\documentclass[12pt,oneside,letterpaper]{report}

\newcommand{\draftfinal}[2]{\ifdefined\draftversion#1\else#2\fi}

\newcommand{\finalonly}[1]{\draftfinal{}{#1}}

\newcommand{\thesistitle}{Deep and Probabilistic Models for Gene Regulatory Network Inference}
\newcommand{\thesisauthor}{Claudia Skok Gibbs}
\newcommand{\thesisadvisor}{Dr. Kyunghyun Cho}
\newcommand{\secondthesisadvisor}{Dr. Richard Bonneau}
\newcommand{\thesisdept}{Data Science}
\newcommand{\gradmonth}{May}
\newcommand{\gradyear}{2026}
\newcommand{\thesisdedication}

\RequirePackage[margin=1in, includefoot, letterpaper]{geometry}

\RequirePackage[prologue]{xcolor}
\definecolor[named]{ThesisBlue}{cmyk}{1,0.1,0,0.1}
\definecolor[named]{ThesisYellow}{cmyk}{0,0.16,1,0}
\definecolor[named]{ThesisOrange}{cmyk}{0,0.42,1,0.01}
\definecolor[named]{ThesisRed}{cmyk}{0,0.90,0.86,0}
\definecolor[named]{ThesisLightBlue}{cmyk}{0.49,0.01,0,0}
\definecolor[named]{ThesisGreen}{cmyk}{0.20,0,1,0.19}
\definecolor[named]{ThesisPurple}{cmyk}{0.55,1,0,0.15}
\definecolor[named]{ThesisDarkBlue}{cmyk}{1,0.58,0,0.21}

\definecolor{SchoolColor}{rgb}{0.3412, 0.0235, 0.5490} 
\definecolor{chaptergrey}{rgb}{0.2600, 0.0200, 0.4600} 
\definecolor{midgrey}{rgb}{0.4, 0.4, 0.4}

\usepackage{hyperref}
\hypersetup{colorlinks,
  linkcolor=ThesisDarkBlue,
  citecolor=ThesisPurple,
  urlcolor=ThesisDarkBlue,
  filecolor=ThesisDarkBlue}

\RequirePackage[labelfont={bf,sf,small,singlespacing},
                textfont={sf,small,singlespacing},
                margin=0pt,
                figurewithin=chapter,
                tablewithin=chapter]{caption}

\usepackage{fix-cm}
\RequirePackage[raggedright,sc]{titlesec}
\definecolor{gray75}{gray}{0.75}
\newcommand{\hsp}{\hspace{20pt}}

\titleformat{\chapter}[hang]
{\Huge\sc}
{\textcolor{SchoolColor}{\thechapter}\hsp\textcolor{gray75}{|}\hsp}
{0pt}{\Huge\sc\raggedright}

\usepackage{setspace}

\finalonly{
  \doublespacing 
}

\usepackage{lipsum}

\newcommand{\nocomments}{}

\newcommand{\notodos}{}

\newcommand{\fcomment}[2]{\ifdefined\nocomments{}\else\footnote{\textcolor{red}{#1:} #2}\fi}

\input{defs}

\begin{document}

\pagenumbering{roman}
%
\thispagestyle{empty}
\vspace*{25pt}
\begin{center}
  {\Large
    \begin{doublespace}
      {\textcolor{SchoolColor}{\textsc{\thesistitle}}}
    \end{doublespace}
  }
  \vspace{.7in}

  by
  \vspace{.7in}

  \thesisauthor
  \vfill

  \begin{doublespace}
    \textsc{
    A dissertation submitted in partial fulfillment\\
    of the requirements for the degree of\\
    Doctor of Philosophy\\
    Center for \thesisdept\\
    New York University\\
    \gradmonth, \gradyear}
  \end{doublespace}
\end{center}
\vfill

\noindent\makebox[\textwidth]{\hfill\makebox[2.5in]{\hrulefill}}\\
\makebox[\textwidth]{\hfill\makebox[2.5in]{\hfill\thesisadvisor}}
\noindent\makebox[\textwidth]{\hfill\makebox[2.5in]{\hrulefill}}\\
\makebox[\textwidth]{\hfill\makebox[2.5in]{\hfill\secondthesisadvisor}}

\newpage

\thispagestyle{empty}
\vspace*{25pt}
\begin{center}
  \scshape \noindent \small \copyright \  \small  \thesisauthor \\
  all rights reserved, \gradyear
\end{center}
\vspace*{0in}
\newpage

\cleardoublepage
\phantomsection
\chapter*{Dedication}
\addcontentsline{toc}{chapter}{Dedication}
\vspace*{\fill}
\begin{center}
    For my parents, who always believed in me, \\
    and for my husband, who carried me through. \\
\end{center}
\vfill
\newpage

\chapter*{Acknowledgements}
\addcontentsline{toc}{chapter}{Acknowledgements}

Before this journey began, I often saw myself through the lens of my struggles with mathematics, and I was not always sure that I belonged in science. Today, I understand that a scientist is not defined by innate talent, but by the determination to keep learning, to grapple with difficult questions, to immerse oneself in complexity, and to find joy in the unknown. It is thanks to my advisors, colleagues, friends, and family that I have come to understand this.

I would first like to thank my advisors Kyunghyun Cho and Richard Bonneau, for their encouragement, kindness, and support over the last five years. Kyunghyun, it has been a pleasure to learn from you and to have been given the freedom to explore the ideas which have always excited me the most. These ideas have benefited enormously from your expertise, creativity, and insight. Rich, I will always be grateful to you for helping launch my scientific career and fostering my love of biology, which ultimately led me to pursue this Ph.D. I have learned so much from both of you, and I am deeply grateful for your mentorship.

I would also like to thank Camille Alexander-Norrell, without whom I would never have been able to wrangle the busy schedules of Kyunghyun and Rich. Thank you for always taking care of me like one of your own, and for nourishing me with lunch, snacks, jokes, and funny gossip. Your love and kindness have carried me through so much of this journey.

To my thesis committee, Carlos Fernandez-Granda, Romain Lopez, and Anirvan Sengupta, thank you for your time, insight, and thoughtful feedback on this dissertation.

There are so many incredible scientists I met along this journey who have also become my closest friends. I would especially like to thank Dongmin for her enthusiasm in our adventures together. Whether it is a lunch date, coffee in the park, or running marathons together, you are always the best company for any activity. I would also like to thank Dan, Harsh, Meet, Lavender, Angie, Maggie, and Omar for their friendship, support, and the insight they brought to so many projects. 

I would also like to thank my early collaborators, who taught me so much useful biology at the beginning of this journey. Chris Jackson, thank you for mentoring me from the very first day and teaching me how to approach a scientific problem. Andreas Tjarnberg, thank you for teaching me how to navigate the tangled mess of biological data and transform it into something clear and beautiful. Neset Ozel, thank you for your collaboration which exposed me to the fascinating world of developmental biology.

Outside of science, I have been surrounded by so many incredible friends who have supported me throughout this journey and have listened to my endless rants about biology, math, and academia. I would like to thank Athena, Kyra, Laurel, Sarah, Cristiana, Aaron, Michelle, Mabelle, Harry, Charlotte, Helene, Romain, Kerda, Randall, Anzar, Katie, Matt, Zoe, Mario, Milly, Jacopo, Irina, and Tanya, for all of the joy and laughter which sustained me throughout this journey.

I am also fortunate to be supported by several workout coaches, who gave me a physical outlet from the mental strain of this thesis. Thank you to Dale Elston, Eric Salvador, Jennifer Spina, and Mike Keohane for reminding me through exercise that I am always capable.

I come from a large family that has supported and loved me from the very first day. To my sisters, Thoma, Victoria, and Sophie, and my brothers, Benjamin, and Sebastian, I am never alone because of the five of you. Thank you for your love, kindness, and for always being there whenever I needed someone to lean on. To their spouses, who have also become like siblings to me, Stuart, Chris, Max, Emma, and Melissa, thank you for loving my favorite people, and for becoming such an important part of our family’s wonderful chaos. To my uncles, David, and Michael, thank you for our adventures, but most of all, for always encouraging me in my pursuit of mathematics and science. Your belief in me and your confidence in my future helped me believe in myself.

In the first year of my Ph.D., I made the best decision of my life, which gave me four additional family members who have made all the difference in this journey. I would first like to thank my mother-in-law, whom we lost in 2022, and whose absence is felt with irreparable sadness. Evelyne, thank you for teaching me that true strength means showing up even on days when it feels hardest, and for showing me that love and kindness are the greatest ambitions of all. Thank you to my father-in-law, Habib, for your love, kindness, and patience. I have learned so much from you, both in language and in life, and have benefited so much from your generosity. Thank you to my brother- and sister-in-law, Allan and Carla, for bringing so much laughter and fun to our adventures together. I cannot wait for the many more to come.

I am so incredibly lucky to have parents who are not only my best friends and biggest supporters, but also my greatest source of motivation and inspiration. To my parents, Jane and Paul, your love, support, and encouragement have shaped every part of my life. Thank you for believing in me so fully, for supporting me through every stage of this Ph.D., and for always reminding me that I was capable of more than I sometimes believed. I love you both more than I could ever adequately put into words. 

Finally, to my husband, Yanis, for whom words will never do justice, in English or in French, I will never find enough ways to thank you for what your partnership has brought to my life and to this Ph.D. You have supported every part of this journey, from helping me through homework assignments to listening patiently to every research idea, frustration, and breakthrough. You picked me up in the moments I doubted myself, reminded me of my strength when I could not see it, and motivated me to give each day my best. I am endlessly grateful not only for everything you have done, but for the love, steadiness, and joy you bring to my life every day. Thank you for making this dream of mine possible.
\newpage

\chapter*{Abstract}
\addcontentsline{toc}{chapter}{Abstract}
Gene regulatory networks (GRNs) link transcription factor (TF) proteins to their target genes, yet reconstructing these networks from genome-wide data remains challenging under practical and methodological constraints. Many methods couple modeling assumptions to a specific inference procedure and rely on heuristic model selection, while evaluation is constrained by incomplete reference networks and point-estimate outputs that lack uncertainty. GRN reconstruction also depends on prior knowledge to constrain TF–gene interactions, yet available priors are often assay-dependent and difficult to transfer across species and less-characterized systems. In this thesis, we develop two complementary frameworks that address these limitations. In the first, PMF-GRN casts GRN inference as a probabilistic graphical model optimized by variational inference, enabling principled model selection and uncertainty-aware edge estimates. In the second, GLM-Prior addresses the prior bottleneck by fine-tuning the pretrained Nucleotide Transformer to predict TF–target gene interactions directly from nucleotide sequence, while generalizing across yeast, mouse, and human settings. Together, PMF-GRN and GLM-Prior motivate a dual-stage view of GRN reconstruction in which sequence-derived priors provide a transferable starting scaffold and probabilistic inference refines regulatory estimates with quantified uncertainty under incomplete evaluation resources.

\newpage

\tableofcontents

\cleardoublepage
\phantomsection
\addcontentsline{toc}{chapter}{List of Figures}
\listoffigures
\newpage

\cleardoublepage
\phantomsection
\addcontentsline{toc}{chapter}{List of Tables}
\listoftables
\newpage

\pagenumbering{arabic} 


\chapter{Introduction}
\label{chp-introduction}
\label{sec:thesis-introduction}

Gene regulatory networks (GRNs) provide a structured representation of transcriptional control by mapping regulatory relationships between transcription factors (TFs) and their target genes \cite{karlebach2008modelling, he2016understanding}. These networks offer a useful lens for studying how coordinated gene expression programs arise in development \cite{ozel2022coordinated}, adapt to environmental signals \cite{jackson2020gene}, and become disrupted in disease \cite{emad2021inference, unger2024gene}. A central challenge, however, is that GRNs cannot be directly measured at genome scale with standard sequencing assays. Instead, regulatory structure must be inferred from high-throughput genomic data that provide indirect, incomplete, and often noisy views of the underlying regulatory processes \cite{badia2023gene}. 

This thesis is motivated by the observation that the difficulty of GRN inference is driven as much by the limitations of available data as by the modeling choices made downstream. First, most experimental assays provide only indirect views of regulation. Expression measurements are noisy and high-dimensional, particularly in single-cell settings where sparsity, dropout, and sampling variability are substantial \cite{nesari2026advances}. Additionally, transcriptomic datasets provide precise snapshots taken at unknown points along the regulatory cascade \cite{gorin2022rna}, causing contemporaneous TF-target gene correlations to be weak or misleading when regulatory effects are delayed. 

Regulation is also combinatorial, with genes integrating inputs from multiple TFs, and each TF influencing many potential target genes, expanding the space of plausible explanations for any observed expressed pattern \cite{dubois2018organizing}. Complementary modalities often help constrain this search space by providing mechanistic context beyond expression alone \cite{miraldi2019leveraging, skok2022high}. For example, chromatin accessibility assays such as ATAC-seq can narrow attention to genomic regions that are permissive to regulation, and TF binding assays such as ChIP-seq can identify loci occupied by a specific TF under a particular condition. At the same time, these measurements remain context dependent and incomplete. Accessibility and occupancy capture regulatory potential and binding, but they do not themselves establish functional influence on transcription or the sign and magnitude of regulatory effects \cite{slattery2014absence, cusanovich2014functional}. 

Limitations in available data for GRN inference also carries through to evaluation. Reference networks and gold standards are typically assembled from incomplete databases and context-specific experimental evidence, capturing only a subset of true regulation in any given system \cite{hegde2025machine}. As a result, many true edges are missing from these reference networks, with unlabeled TF-gene pairs usually reflecting a lack of evidence or coverage, rather than providing evidence that an interaction does not occur \cite{kernfeld2024transcriptome}. This partial observability of both regulation and evaluation presents GRN inference as an ill-posed inverse problem \cite{schafer2005empirical}, where many distinct regulatory explanations can be consistent with the same observed data.

Beyond data limitations, existing GRN inference methods introduce practical constraints that further limit what can be inferred reliably in practice. Many existing algorithms tightly couple their statistical model to a specific inference procedure, often embedding assumptions that are tailored to a particular data modality or sampling regime \cite{hu2020integration}. As genomic technologies have shifted from microarrays to bulk RNA-seq and then to single-cell measurements, GRN inference methods have frequently required redesign to accommodate new measurement characteristics, rather than providing a stable modeling scaffold capable of absorbing new assumptions. 

Even within a fixed data regime, additional methodological choices further shape what can be inferred. Model selection is often treated heuristically, with many approaches committing to a single algorithmic form or fixed set of hyperparameters without systematically comparing alternative modeling choices, even though different assumptions can yield qualitatively different networks from the same data. Finally, GRNs are commonly reported as point estimates without an accompanying measure of confidence, despite the fact that evaluation typically relies on incomplete and context-dependent references. Without uncertainty estimates, it becomes difficult to separate robust interactions from weakly identified edges, and to prioritize predictions in settings where labels are sparse, missing, or context-mismatched. 

In addition to inference and model selection limitations, the accuracy of GRN reconstruction is often determined by the availability and quality of prior knowledge. In realistic single-cell settings, expression data alone is insufficient to anchor regulatory structure, and meaningful results typically require incorporating prior evidence that breaks symmetries and restricts attention to biologically plausible edges. In current pipelines, priors are most commonly constructed by combining experimental and computational signals, for example by linking TF motifs to genes through regions of open chromatin, or by aggregating TF binding evidence where it exists \cite{skok2022high, kamimoto2023dissecting, van2020scalable}. These strategies are powerful, but they also remain limited as they tend to emphasize promoter-proximal interactions and depend on cell-type specific assays that are unavailable in many contexts. In model organisms, curated regulatory databases provide an alternative source of prior edges, but these remain incomplete even across TFs and conditions. 

Together, these methodological limitations motivate the need for GRN inference frameworks that decouple model specification from the inference algorithm, support principled model selection across modeling assumptions and hyperparameters, and provide uncertainty-aware network estimates that can be interpreted and prioritized even when reference labels are incomplete or unavailable. These limitations also motivate the need for priors that do not depend on cell type-specific experimental assays, can capture regulatory logic beyond promoter-proximal interactions, and support the transfer of regulatory information from well-studied organisms to less-characterized systems, so that downstream inference remains well constrained even when direct prior evidence is sparse or unavailable.

The remainder of this thesis is organized to first establish the biological and methodological context for GRN inference, and then to develop two complementary approaches that address the central limitations outlined above. In Chapter \ref{sec:thesis-background}, we first situate GRNs within the central dogma by emphasizing the feedback loop in which TF proteins return to the genome to shape transcription, motivating a network-based representation of regulation. We then survey the experimental modalities that make GRN reconstruction possible in practice, including transcriptomic assays and chromatin-based assays that provide observable and mechanistic context. Finally, we introduce the machine learning framing used throughout the remainder of this thesis, describing inference over latent regulatory structure under strong data limitations, the role of priors in anchoring interpretability and identifiability, and the consequence of incomplete reference networks for evaluation. Together, this background chapter provides the conceptual toolkit required to understand why GRN inference is challenging, what information different data modalities contribute, and how modeling assumptions determine what can be learned from modern genomics datasets. 

Chapter \ref{sec:PMF-GRN-Chapter} then focuses on the inference problem itself and introduces a probabilistic framework designed to address methodological limitations that arise even when appropriate data and priors are available \cite{skok2024pmf}. The chapter develops a generative, latent-variable view of GRN inference in which TF activities and TF-gene influences are treated as unobserved quantities that must be inferred from noisy expression measurements. By separating model specification from the inference procedure, this framework makes it possible to modify distributional assumptions and incorporate additional biological structure without redesigning the optimization machinery used for inference. The probabilistic formulation also enables principled model selection by comparing alternative modeling assumptions and hyperparameters in a systematic way, rather than relying on a single default algorithmic configuration. Finally, posterior uncertainty provides an explicit notion of confidence for inferred TF-gene interactions, supporting uncertainty-aware prioritization in settings where evaluation resources are incomplete or unavailable.

Chapter \ref{sec:GLM-Prior-Chapter} shifts the emphasis from inference to prior construction, motivated by the observation that the quality and availability of prior knowledge often determines whether GRN inference is usable in practice. This chapter introduces a sequence-based approach for constructing TF-gene prior networks that does not rely on cell type-specific experimental assays and can generalize beyond the limited contexts where curated priors exist \cite{gibbs2025glm}. Instead of deriving priors primarily from promoter-proximal accessibility and motif heuristics, the approach learns regulatory sequence features directly from data using a transformer-based genomic language model, enabling prior construction at genome scale. A central goal of this chapter is to characterize how such priors generalize across settings, including single-species training, transfer learning between organisms, and multi-species training. Further, we demonstrate how stronger, more transferable priors improve downstream GRN reconstruction in complex mammalian systems where experimental priors are often noisy, incomplete, or unavailable.

Together, Chapters \ref{sec:PMF-GRN-Chapter} and \ref{sec:GLM-Prior-Chapter} support a dual-stage perspective on GRN reconstruction in which prior construction and expression-based inference play distinct, complementary roles. Prior knowledge provides TF-specific structure that narrows the space of plausible regulatory explanations and anchors interpretability, while expression-based inference refines this scaffold in a context-dependent manner and quantifies uncertainty in the resulting network. This division of labor directly addresses the challenges outlined above, where noisy and incomplete measurements, imperfect and context-dependent evaluation resources, and methodological constraints have historically limited robustness and interpretability of GRN inference. The central argument developed across this thesis is that progress in GRN reconstruction depends jointly on inference frameworks that are flexible, principled, and uncertainty-aware, as well as on priors that can capture regulatory logic beyond promoter-proximal signals and can transfer across biological systems where direct regulatory evidence is sparse or unavailable.

\chapter{Background}
\label{chp-background}
\label{sec:thesis-background}

This chapter lays the groundwork for the chapters that follow by introducing essential context on topics related to gene regulatory network inference. In Section \ref{sec:central-dogma}, we describe the central dogma of biology and its natural connection to gene regulatory networks, highlighting how transcriptional regulation introduces feedback that links gene expression to protein function. This perspective motivates a network-based view of regulation which is formalized in Section \ref{sec:gene-reg-nets} through the definition of gene regulatory networks. Following this, in Section \ref{sec:grn-data}, we introduce the major experimental data modalities that provide indirect or partial evidence of regulatory activity and form the empirical basis for gene regulatory network inference. 

In Section \ref{sec:ml-foundations-grn-inference}, we then define the GRN inference problem from a machine learning perspective, and outline key challenges that arise from both the properties of modern genomics datasets and the complexity of transcriptional regulation. Then, we review the major families of GRN inference algorithms and introduce the statistical foundations that will be used throughout this thesis, including latent variable modeling, probabilistic generative frameworks with variational inference, and sequence-based models for construction informative prior knowledge. Finally, we describe evaluation metrics used throughout this thesis to evaluate the predictive accuracy of inferred GRNs using available reference gold standards.

\section{The Central Dogma of Biology}
\label{sec:central-dogma}

\begin{figure}[!htbp]
    \centering
    \includegraphics[width=1.0\textwidth]{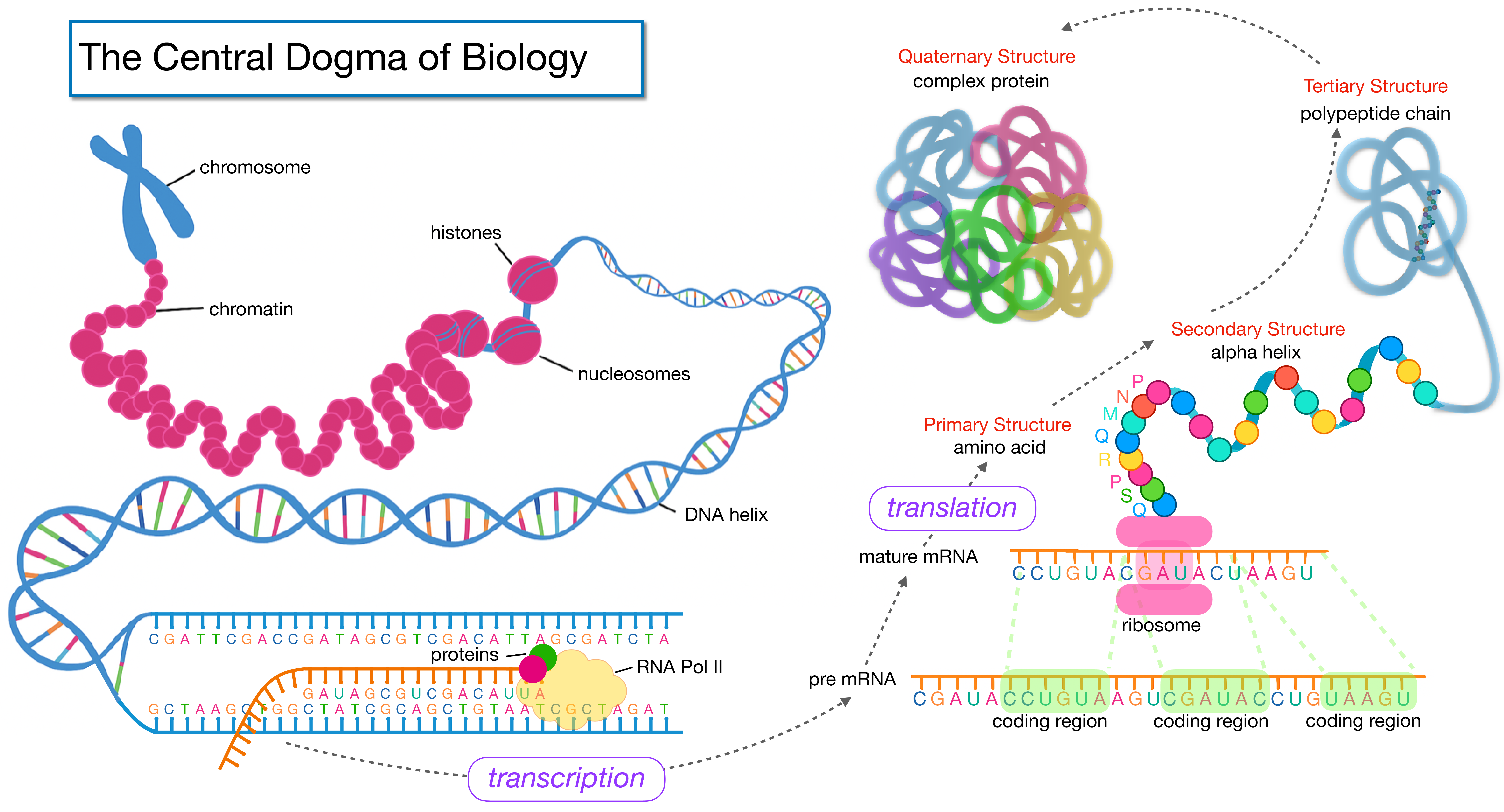} 
    \caption{The central dogma of biology, illustrating the flow of genetic information from DNA to RNA via transcription and from RNA to protein via translation.}
    \label{fig:central-dogma}
\end{figure}

The central dogma of biology describes the flow of information within living systems, where genetic information encoded in the DNA is transcribed into RNA, and then translated into protein (Figure \ref{fig:central-dogma}) \cite{crick1970central}. DNA is a complex double helix molecule consisting of four chemical bases, adenine (A), thymine (T), cytosine (C), and guanine (G), \cite{klug2004discovery}, and contains the genetic instructions required for the development, function, and growth of all known organisms. Within this sequence, genes define specific, heritable regions of the genome that are transcribed to produce RNA molecules \cite{chaffey2003alberts}. RNA shares many structural similarities with DNA, but differs by using uracil (U) in place of thymine (T), giving it a nucleotide alphabet of A, U, C, and G. 

RNA is synthesized through a process called transcription, in which the DNA sequence of a gene is read by the enzyme RNA polymerase to produce a complementary RNA strand. To initiate transcription, the DNA double helix locally unwinds, allowing RNA polymerase to bind to the promoter (or start of a gene), and begin RNA synthesis \cite{borukhov2008rna}. As RNA polymerase moves along the DNA template, it assembles the RNA transcript by sequentially adding RNA bases that are complementary to the DNA bases \cite{abbondanzieri2005direct}. The resulting RNA molecule may function directly as a non-coding RNA, or more commonly, as messanger RNA (mRNA), which carries the gene's information forward and acts as a template to build proteins \cite{cooper2022cell}.

Proteins are functional molecules responsible for executing most cellular processes. Unlike DNA and RNA, which are composed of four nucleotide bases, proteins are made up of chains of $20$ different amino acids, giving rise to a more complex biochemical alphabet \cite{chaffey2003alberts}. Proteins synthesis occurs through a process called translation, in which the nucleotide sequence of an mRNA molecule is read in groups of three bases, known as codons \cite{cooper2022cell}. Each codon corresponds to a specific amino acid, which is added sequentially to a growing polypeptide chain. Once synthesized, this chain folds into a three-dimensional structure, with this structure largely determining the protein's biological function \cite{sanvictores2020biochemistry}. 

Among the many types of proteins produced in the cell, transcription factors (TFs) play a central role in controlling gene regulation. TFs are DNA-binding proteins that recognize specific sequence motifs and modulate transcription by influencing the recruitment and activity of the transcriptional machinery, including RNA polymerase II, at gene regulatory regions \cite{latchman1997transcription, lambert2018human}. In eukaryotes, TFs act at gene promoters as well as at distal regulatory elements such as enhancers. TF-mediated regulation is typically combinatorial; a single gene integrates input from multiple TFs, and the regulatory outcome depends on the combination of TF partners, cofactors, and the broader cellular context \cite{mitsis2020transcription}. This complexity allows cells to precisely control when transcription is initiated and to what extent each gene is expressed.

In this way, the central dogma of biology, describing the flow of information from DNA to RNA to protein via transcription and translation, serves as the backbone for gene expression, while gene regulation determines how this flow is modulated across space, time, and condition. Importantly, regulation introduces feedback that closes the loop; many proteins produced during translation, including TFs, return to bind DNA, and influence subsequent rounds of transcription. Because each TF can regulate multiple genes, and each gene can be regulated by multiple TFs, this many-to-many, combinatorial architecture gives rise to complex, distributed control systems. Feedback interactions between TFs and their target genes generate coordinated gene expression programs across the genome. This structure motivates a network-based perspective of regulation, formalized as gene regulatory networks (GRNs), which is the focus of the next section.

\section{Gene Regulatory Networks}
\label{sec:gene-reg-nets}
\begin{figure}[!htbp]
    \centering
    \includegraphics[width=1.0\textwidth]{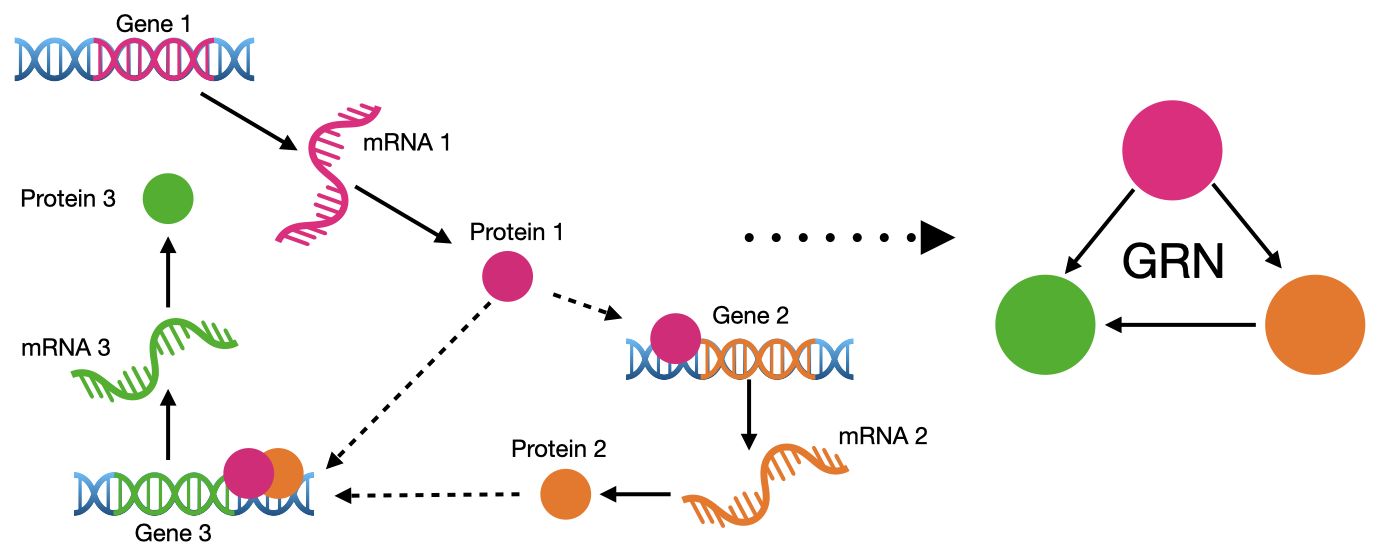} 
    \caption{Schematic linking the central dogma to gene regulatory networks. A gene is transcribed into mRNA and translated into protein; when this protein is a transcription factor, it binds to DNA to regulate the transcription of a target gene, initiating a subsequent round of transcription and translation. This repeated feedback process is summarized as a directed graph, where nodes represent genes (or TFs) and directed edges represent regulatory influence from TFs to their target genes.}
    \label{fig:grn-schematic}
\end{figure}

The feedback described in the previous section motivates a simple but powerful abstraction for transcriptional control. When a gene is transcribed and translated, the resulting protein can act back on the genome to influence the transcription of other genes. In particular, TF proteins bind specific regions of DNA and modulate the transcription of their target genes, thereby initiating new rounds of RNA production and protein synthesis. Gene regulatory networks (GRNs) encapsulate this process by representing the relationships between TFs and their target genes within a biological system (illustrated in Figure \ref{fig:grn-schematic}) \cite{delgado2019computational, badia2023gene}.

GRNs are commonly represented as directed graphs \cite{schlitt2007current}. Formally, a GRN can be defined as a graph $G = (V, E)$, where $V = \{1, \dots, p\}$ is the set of nodes corresponding to genes (or gene products, such as TFs), and $E \subseteq V \times V$ is the set of directed edges representing regulatory influences. For each node $j \in V$, the parent set $Pa(j) = \{i \in V : (i \rightarrow j) \in E\}$ denotes the regulators of gene $j$. In this representation, an edge from TF $A$ to gene $B$ indicates that the activity of TF $A$ influences the expression of gene $B$. Importantly, regulation is not governed by isolated one-to-one relationships. Individual TFs typically regulate many genes, individual genes integrate inputs from multiple TFs, and the sign and strength of regulation can depend on the cellular context \cite{hobert2008gene}. The central goal of constructing a GRN is to determine which TFs are responsible for the activation or repression of each gene in a given biological setting, and to do so at genome scale \cite{babu2004structure, marbach2012wisdom, badia2023gene}.

Reconstructing these relationships provides a systems-level view of regulatory programming. A GRN offers a mechanistic explanation for observed gene expression patterns and supports hypotheses about unobserved drivers of phenotypes, such as latent regulatory activities or upstream perturbations. By uncovering regulatory relationships between TFs and their target genes, GRNs can help clarify mechanisms that distinguish healthy from diseased cellular states and can inform downstream analyses relevant to drug discovery and targeted therapies \cite{kim2023gene}. GRNs also provide a framework for understanding developmental processes and cellular differentiation, where cell fate transitions reflect changes in regulatory programs over time \cite{allaway2021genetic, ozel2022coordinated}. More broadly, GRNs can be used to study how regulatory mechanisms evolve and diverge across organisms, and how biological systems respond to environmental stimuli or stressors through coordinated changes in transcriptional control \cite{jackson2020gene}.

A central challenge of reconstructing GRNs is that the TF-target gene edges are not directly observable at scale with standard sequencing assays. While specific regulatory interactions can be validated experimentally, current technologies do not provide a complete, direct readout of TF-target gene relationships across all genes and conditions. Reconstructing a GRN therefore relies on integrating measurable genomic data to infer the most plausible regulatory landscape and underlying regulatory connections \cite{skok2022high, skok2024pmf, kamimoto2023dissecting}. This motivates the next section, which describes the major data modalities and experimental settings that provide evidence for regulatory relationships and define the practical regimes in which GRNs can be inferred.

\section{Data Modalities for Gene Regulatory Networks}
\label{sec:grn-data}
The reconstruction of GRNs relies on integrating genomic data that provides indirect or partial evidence of regulatory activity. TF binding and target gene regulation are typically not observable at scale through direct measurement, requiring GRN inference to draw upon diverse data modalities that reflect the downstream consequences or correlates of regulation. These data modalities include gene expression profiles across conditions, chromatin accessibility landscapes, TF binding data, and DNA sequence features that encode regulatory specificity. Each data type captures a different aspect of the regulatory system, with its own strengths, biases, and assumptions. 

The order in which these technologies emerged, beginning with microarrays, followed by bulk RNA-seq, and more recently single cell RNA-seq, has shaped how transcriptional activity is measured and continues to influence the types of biological questions that can be addressed. This section describes the major experimental modalities used to characterize regulatory activity and provides context for the types of data available in modern genomics to reconstruct GRNs.

\subsection{Microarray Gene Expression Profiling}
\label{sec:microarray}

\begin{figure}[!htbp]
    \centering
    \includegraphics[width=1.0\textwidth]{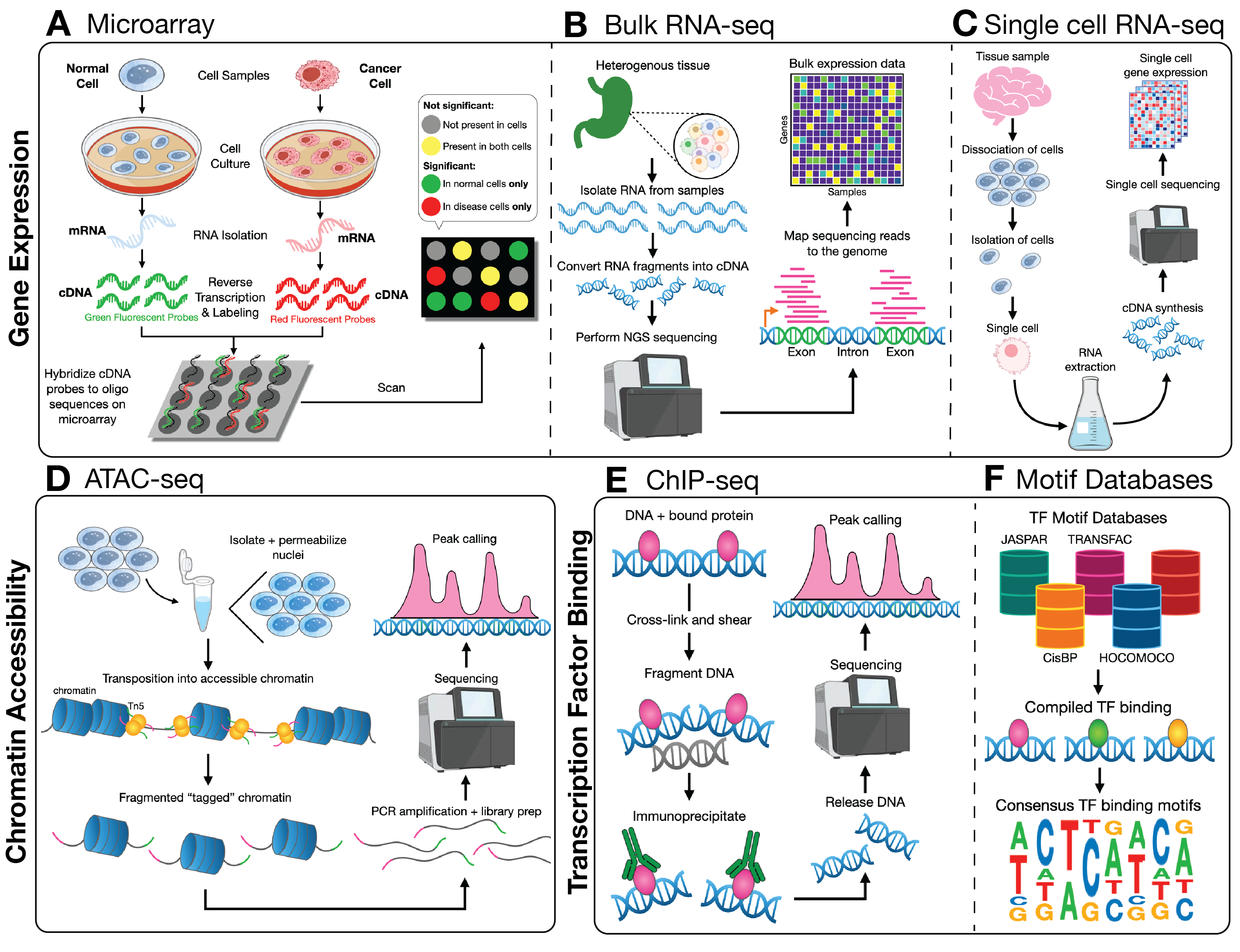} 
    \caption{Schematic overview of data modalities commonly used to reconstruct gene regulatory networks. (\textbf{A}) Microarray workflow: labeled cDNA hybridizes to probes on a fixed array to measure transcript abundance via fluorescence intensity. (\textbf{B}) Bulk RNA-seq: population-level transcriptomes are sequenced to quantify gene expression across conditions or samples. (\textbf{C}) Single-cell RNA-seq: RNA from individual cells is barcoded, sequenced, and aggregated into a cell-by-gene count matrix. (\textbf{D}) ATAC-seq: a transposase preferentially inserts adapters into accessible chromatin, enabling the identification of open regulatory regions. (\textbf{E}) ChIP-seq: immunoprecipitation of DNA-bound TFs allows direct measurement of TF binding sites genome-wide. (\textbf{F}) Motif databases: position weight matrices (PWMs) derived from experimental binding data represent sequence-specific TF binding preferences and are used to scan genomes for candidate binding sites.}
    \label{fig:grn-data}
\end{figure}

DNA microarrays were one of the earliest genome-scale technologies for measuring gene expression, providing the first high-throughput view of transcriptional programs across conditions. In a microarray experiment, cellular RNA is extracted from a biological sample and reverse-transcribed into fluorescently labeled complementary DNA (cDNA) (illustrated in Figure \ref{fig:grn-data}A). cDNA is then hybridized to an array containing thousands of short oligonucleotide probes, each designed to target a specific gene or transcript \cite{jaksik2015microarray}. The strength of the hybridization between the cDNA and each probe is measured via fluorescence intensity, which serves as a proxy for transcript abundance. Since each probe set is predefined, microarrays can only measure transcripts that are already annotated and represented on the array \cite{dufva2009introduction, blohm2001new, howbrook2003developments}. 

Microarrays are widely used in regulatory studies because they provide an efficient and cost-effective snapshot of gene expression across diverse conditions, perturbations, and timepoints \cite{tarca2006analysis}. Such measurements are often used to identify genes that change together across biological states, offering indirect evidence for shared regulation \cite{lee2004coexpression}. 

Computationally, microarray data is represented as a gene expression matrix $W \in \mathbb{R}^{N \times M}$, where $N$ is the number of samples or experimental conditions, and $M$ is the number of genes (or transcripts) probed on the array \cite{smyth2005limma}. Each entry $W_{n,m}$ contains the normalized, often log-transformed fluorescence intensity for gene $m$ in sample $n$, reflecting its estimated expression level \cite{ritchie2015limma}. These matrices provide a standardized representation of expression dynamics across conditions and form the basis for many statistical analyses of transcriptional regulation.

Several limitations affect the interpretability of microarray data \cite{russo2003advantages, abdullah2006technology}. First, because probe design is limited to known transcripts, microarrays lack the ability to detect unannotated genes, novel isoforms, or transcript structure \cite{jaluria2007perspective}. Second, cross-hybridization, where similar sequences partially bind non-target probes, can introduce noise and confound gene-specific quantification \cite{murphy2002gene}. Third, the dynamic range of fluorescence measurements is limited \cite{nadon2002statistical}. Highly expressed genes may saturate the signal, while weakly expressed genes may fall below the detection threshold, leading to distorted estimates at expression extremes. Finally, microarray data are prone to batch effects and platform-specific artifacts, which hinder integration across studies and require rigorous normalization \cite{forster2003experiments}. Despite these challenges, microarrays provided the first genome-scale views of transcriptional activity and played a foundational role in early efforts to uncover regulatory structure from gene expression data.

\subsection{Bulk RNA Sequencing}
\label{sec:RNA-seq}

Bulk RNA-sequencing (RNA-seq) emerged as a sequencing-based alternative to microarrays, offering a more flexible and quantitative platform for transcriptome-wide expression profiling \cite{wang2009rna}. Unlike microarrays, which rely on pre-defined probes, RNA-seq does not require prior knowledge of transcript sequences, making it suitable for detecting novel genes, splice variants, and isoforms. In a typical RNA-seq workflow (Figure \ref{fig:grn-data}B), total RNA is extracted from a population of cells, converted into a library of cDNA, and sequenced using high-throughput sequencing technologies. The resulting reads are aligned to a reference genome or transcriptome, and gene- or transcript-level abundances are estimated based on the number of aligned reads \cite{marguerat2010rna}.

RNA-seq provides a high-resolution view of gene expression programs across experimental conditions, perturbations, or timepoints \cite{marguerat2008next}. It offers a wider dynamic range than microarrays and can more accurately quantify both lowly and highly expressed transcripts, making it an adopted modality for studying transcriptional regulation and for generating datasets used in GRN analysis \cite{zhao2014comparison, rai2018advantages}.

Computationally, RNA-seq data is typically represented as a gene expression matrix $W \in \mathbb{R}^{N \times M}$, where $N$ is the number of samples and $M$ is the number of genes (or transcripts) included in the quantification. Each entry $W_{n,m}$ denotes the normalized expression level of gene $m$ in sample $n$, commonly reported on a log scale after normalization procedures such as transcripts per million (TPM), counts per million (CPM), or variance-stabilizing transformations \cite{ghosh2016analysis}. This matrix serves as a standardized representation of transcript abundance across samples and is used as a primary input for downstream statistical and network-based analysis.

Several limitations of bulk RNA-seq affect the interpretation and downstream modeling of expression measurements. First, the approach aggregates RNA from many cells within a sample, yielding an average expression profile that can obscure heterogeneity among individual cells \cite{hegenbarth2022perspectives}. As a result, gene expression differences arising from rare cell types or dynamic regulatory states may be masked. Second, RNA-seq data are subject to technical variation stemming from library preparation protocols, sequencing depth, and alignment biases \cite{seqc2014comprehensive, degner2009effect}. These factors can introduce systematic shifts in expression estimates across samples, necessitating careful normalization and experimental design to ensure comparability. Despite these challenges, bulk RNA-seq has become a useful tool for transcriptomics and continues to assist large-scale efforts to characterize transcriptional programs in health and disease \cite{costa2013rna}.

\subsection{Single-Cell RNA Sequencing}
\label{sec:scrna-seq}

Single-cell RNA-sequencing (scRNA-seq) was developed to measure gene expression at the resolution of individual cells, addressing a key limitation of bulk RNA-seq, which averages expression across heterogeneous populations \cite{saliba2014single}. In complex tissues composed of diverse cell types and dynamic cellular states, such averaging can obscure important regulatory differences \cite{kolodziejczyk2015technology}. scRNA-seq preserves this heterogeneity by isolating and profiling the transcriptomes of thousands to millions of individual cells within a single experiment. In a typical scRNA-seq workflow (Figure \ref{fig:grn-data}C), single cells are encapsulated, barcoded, and lysed, after which their RNA is reverse-transcribed into cDNA, amplified and sequenced \cite{slovin2021single}. Barcodes enable each read to be assigned to its cell of origin, resulting in a cell-by-gene matrix that reflects transcriptional variation across individual cells.

The level of resolution achieved by scRNA-seq has transformed the study of complex biological systems. scRNA-seq enables the identification of discrete cell types, the characterization of continuous developmental or differentiation trajectories, and the exploration of gene expression patterns specific to transient or rare cell states \cite{chen2019single}. These properties make scRNA-seq an especially valuable modality for studying dynamic and cell-type specific regulatory programs, and for generating data that can inform models of gene regulation at single-cell resolution.

scRNA-seq data is typically represented as a matrix $W \in \mathbb{R}^{N \times M}$, where $N$ is the number of individual cells and $M$ is the number of genes. Each entry $W_{n,m}$ records the number of unique molecular identifiers (UMIs) detected for gene $m$ in cell $n$, serving as a count of RNA molecules \cite{luecken2019current}. Total UMI counts can vary substantially between cells, due to differences in RNA content, capture efficiency, or sequencing-depth, and can cause raw counts to not be directly comparable. Downstream analyses typically operate on normalized or transformed versions of $W$, where methods such as library size normalization, log-transformation, or variance stabilization are used to mitigate technical variability and improve interpretability of gene expression differences across cells \cite{hafemeister2019normalization}. 

Several technical challenges affect scRNA-seq data. First, expression matrices are characteristically sparse and noisy, with many entries containing zeros due to limited RNA capture per cell, a phenomenon often referred to as dropout \cite{van2018recovering, qiu2020embracing}. This sparsity complicates downstream modeling and can obscure signals from lowly expressed genes. Second, technical variability across cells, including batch effects, differences in sequencing depth, and variation capture efficiency, can introduce systematic biases that must be carefully controlled \cite{tung2017batch}. Third, observed variability across cells may reflect a combination of true biological differences and technical noise, making it difficult to disentangle meaningful regulatory signals without appropriate processing and modeling strategies \cite{vallejos2017normalizing}. Despite these challenges, scRNA-seq has become a central tool in modern genomics and provides a rich modality for studying gene regulation at cellular resolution \cite{saliba2014single}.

\subsection{ATAC Sequencing for Chromatin Accessibility}
\label{sec:atac-seq}

While transcriptomic technologies measure the consequence of gene regulation, chromatin accessibility assays provide insight into the regulatory landscape itself. The Assay for Transposase-Accessible Chromatin using sequencing (ATAC-seq) \cite{buenrostro2013transposition} is a widely used method for identifying regions of open chromatin, which are often indicative of active regulatory elements such as promoters, enhancers, and transcription factor binding sites \cite{grandi2022chromatin}. Open chromatin is characterized by nucleosome depletion, allowing regulatory proteins to access the DNA. ATAC-seq leverages this property by using a hyperactive transposase enzyme (Tn5) to preferentially insert sequencing adapters into accessible regions of the genome. These tagged fragments are then PCR amplified and sequenced, enabling genome-wide mapping of chromatin accessibility with high resolution and minimal input material (Figure \ref{fig:grn-data}D) \cite{buenrostro2015atac}.

Following sequencing, ATAC-seq data undergoes alignment to a reference genome, after which regions of high read density are identified through a process known as peak calling \cite{yan2020reads}. Peaks represent genomic intervals where the transposase inserted more frequently, suggesting accessible chromatin. These peaks can be classified as open regulatory elements and annotated relative to nearby genes \cite{sun2019detect}. Importantly, accessibility at a given region varies across cell types and conditions, reflecting context-specific regulatory activity \cite{grandi2022chromatin, corces2018chromatin}. Peaks overlapping gene promoters often indicate transcriptional readiness, while distal peaks may correspond to enhancers or other regulatory elements.

Bulk ATAC-seq can be represented as a binary or quantitative matrix $C \in \mathbb{R}^{N \times P}$, where $N$ is the number of samples and $P$ is the number of genomic regions (e.g., peaks) identified across the dataset. Each entry $C_{n,p}$ indicates whether region $p$ is accessible in sample $n$, often quantified by read counts or normalized signal intensity (e.g., reads per kilobase of transcript per million mapped reads or counts per million). For downstream analyses, these regions can be further linked to genes by proximity (e.g., nearest transcription start site), overlapped with promoter annotations, or through regulatory annotations such as chromatin state maps or enhancer-gene linkage datasets \cite{nasser2021genome}.

ATAC-seq provides important complementary information to expression-based modalities. While transcriptomic data reflect the outcome of the regulatory processes, chromatin accessibility captures upstream regulatory potential and TF binding opportunity. However, ATAC-seq still presents several limitations \cite{yan2020reads}. First, peak calling can be sensitive to sequencing depth and noise, particularly in regions with low accessibility or high background signal \cite{grandi2022chromatin}. Second, many accessible regions are distal to known genes, complicating assignment of regulatory influence. Third, bulk ATAC-seq represents an average over many cells, potentially obscuring cell type-specific regulatory elements \cite{buenrostro2015single}. While single cell ATAC-seq technologies have been developed to address this limitation, they introduce sparsity and increase noise \cite{ji2020single}. Despite these challenges, ATAC-seq remains a core modality for probing the regulatory landscape and identifying candidate cis-regulatory elements genome-wide.

\subsection{ChIP Sequencing for Transcription Factor Binding}
\label{sec:chip-seq}

While chromatin accessibility assays such as ATAC-seq provide indirect evidence of regulatory potential, chromatin immunoprecipitation follow by sequencing (ChIP-seq) enables direct measurement of TF binding at specific genomic loci \cite{mardis2007chip}. In a ChIP-seq experiment (Figure \ref{fig:grn-data}E), proteins are crosslinked to DNA in living cells, the chromatin is fragmented, and an antibody specific to the TF of interest is used to immunoprecipitate the protein-DNA complexes \cite{park2009chip}. After reversing the crosslinks, the co-precipitated DNA is purified, sequenced, and mapped to a reference genome. Regions enriched for sequencing reads reflect genomic loci bound by the targeted TF under the profiled conditions. 

Following sequencing, downstream analysis of ChIP-seq data involves peak calling to identify regions of significant enrichment compared to background, using control input samples or statistical models to correct for sequencing biases \cite{thomas2017features}. These peaks represent putative binding sites of the immunoprecipitated protein and are often annotated to known regulatory elements such as gene promoters, enhancers, or other cis-regulatory features \cite{jeon2020comparative}. ChIP-seq is therefore a direct method for characterizing TF binding landscapes genome-wide, under specific cellular and environmental conditions. 

ChIP-seq data can be represented as a binary or continuous signal track across the genome, with enriched regions (peaks) summarized into a matrix $T \in \mathbb{R}^{K \times P}$, where $K$ is the number of TFs profiled and $P$ is the number of genomic loci or peaks identified across experiments. Each entry $T_{k,p}$ indicates whether a TF $k$ binds to a region $p$, or reflects the binding strength at that region, depending on whether peaks or signal intensities are used. These peaks can then be linked to candidate target genes based on proximity or experimentally informed annotations, yielding TF-gene interaction datasets that are widely used in regulatory genomics.

ChIP-seq offers high-resolution, TF-specific information that is widely valuable for mapping the regulatory wiring of the genome. However, there are several limitations to the ChIP-seq approach. First, it requires high-quality, TF-specific antibodies, which are often unavailable for all proteins or species, limiting general applicability \cite{tahara2025unmeasured}. Second, ChIP-seq typically profiles one TF at a time, making large-scale experiments resource-intensive \cite{park2009chip}. Third, peak calling can be confounded by noise, indirect binding, or chromatin accessibility biases, and binding does not always equate to functional regulation \cite{nakato2017recent}. Finally, ChIP-seq reflects binding in a particular cellular context and timepoint, making it difficult to generalize across dynamic or heterogeneous systems. Despite these challenges, ChIP-seq remains the gold standard for identifying in vivo TF binding sites and has been essential for building reference regulatory maps in many model systems.

\subsection{Transcription Factor Binding Motifs}
\label{sec:TF-motifs}

TFs regulate gene expression by binding to specific DNA sequences in the genome. These sequences, known as TF binding motifs, are typically short ($6$-$15$bp) and degenerate, allowing variation at some positions while retaining overall binding specificity \cite{inukai2017transcription}. Motifs are often enriched at cis-regulatory elements such as promoters and enhancers, where TFs bind to modulate transcriptional activity \cite{spitz2012transcription}. Identifying the presence of TF binding motifs in regulatory regions provides a sequence-level signature of potential TF-DNA interactions and enables genome-wide scans for candidate binding sites \cite{grant2011fimo}. 

Motifs are most commonly represented as position weight matrices (PWMs), which describe the base preference of a TF at each position in the motif \cite{stormo2000dna}. A PWM is derived by aligning experimentally determined binding sites and computing the frequency of each nucleotide at each position, often normalized by background base frequencies. These matrices can then be used to scan DNA sequences and assign binding scores, indicating the likelihood that a given TF binds at a specific genomic locus \cite{grant2011fimo}.

Several large-scale databases provide curated collections of TF motifs derived from diverse experimental and computational sources. These databases, such as JASPAR \cite{sandelin2004jaspar, rauluseviciute2024jaspar}, TRANSFAC \cite{wingender1996transfac}, HOCOMOCO \cite{kulakovskiy2013hocomoco, vorontsov2024hocomoco}, and CisBP \cite{weirauch2014determination}, serve as key resources for annotating potential TF binding sites in regulatory regions (Figure \ref{fig:grn-data}F). Annotations can be used to interpret chromatin accessibility and ChIP-seq peaks. While motifs alone do not confirm in vivo binding, they provide valuable sequence-level hypotheses about where and how transcriptional regulation may occur.

Taken together, these data modalities offer complementary views of the regulatory landscape. Transcriptomic technologies such as microarrays, bulk RNA-seq, and single-cell RNA-seq measure the downstream effects of gene regulation, capturing expression patterns that reflect underlying transcriptional programs. In contrast, chromatin-based assays like ATAC-seq and ChIP-seq provide more direct information about regulatory potential and TF binding activity, offering insight into the mechanisms that shape gene expression. Sequence-derived features, including TF motifs from databases like JASPAR, TRANSFAC, HOCOMOCO, and CisBP, capture the intrinsic specificity of TF-DNA interactions encoded in the genome itself. While each modality has unique strengths and limitations, they all serve as valuable sources of evidence for reconstructing GRNs. The next chapter describes computational methods that leverage these data types, both individually or in combination, to infer regulatory relationships and model the structure of transcriptional control.

\section{Machine Learning Foundations for Gene Regulatory Network Inference}
\label{sec:ml-foundations-grn-inference}
Machine learning provides a unifying framework for formalizing GRN inference as a problem of recovering latent regulatory structure from observed genomic data and for comparing the diverse algorithms used in practice. This section first defines the GRN inference task and describes the core data and modeling challenges that make it ill-posed in realistic settings. We then review major families of GRN inference methods, spanning unsupervised, supervised, and prior-informed approaches, and introduce the statistical foundations that underpin the methods developed in Chapters \ref{sec:PMF-GRN-Chapter} and \ref{sec:GLM-Prior-Chapter}. Finally, we summarize evaluation metrics and uncertainty-based diagnostics used to benchmark inferred networks against available reference gold standards.

\subsection{Problem Definition}
GRN inference aims to reconstruct the regulatory relationships that govern transcriptional control within biological systems. Formally, a GRN is represented as a directed acyclic graph in which nodes correspond to genes (or gene products such as TFs) and directed edges represent regulatory influences from TFs to their target genes. An edge from TF $i$ to gene $j$ indicates that the activity of TF $i$ modulates the transcriptional output of gene $j$, either through activation or repression.

\begin{figure}[!htbp]
    \centering
    \includegraphics[width=0.5\textwidth]{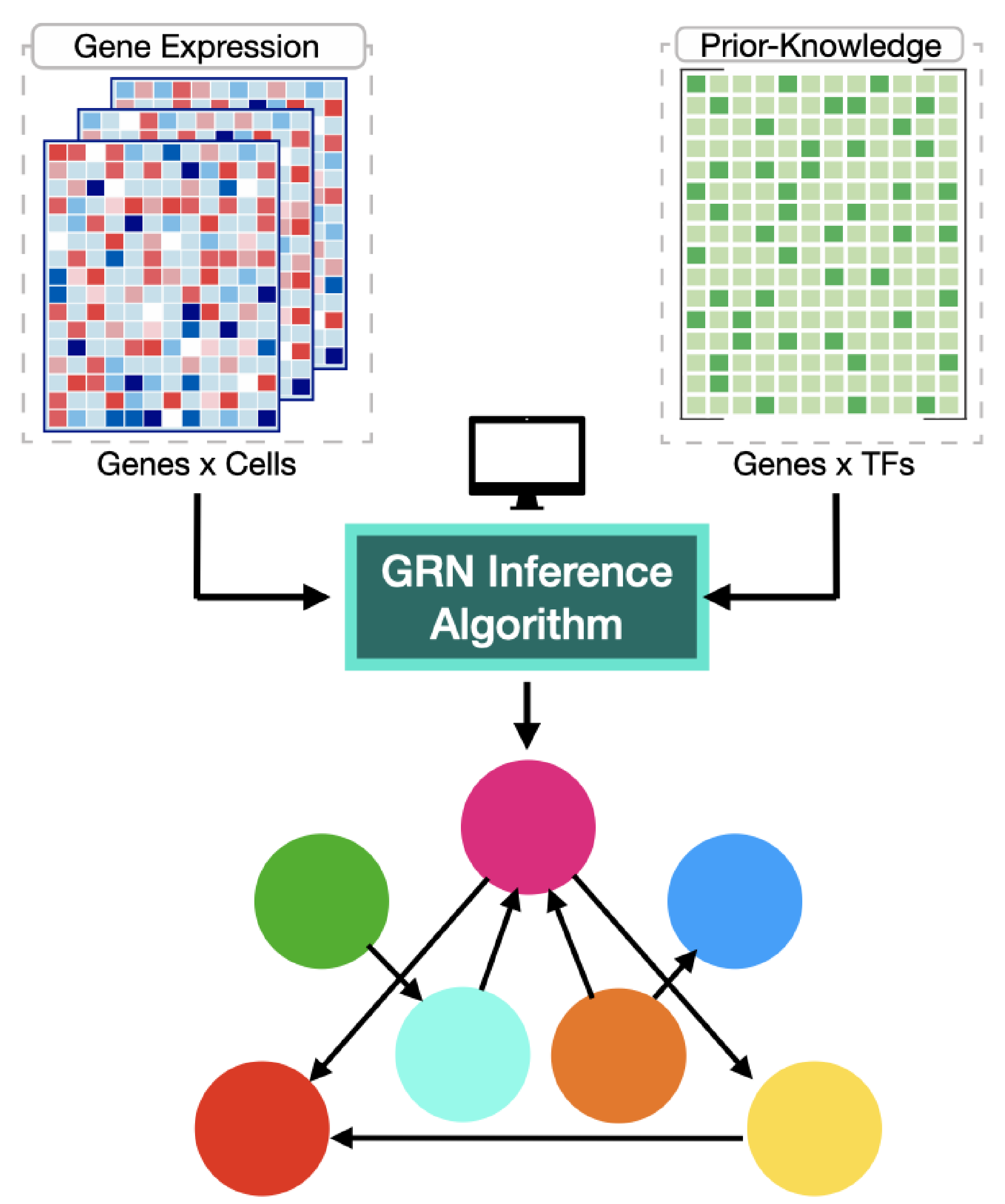} 
    \caption{Schematic overview of the input and output of GRN inference algorithms. The primary input is a gene expression matrix, and optional secondary input is a prior knowledge matrix. The selected GRN inference algorithm will produce a directed acyclic graph describing predicted releationships between TFs and their target genes.}
    \label{fig:grn-input-schematic}
\end{figure}

While a GRN describes direct regulatory interactions between TFs and their target genes, such relationships are not directly measurable by existing genomic assays. Instead, GRN inference algorithms operate on measurable genomic data that reflect the consequences or correlates of regulation. The primary input to most GRN inference methods is a gene expression matrix $W \in \mathbb{R}^{N \times M}$, where $N$ denotes the number of samples, conditions, or cells, and $M$ denotes the number of genes. Each entry of $W$ captures the observed expression level of a gene under a particular condition. In some settings, inference algorithms also incorporate prior knowledge in the form of an auxiliary matrix $A \in \mathbb{R}^{K \times M}$, where $K$ corresponds to TFs and entries encode prior evidence that a TF may regulate a given gene. The output of GRN inference is an estimated regulatory network, typically described as a weighted, directed adjacency matrix (Figure \ref{fig:grn-input-schematic}).

The central modeling goal of GRN inference is to estimate latent regulatory relationships that can best explain the observed data. Given expression measurements $W$, and optionally prior information $A$, algorithms aim to infer which TFs regulate which genes, how strongly, and under what conditions. Crucially, TF-target gene relationships are latent variables; expression data alone provides indirect evidence of regulation, and chromatin-based assays capture regulatory potential rather than functional regulatory effects. As a result, GRN inference is a fundamentally ill-posed problem that requires statistical assumptions, regularization, or inductive biases to constrain the space of plausible solutions.

Different data modalities contribute complementary information to this inference task. Expres-sion-based measurements, including microarrays (Section \ref{sec:microarray}), bulk RNA-seq (Section \ref{sec:RNA-seq}), and single-cell RNA-seq (Section \ref{sec:scrna-seq}), reflect the downstream outcomes of regulatory activity and provide the primary signal used to infer regulation and regulatory influence. In contrast, chromatin accessibility data such as ATAC-seq (Section \ref{sec:atac-seq}), TF binding assays such as ChIP-seq (Section \ref{sec:chip-seq}), and sequence-derived features such as TF binding motifs (Section \ref{sec:TF-motifs}), encode upstream regulatory potential and specificity, and are often used to construct prior knowledge or structural constraints on the inferred network. Machine learning methods for GRN inference differ not only in how they integrate these diverse data types, but also in the way they formulate the inference problem. Approaches can range from unsupervised co-expression analysis to probabilistic generative models and supervised classification frameworks. The following sections aim to describe these modeling approaches and the statistical foundations that underpin them.

\subsection{Challenges in Gene Regulatory Network Inference}
Despite decades of research, GRN inference remains a fundamentally challenging problem, from both a data and a modeling perspective. High-throughput genomic technologies offer only indirect, incomplete views of the underlying regulatory architecture, and the complexity of transcriptional regulation itself poses deep statistical and computational challenges. This section outlines five core challenges that motivate the development of advanced machine learning approaches for inferring GRNs.

\subsubsection{Noisy, High-Dimensional, and Sparse Expression Measurements}
Gene expression matrices are challenging inputs for GRN inference, particularly scRNA-seq, where measurements are noisy, high-dimensional, and sparse. Technical variation, which includes capture efficiency, sequencing depth, and amplification bias, introduces substantial cell-to-cell measurement noise that can obscure regulatory signal \cite{chu2022comprehensive}. In addition, scRNA-seq count matrices contain many zeros due to a combination of true biological non-expression and technical dropout, which complicates the distinction between absence of transcription and failure to detect transcripts \cite{xu2022evaluating}. Although bulk RNA-seq reduces sparsity by averaging across many cells, this averaging also obscures cell-to-cell variability that is often essential for identifying regulatory relationships \cite{li2019dropout}. Together, these properties make the statistical problem of recovering accurate regulatory structure from expression matrices ill-conditioned. The dimensionality is large, the effective signal-to-noise ratio is low, and the data is dominated by zeros and sampling variability.

\subsubsection{Snapshot Sampling and Temporal Mismatch}
An additional limitation of expression data is that transcriptomic measurements provide a snapshot of RNA abundance at the moment of capture, rather than a direct record of regulatory events over time \cite{marr2016single}. Regulatory programs frequently act upstream of the observed mRNA state, and their effects can be observed with delays that break simple contemporaneous associations between regulator TFs and target genes \cite{qiu2020inferring}. In particular, TF activity can be transient and is often shaped by processes that are not reflected by TF mRNA abundance, including protein accumulation, nuclear localization, post-translational modifications, and regulated degradation \cite{brent2016past}. Post-transcriptional mechanisms such as mRNA stabilization and degradation further decouple mRNA levels from the timing and magnitude of upstream regulatory inputs. 

\begin{figure}[!htbp]
    \centering
    \includegraphics[width=0.7\textwidth]{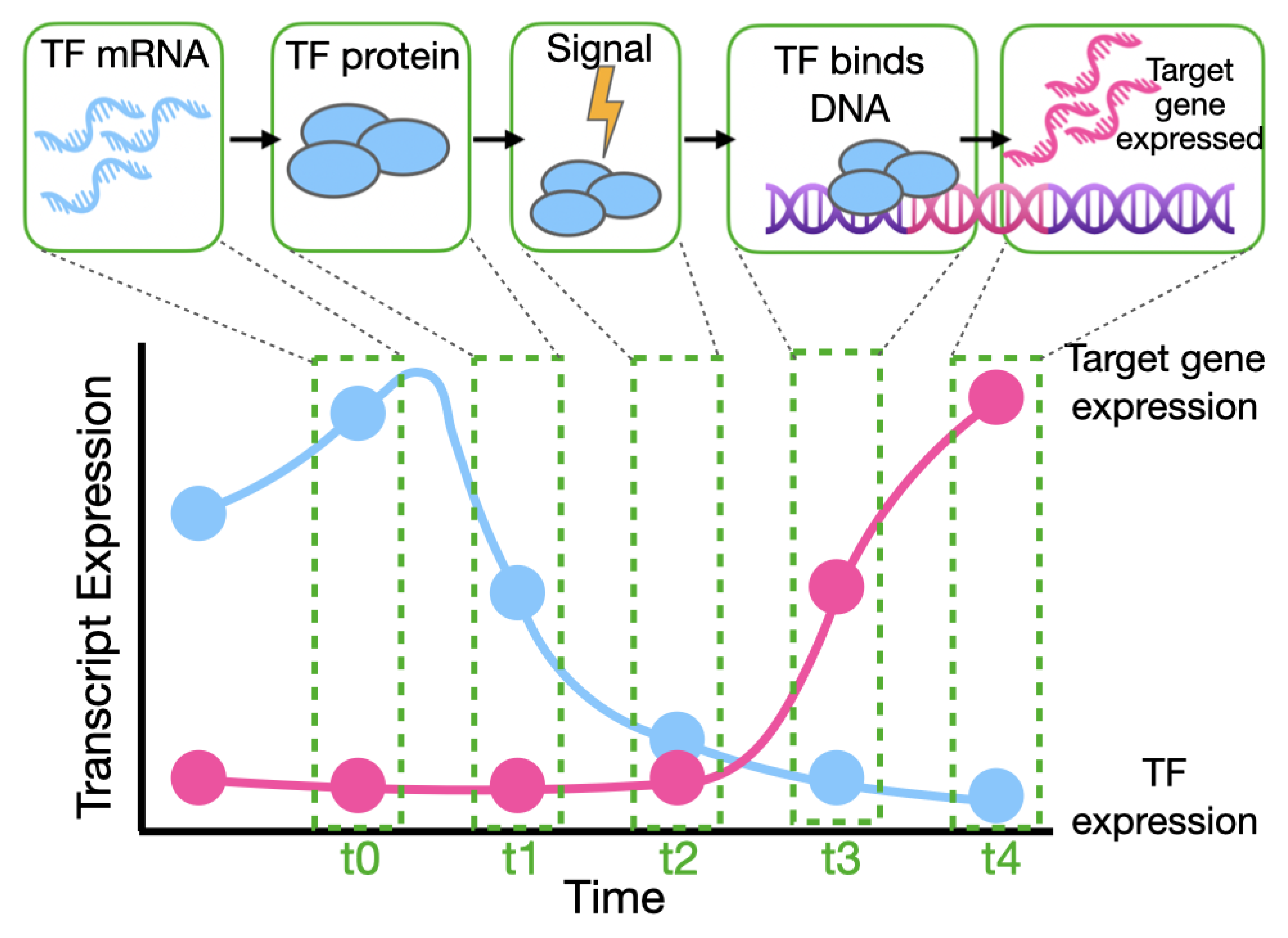} 
    \caption{Snapshot sampling and temporal mismatch between TF transcripts and target-gene expression. Schematic time course illustrating how TF mRNA abundance can be offset in time from the regulatory events that drive target gene transcription. At $t_0$, TF mRNA is highly expressed and then declines over subsequent time points, while TF protein accumulates and becomes positioned for regulation. Following an activating signal at $t_2$, the TF binds DNA at $t_3$, and the downstream target gene exhibits increased transcript abundance only later at $t_4$.}
    \label{fig:snapshot-data}
\end{figure}

As illustrated schematically in Figure \ref{fig:snapshot-data}, profiling expression at different moments can capture qualitatively different snapshots along this regulatory cascade (for example, during TF transcription, after protein accumulation, during activation and DNA binding, or only after the target gene has been expressed), yet the expression matrix alone does not reveal which snapshot of the regulatory landscape has been captured \cite{sha2024reconstructing}. This temporal ambiguity is especially problematic for causal interpretation, where observed correlations can reflect delayed responses, indirect pathways, or regulatory events that occurred prior to measurement, rather than direct TF-target regulation.

\subsubsection{Limited Observability of Regulatory Mechanisms}
Even when expression measurements are high quality, the underlying regulatory mechanisms that generate these profiles are only partially observable. Expression data provides indirect evidence of transcriptional outcomes, but it does not specify which regulators were active, which genes they targeted, or whether the net effect was activating or repressive. Additional assays can constrain the search space of plausible interactions \cite{kim2023gene}, but they remain incomplete and context-dependent. For example, chromatin accessibility determined by ATAC-seq, and TF binding profiles defined by ChIP-seq identify regulatory potential and occupancy, but binding and accessibility alone do not establish functional relevance or quantify regulatory effect sizes \cite{slattery2014absence}.

Moreover, gene regulation depends on molecular processes beyond TF-DNA binding, including cofactor recruitment \cite{inge2024rapid}, chromatin remodeling \cite{aoyagi2008dynamics}, enhancer-promoter communication through 3D genome organization \cite{chaumeil2012role}, and other mechanisms that are not directly captured by standard assays. These layers create nonlinear and context-dependent relationships between regulatory inputs and transcriptional outputs. The same binding event can have different consequences across cell types, developmental stages, or signaling conditions. Limited observability introduces an irreducible ambiguity in GRN inference, where measured data often lack the resolution and completeness needed to uniquely reconstruct regulatory interactions. This challenge motivates models that treat regulation as a latent process and methods that integrate multiple modalities to reduce uncertainty.

\begin{figure}[!htbp]
    \centering
    \includegraphics[width=1.0\textwidth]{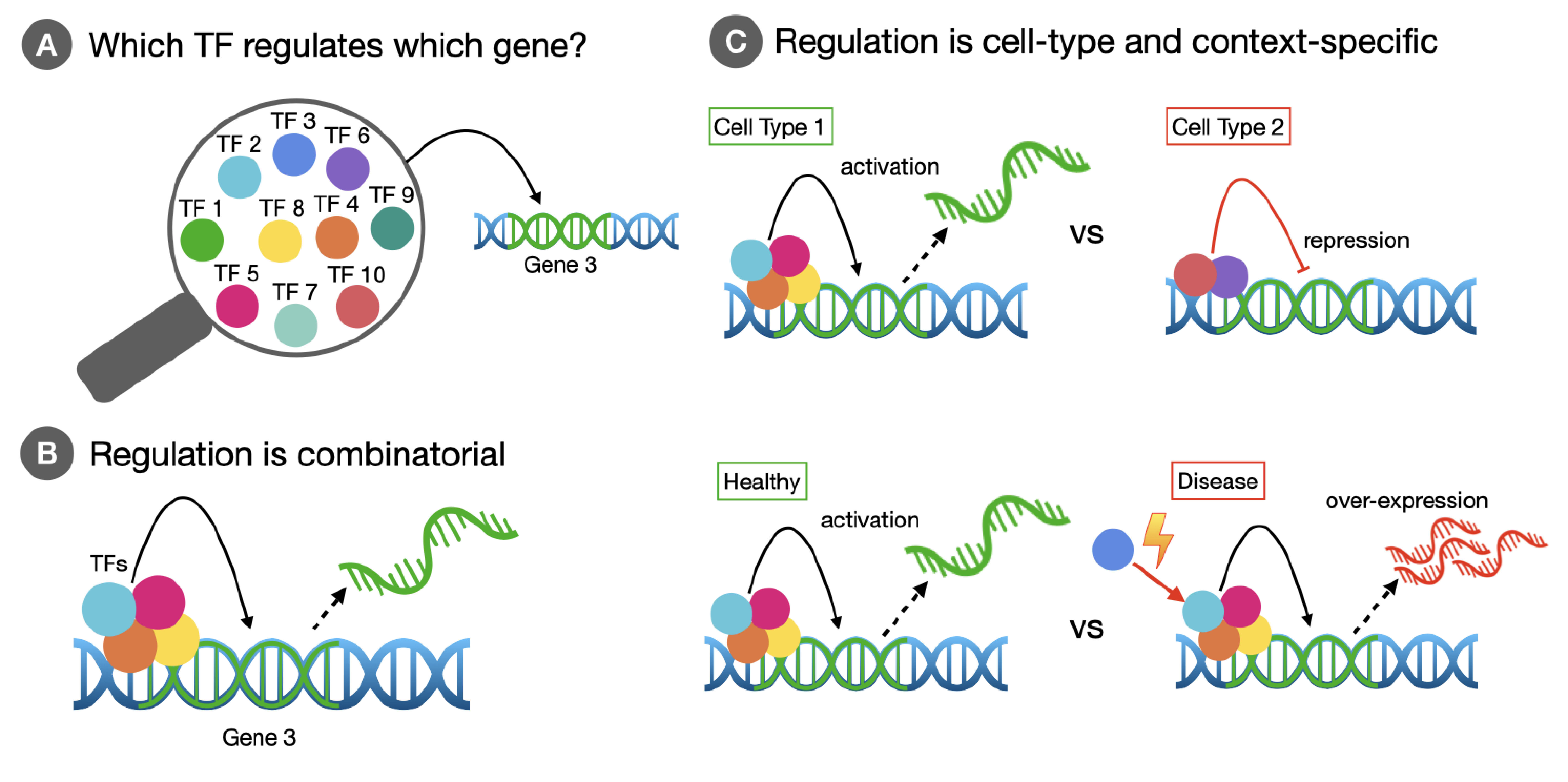} 
    \caption{Key challenges in gene regulatory network inference. \textbf{(A)} A central challenge in GRN reconstruction is determining which transcription factors (TFs) regulate which target genes. \textbf{(B)} Regulation is often combinatorial, with multiple TFs acting cooperatively or competitively to control the same gene, while individual TFs may regulate many targets. \textbf{(C)} Regulatory interactions are also cell-type and context-specific, such that the active GRN can vary across cellular states or conditions.}
    \label{fig:grn-challenges}
\end{figure}

\subsubsection{Combinatorial and Context-Specific Regulation}
Transcriptional regulation is inherently combinatorial (Figure \ref{fig:grn-challenges}). For example, multiple TFs can cooperatively or competitively regulate a single gene, and each TF may regulate many genes in different contexts \cite{dubois2018organizing}. This many-to-many structure leads to regulatory networks that are sparse but structured, with modular or hierarchical organization \cite{ravasi2010atlas}. Capturing this complex structure requires models that can represent conditional dependencies and interactions between regulators, rather than treating regulatory effects as independent or additive. 

Complexity increases in dynamic or heterogeneous systems, where GRNs may shift over time or between cell states. In these settings, a single regulatory interaction may be active only in a specific cell type, condition, or developmental stage. When expression data is aggregated across diverse states, such context-specific patterns can be obscured, and static prior-knowledge networks may also fail to reflect the regulatory program operating in the sampled cells. Methods that infer GRNs must therefore balance generalization with specificity, and account for context-dependent regulatory when possible.

\subsubsection{Incomplete Ground Truth for Evaluation}
Evaluating the accuracy of predicted GRNs remains a major obstacle caused by the scarcity of reliable gold standard annotations. While experimentally validated TF-target gene relationships exist for select interactions in well-studied organisms like yeast or human, these datasets are incomplete, context-specific, and often biased toward particular pathways or cell types \cite{karamveer2024approaches}. As a result, evaluation can only assess a small subset of the predicted network.

Limited access to ground truth TF-target gene interactions complicates not only supervised model training, but also objective benchmarking and performance comparisons across inference methods. Without comprehensive labels, most studies resort to proxy metrics such as overlap with ChIP-seq peaks or motif enrichment near predicted targets \cite{kernfeld2024transcriptome}. Alternatively, performance can be computed on a held-out set of experimentally validated interactions, assessing how well the model recovers known regulatory edges. While informative, these metrics are indirect and can reflect signal unrelated to the true regulatory activity. In cross-organism or cross-cell-type settings, the challenge is further confounded by the lack of transferable evaluation benchmarks. 

Some approaches attempt to address this gap by generating synthetic networks or simulating expression data \cite{pratapa2020benchmarking}, but these strategies risk introducing unrealistic assumptions or artifacts that do not reflect biological complexity. The lack of robust, standardized, and generalizable evaluation frameworks remains a fundamental barrier to assessing GRN inference quality.

\subsubsection{Computational Complexity and Scalability}
GRN inference typically involves evaluating regulatory relationships between thousands of TFs and tens of thousands of genes, leading to a combinatorial number of possible interactions. Even simple modeling approaches that estimate pairwise associations (e.g., correlation or mutual information) must process millions of TF-gene pairs, while more complex methods that fit structured models (e.g., Bayesian networks, probabilistic matrix factorization, or graphical models) face high computational costs \cite{banf2017computational}.

These challenges are compounded in single-cell datasets, where the number of observations (cells) regularly exceed $100,000$, and algorithms must scale to high-dimensional input matrices with efficient memory and runtime performance \cite{skok2022high}. Regularization techniques, dimensionality reduction, or batching strategies are often used to address this, but scalability remains a concern, particularly for iterative inference frameworks or models that incorporate multiple data modalities. Advances in approximate inference, sparse modeling, and distributed high performance computing have helped mitigate these constraints, enabling more expressive models to be applied at scale.

\subsection{Gene Regulatory Network Inference Algorithms}

GRN inference algorithms aim to reconstruct regulatory relationships between TFs and their target genes using high-throughput  experimental datasets, such as gene expression data. Direct experimental validation for regulatory edges is expensive and incomplete, requiring computational approaches to generate candidate networks in order to study how regulatory programs vary across conditions, cell types and time. Existing GRN inference methods span a broad spectrum of assumptions and data requirements, ranging from approaches that rely only on statistical structure in expression matrices to models that learn from curated regulatory interactions, or incorporate prior biological knowledge as constraints \cite{hegde2025machine}. In the following section, we first review unsupervised methods that infer networks directly from expression data, and then contrast them with supervised and semi-supervised approaches that use labeled regulatory edges to train predictive models.

\subsubsection{Unsupervised Methods}
Unsupervised methods for GRN inference aim to reconstruct network structure directly from gene expression data, without access to labeled regulatory interactions or experimentally validated edges. These methods rely on the statistical structure of the expression matrix itself and often assume that regulatory relationships can be uncovered through co-variation, conditional dependence, or sparse predictive modeling. Unsupervised approaches were first developed in the context of microarray experiments, where large numbers of samples across diverse conditions made statistical inference feasible. As high-throughput sequencing technologies evolved, many of these methods were adapted to bulk RNA-seq, and more recently single-cell RNA-seq, often with modifications to account for data sparsity and the lack of replicates.

\paragraph{Information-Theoretic Approaches}\mbox{}\\
One of the earliest families of approaches were co-expression or association networks, which infer putative interactions based on pairwise relationships between gene expression profiles \cite{madhamshettiwar2012gene}. Relevance networks constructed using Pearson or Spearman correlation were among the first to be applied, providing a simple yet powerful mechanism to identify gene pairs with correlated activity. These networks are typically undirected and best suited to capturing gene modules rather than causal regulatory interactions. 

Mutual information (MI) was later introduced to capture nonlinear dependencies, leading to MI-based relevance networks \cite{butte1999mutual} and more advanced formulations like Weighted Gene Co-expression Network Analysis (WGCNA) \cite{zhang2005general}, which groups genes into co-expression modules with potential shared regulation. These methods are broadly applicable to microarray and bulk RNA-seq data, and have been adapted to scRNA-seq \cite{morabito2023hdwgcna} by averaging cells into pseudobulk profiles or applying data-smoothing techniques.

A refinement of co-expression networks came with the development of information-theoretic approaches that attempt to distinguish direct from indirect interactions by incorporating statistical corrections and pruning strategies. One of the most influential of these is ARACNe \cite{margolin2006aracne}, which applies the Data Processing Inequality (DPI) to remove edges that can be explained by a shared intermediate regulatory, enriching for direct regulatory interactions in mutual information (MI) networks. Building on this idea, the Context Likelihood of Relatedness (CLR) algorithm extends MI by normalizing each gene-gene MI value against the empirical distribution of MI scores for each gene, yielding a context-aware z-score matrix that emphasizes unusually strong associations \cite{faith2007large, zhu2016algorithms}. 

In contrast, Minimum Redundancy Network (MRNET) adopts a feature selection framework, treating each gene as a response variable and iteratively selecting regulators based on both their MI with the target and their redundancy with already selected features \cite{meyer2007information}. Alternatively, Conservative Causal Core Network (C3NET) takes a highly selective approach by retaining only the single most significant MI edge for each gene, thereby prioritizing high-confidence associations over network completeness \cite{altay2010inferring}. Finally, Partial Information Decomposition and Context (PIDC), which was specifically designed for scRNA-seq data, decomposes MI into unique, redundant, and synergistic components to account for the complex dependencies and sparsity inherent in scRNA-seq \cite{chan2017gene}. These methods collectively offer improved specificity over simple MI networks, particularly in high-sample regimes, with PIDC retaining relevance for its adaptation to single-cell data.

\paragraph{Sparse Regression Approaches}\mbox{}\\
Another class of unsupervised methods formulates GRN inference as a sparse regression problem, treating gene expression as a predictive modeling task. In this framework, the expression of each gene is modeled as a response variable, while the expression of candidate regulators, typically TFs, serve as the predictor set. One of the earliest and most influential methods in this category is the Inferelator \cite{greenfield2013robust}, which employs regularized regression to identify a parsimonious set of regulators per gene, often incorporating temporal or experimental design information to improve interpretability. This strategy has since been extended through the use of sparsity-inducing penalties such as LASSO or Elastic Net, enabling robust inference in high-dimensional settings \cite{skok2022high, miraldi2019leveraging}. 

Methods like TIGRESS \cite{haury2012tigress} further stabilize inference by coupling regression with bootstrapping and stability selection, providing more reproducible edge estimates across data perturbations. Building on the same sparse modeling principle, ensemble-based methods such as GENIE3 \cite{huynh2010inferring} and GRNBoost2 \cite{moerman2019grnboost2} recast the problem as one of feature importance, using random forests or gradient-boosted trees to estimate the contribution of each TF to the expression of target genes. These approaches have gained popularity for single-cell applications due to their scalability, robustness to sparsity, and compatibility with datasets lacking explicit temporal or experimental labels.

\paragraph{Conditional Dependence and Graphical Models}\mbox{}\\
Beyond direct association or regression, a separate set of methods aims to reconstruct regulatory relationships by estimating conditional dependencies among genes. These approaches are rooted in the principle that true regulatory interactions are best revealed by controlling for indirect effects, typically through partial correlation or graphical modeling techniques. Gaussian graphical models are a widely used formulation, where the inverse covariance (precision) matrix encodes the conditional dependence structure of the system. Estimation procedures such as the Graphical Lasso \cite{friedman2008sparse} improve sparsity constraints to recover interpretable and data-efficient networks. Complementary strategies based on shrinkage, as implemented in methods like GeneNet \cite{schafer2006reverse}, provide stable estimates in high-dimensional but small-sample regimes. While these models are well suited to bulk expression data with large numbers of samples, their reliance on Gaussian assumptions and sensitivity to noise can limit their utility in single-cell contexts, unless mitigated by aggregation, denoising, or imputation techniques. 

\paragraph{Bayesian Networks}\mbox{}\\
Bayesian network learning introduces a probabilistic perspective to GRN inference by modeling the joint distribution over gene expression and identifying directed edges that represent conditional dependencies \cite{friedman2000using}. These methods offer a principled framework for capturing regulatory directionality and uncertainty, with the added advantage of allowing incorporation of prior knowledge into the learning process \cite{werhli2007reconstructing}. However, Bayesian networks are computationally intensive and typically require discretizations of continuous expression values, which can obscure subtle expression changes. Dynamic Bayesian networks extend this approach to time-series data, modeling regulatory programs as evolving over discrete temporal intervals. These have proven particularly useful in developmental studies or perturbation experiments where time-course data is available and temporal ordering among genes is biologically meaningful \cite{husmeier2003sensitivity}.

\paragraph{Trajectory-Aware Methods}\mbox{}\\
The advent of single-cell transcriptomics has spurred the development of trajectory-aware GRN inference methods that exploit the pseudo-temporal structure of differentiating or transitioning cells. These methods assume that cells can be ordered along a latent temporal axis and seek to uncover regulators that drive expression changes across this inferred trajectory. Methods such as SCODE \cite{matsumoto2017scode} model gene expression dynamics using linear ordinary differential equations, positing that observed expression levels arise from regulatory activity along pseudotime. GRISLI \cite{aubin2020gene} adopts a similar framework but incorporates sparse regression to infer dynamic, time-varying interactions. Methods like SINCERITIES \cite{papili2018sincerities} leverage timestamped single-cell measurements, while LEAP \cite{specht2017leap} uses lagged correlation to infer putative regulatory delays between genes along the trajectory. These approaches are specifically tailored to the challenges and opportunities of single-cell data, including sparsity, asynchronous sampling, and nonlinear expression dynamics, and they represent a growing frontier in unsupervised GRN modeling. 

\paragraph{Summary}\mbox{}\\
Together, these methods highlight the breadth of statistical strategies available for unsupervised GRN inference using expression data alone. From simple pairwise correlations to sophisticated probabilistic models, each approach balances different trade-offs in scalability, interpretability, and robustness to noise. While unsupervised methods can reveal informative patterns in high-throughput data, they ultimately operate on the assumption that statistical dependencies reflect underlying regulatory mechanisms. In the absence of experimental validation or orthogonal priors, these networks should be interpreted with caution, particularly when used to derive biological hypotheses or inform downstream analyses.

\subsubsection{Supervised and Semi-Supervised Methods}
Supervised learning approaches for GRN inference leverage labeled examples of known regulatory interactions to train predictive models that generalize to unseen TF-gene pairs. In contrast to unsupervised methods, which infer regulatory relationships solely from expression patterns, supervised methods explicitly rely on curated gold standards. Examples of these curated gold standards, which guide the learning process, consist of interaction labels derived from ChIP-based experiments, genetic perturbations, or high-confidence database annotations. When sufficient labeled regulatory edges are available, supervised models can more accurately distinguish true regulatory interactions from spurious associations by learning discriminative patterns that differentiate positive from negative examples. 

Within the space of supervised methods, SIRENE \cite{mordelet2008sirene} reformulates GRN inference as a series of local binary classification problems. For each TF, SIRENE trains a support vector machine (SVM) using expression-based features and known TF-target labels, with the assumption that co-regulated targets exhibit similar expression profiles. Another supervised approach that extends traditional machine learning to temporal expression profiles is dynGENIE3 \cite{huynh2018dyngenie3}, which builds upon the random forest framework of GENIE3 \cite{huynh2010inferring} by incorporating time-series information. Although GENIE3 is typically categorized as unsupervised learning, dynGENIE3 adds supervision through semi-parametric modeling of dynamic regulatory effects, integrating temporal gradients with ensemble regression to better capture time-dependent regulatory influences.

In recent years, deep learning has been applied extensively to supervised GRN inference, particularly in contexts where sequence and chromatin features complement expression data. DeepIMAGER \cite{zhou2024deepimager} uses a convolutional neural network (CNN) architecture to convert co-expression patterns and TF binding features into an image-like format for classification, training on labeled regulatory pairs derived from scRNA-seq and ChIP-seq data. Similarly, DeepSEM \cite{shu2021modeling} employs a deep structural equation model that jointly learns latent regulatory influences and gene expression dependencies, using supervision from curated TF-target gene interactions to refine its parameter estimates. 

Transformer-based models, which leverage self-attention mechanisms to capture long-range dependencies, have emerged as powerful supervised learners for GRN tasks. STGRNs \cite{xu2023stgrns} applies a transformer encoder to gene expression and positional features, learning representations that encapsulate regulatory context before classification. Parallel to this, GRNFormer \cite{hegde2025grnfomer} leverages a variational graph transformer autoencoder to model both global and local regulatory patterns in single-cell data. In GRNFormer, subgraphs centered on TFs are embedded using attention layers, and a variational decoder estimates a probabilistic adjacency matrix, trained with ground-truth regulatory data using a composite loss that balances prediction accuracy and latent regularization.

Another supervised method is RSNET \cite{jiang2022rsnet}, which combines mutual information-based filtering with a sparse linear regression model. RSNET uses MI to identify candidate TF-gene regulatory pairs, then applies a redundancy-silencing optimization that penalizes redundant regulators and enhances high-confidence edges. This approach improves GRN inference by prioritizing direct regulatory interactions and reducing spurious co-expression effects.

Finally, alternative supervised strategies explore different problem formulations. AnomalGRN \cite{zhou2025anomalgrn} reframes GRN inference as a graph anomaly detection task, in which interacting gene pairs are treated as rare or abnormal nodes, and non-interacting pairs as normal nodes. The method reconstructs a graph over gene-pair nodes using expression-derived features, then applies graph neural networks with sparsification to identify meaningful regulatory interactions in the presence of extreme class imbalance and noisy connectivity. This perspective is particularly well-suited to single-cell data, where dropout, sparsity, and the scarcity of positive regulatory links complicate standard supervised classification.

Collectively, these supervised and semi-supervised methods illustrate the evolution of GRN inference from early binary classifiers to modern frameworks based on deep learning, graph neural networks, and alternative formulations such as anomaly detection. These methods demonstrate how supervised learning can integrate heterogeneous biological features, including expression, sequence, and chromatin context, while learning regulatory patterns from available labeled data. However, because such labels remain sparse outside of well-characterized organisms and experimental settings, the performance of these approaches is ultimately constrained by the scale and quality of known interactions. In the next section, we examine hybrid frameworks that incorporate prior knowledge directly into statistical models of expression, aiming to combine the fidelity of supervised learning with the scalability of unsupervised inference.

\subsubsection{Prior-Informed Methods}
\begin{figure}[!htbp]
    \centering
    \includegraphics[width=1.0\textwidth]{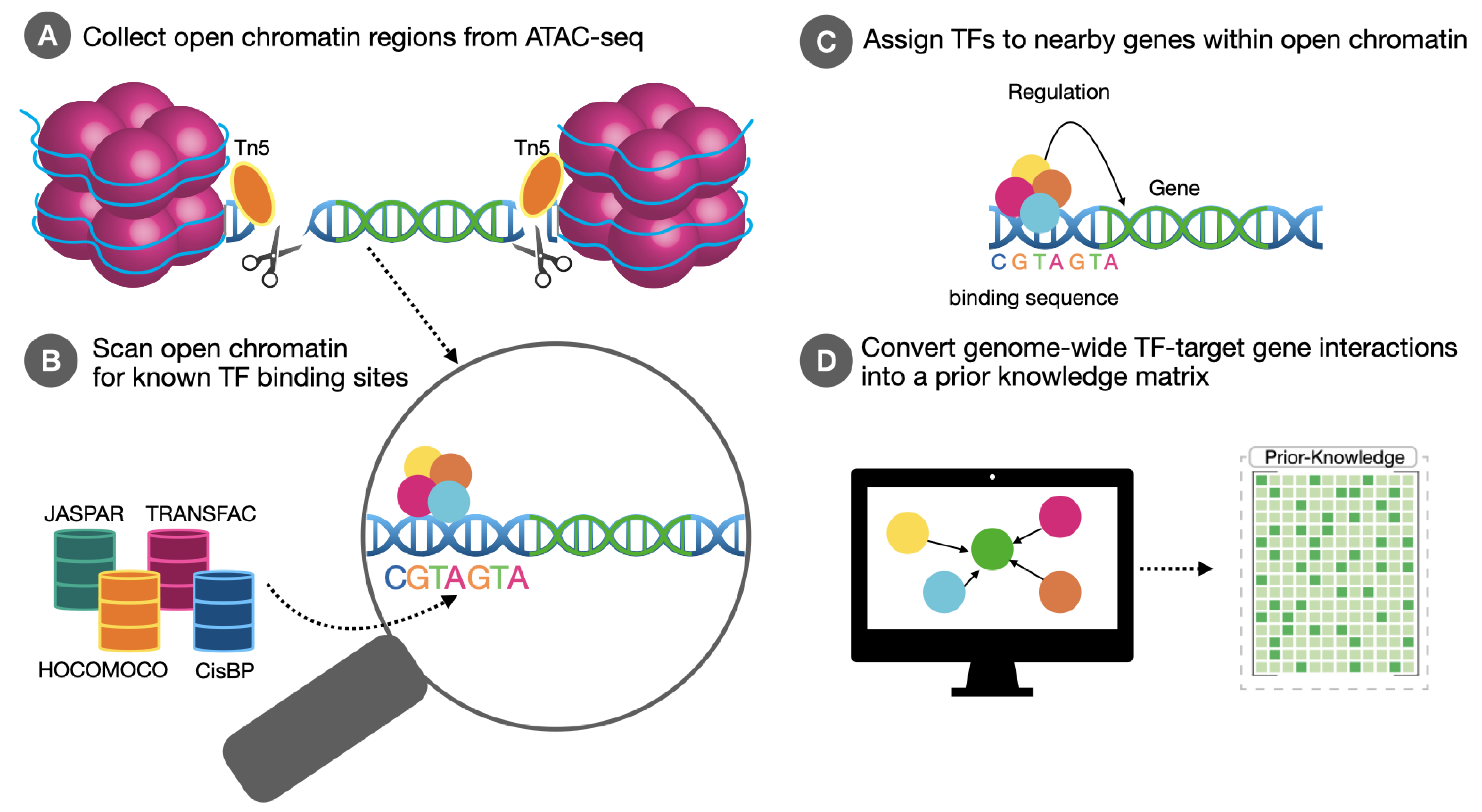} 
    \caption{Typical pipeline for constructing prior-knowledge matrix for downstream GRN inference. \textbf{(A)} Open chromatin regions are identified from ATAC-seq data to define candidate regulatory DNA accessible to transcription factor (TF) binding. \textbf{(B)} These accessible regions are scanned for known TF binding motifs to identify potential TF occupancy. \textbf{(C)} Motif-containing accessible regions are assigned to nearby genes, yielding candidate TF-target regulatory interactions. \textbf{(D)} Genome-wide interactions are aggregated into a prior-knowledge matrix, in which rows correspond to target genes, columns correspond to TFs, and entries indicate putative regulatory relationships. This prior matrix can then be incorporated into GRN inference methods to constrain the search space toward biologically plausible edges.}
    \label{fig:prior-knowledge}
\end{figure}

An important class of GRN inference methods enhances expression-based modeling by incorporating prior biological knowledge. These priors, typically constructed from sequence motifs, chromatin accessibility, TF binding profiles, or curated regulatory databases, serve as structured constraints that guide or regularize the inference process (overview illustrated in Figure \ref{fig:prior-knowledge}). Rather than treating every possible TF-gene pair as a candidate interaction, prior-informed models leverage biological evidence to define a subset of likely regulatory edges. This approach not only reduces the combinatorial search space, but also increases the biological plausibility and interpretability of the inferred networks.

Of these prior knowledge constrained GRN inference algorithms, the Inferelator \cite{skok2022high, greenfield2013robust, arrieta2015experimentally, miraldi2019leveraging} constructs a prior matrix by scanning for TF binding motifs within accessible chromatin regions, typically promoter-proximal ATAC-seq peaks, and integrates this prior with gene expression using sparse regression models. The prior matrix represents binary or weighted confidence scores for potential regulatory edges and serves as a constraint during model fitting, encouraging the selection of edges that are supported by the data and consistent with known biology. 

Similarly, CellOracle \cite{kamimoto2023dissecting} builds motif-informed prior networks from promoter regions that intersect open chromatin. These priors are then integrated with single-cell expression data using linear models or dynamical simulations to estimate TF-gene relationships and predict perturbation outcomes. Additionally, pySCENIC \cite{van2020scalable} provides another prior-constrained GRN inference approach, by incorporating motif-based priors as a post-hoc filtering step in a hybrid inference pipeline. pySCENIC uses GENIE3 to infer co-expression modules from gene expression data. These candidate regulatory modules are then refined by pruning edges lacking motif support in the promoter regions of the predicted target genes. 

By constraining GRN inference to a biologically plausible sub-network, prior-informed methods improve statistical power, particularly in low-sample or high-noise regimes. They also enable transferability across datasets by anchoring inference in shared regulatory logic rather than dataset-specific expression correlations. Importantly, the effectiveness of these methods depends on the quality of the priors themselves. Mis-annotated or overly inclusive priors can bias the model or suppress the discovery of novel edges. Moreover, in organisms or cell types where chromatin or TF binding data are limited, constructing meaningful priors may be challenging, prompting recent efforts to generate priors directly from sequence using learned models, as is discussed in Chapter \ref{sec:GLM-Prior-Chapter}.

\subsection{Machine Learning and Statistical Foundations}
This section introduces the core statistical and machine learning ideas that underpin the GRN inference methods developed in Chapters \ref{sec:PMF-GRN-Chapter} and \ref{sec:GLM-Prior-Chapter}. We begin with latent-variable representations of expression via matrix factorization and use identifiability to motivate why external information is needed to align latent dimensions with named TFs. We then place factorization within a probabilistic generative framework and describe variational inference as a practical approach for approximate posterior inference with quantified uncertainty. Finally, we introduce sequence language models and standard classification objectives as the tools used to construct sequence-derived priors that provide TF-specific structure to constrain downstream GRN inference.

\subsubsection{Latent Variables and Matrix Factorization}
GRN inference is fundamentally a problem of reasoning about unobserved causes from noisy measurements. Gene expression assays provide an observed snapshot of mRNA transcript abundance, however, the regulatory quantities that generate these measurements, such as TF activity and TF-target gene influence, are not directly measured and may not be well-approximated by TF mRNA abundance alone. Latent variable models formalize this distinction by introducing unobserved variables that capture the underlying regulatory state, using them to explain the observed expression matrix. 

Let $W \in \mathbb{R}^{N \times M}$ denote an observed gene expression matrix, where $N$ is the number of samples or cells and $M$ is the number of genes. Matrix factorization assumes that the dominant structure in $W$ can be represented by a smaller number of latent factors $K \ll min(N, M)$, yielding the approximation
\begin{equation}
    W \approx UV^{\top},
\end{equation}
where $U \in \mathbb{R}^{N \times K}$ and $V \in \mathbb{R}^{M \times K}$. The matrix $U$ assigns each sample $n$ a vector of $K$ latent factor values $U_{n,:}$, while the matrix $V$ assigns each gene $m$ a vector of factor loadings $V_{m,:}$. Under this model, the predicted expression of gene $m$ in sample $n$ is a weighted sum of factor contributions,
\begin{equation}
    (W_{pred})_{n,m} = \sum_{k=1}^K U_{n,k} V_{m,k}.
\end{equation}
This decomposition separates sample-specific variation, captured by $U$, from gene-specific response patterns, captured by $V$, providing a compact representation of coordinated expression programs \cite{stein2018enter}.

In the GRN setting, the latent factors are often interpreted in relation to TF-driven regulation. A common choice is to set $K$ to the number of TFs under study and interpret the columns of $U$ as latent TF activities across samples or cells, and the columns of $V$ as gene-specific responses to those activities. With this interpretation, each entry $U_{n,k}$ represents the activity of TF $k$ in sample $n$, and each entry $V_{m,k}$ represents the extent to which TF $k$ is associated with changes in expression of gene $m$. The matrix product $UV^{\top}$ then aggregates the contributions of many TFs to each gene, consistent with combinatorial regulation.

The latent variable framing is useful for two reasons. First, it permits drivers of regulation (TFs) to be inferred even when they are not directly observed, allowing the model to capture regulatory signals that may arise from post-transcriptional processes or upstream inputs not reflected in TF mRNA measurements. Second, it provides a natural route to reconstructing regulatory structure. The gene-by-factor loadings in $V$ can be interpreted as a candidate TF-gene interaction, while $U$ captures the cell- or condition-specific activity state for a TF, determining its contribution to a target gene's observed expression.

Matrix factorization alone, however, does not uniquely determine the latent factors. Multiple pairs $(U,V)$ can yield the same product $UV^{\top}$, and additional constraints or side information are generally required to assign a stable interpretation to individual factors. This issue is particularly important in GRN inference, when factors are intended to correspond to specific TFs. The next subsection addresses this identifiability challenge and motivates the role of prior knowledge in anchoring the latent space to interpretable regulatory components. 

\subsubsection{Identifiability and the Role of Priors}
\label{sec:identifiability-background}
A central limitation of matrix factorization models is that the latent factors are not uniquely determined by the observed data \cite{papastamoulis2022identifiability}. The approximation $W \approx UV^{\top}$ specifies only the product of two matrices, not a unique choice of $U$ and $V$. As a result, multiple distinct latent representations can explain the same expression matrix equally well. This non-uniqueness is referred to as an identifiability issue, and it becomes particularly consequential when the goal is to interpret latent dimensions as specific TFs.

The simplest demonstration of non-identifiability arises from permutations of the latent dimensions. Let $P \in R^{K \times K}$ be any permutation matrix. Then
\begin{equation}
    UV^{\top} = (UP)(VP)^{\top},
\end{equation}
since $(VP)^{\top} = P^{\top}V^{\top}$ and $PP^{\top} = I$. This implies that the ordering of $K$ latent factors is arbitrary; swapping two columns of $U$ and the corresponding columns of $V$ leaves the reconstructed expression unchanged. More simply, without additional information, there is no principled way to determine which latent dimension corresponds to which TF. When $K$ is large, the number of equivalent permutations grows as $K!$, so a factorization aligned with TF identities is unlikely to emerge consistently from expression data alone.

In GRN inference, interpretability requires more than recovering a low-rank approximation to $W$. The latent dimensions are intended to correspond to named TFs, so the inferred regulatory structure must be aligned to those TF identities. Priors are essential for introducing external information that breaks the symmetry of the latent space by preferring solutions consistent with known or hypothesized TF-gene relationships. Conceptually, priors act as anchors biasing the model toward factorization solutions in which particular genes load onto particular TF-associated dimensions, allowing a stable and interpretable mapping between latent factors and TF identities \cite{leung2016order}. 

Within probabilistic formulations of factor models, priors can be placed directly on the gene-by-factor loadings $V$, or on a structured representation of TF-gene interactions that governs $V$. The key requirement is that the prior must encode TF-specific information, instead of generic regularization. For example, shrinkage priors that promote sparsity can improve stability and reduce overfitting. These priors, however, do not themselves resolve which factor corresponds to which TF. In contrast, a prior derived from a combination of TF binding profiles and chromatin accessibility, or curated regulatory interactions, supplies asymmetric information across TFs and genes, and therefore can distinguish one latent dimension from another.

The role of priors is not only a technical fix, but reflects a broader principle in GRN inference. The space of candidate regulatory networks consistent with expression data is extremely large, and many networks can generate similar expression patterns under realistic noise levels and sampling constraints. Prior knowledge narrows this space by restricting attention to biologically plausible edges and by aligning latent factors with TF identities so they can be interpreted. Later sections will instantiate this idea in two complementary ways, the first by using priors to anchor latent-variable inference over regulatory structure (Chapter \ref{sec:PMF-GRN-Chapter}), while the second focuses on constructing higher-quality priors from sequence to better constrain downstream inference (Chapter \ref{sec:GLM-Prior-Chapter}).

\subsubsection{Probabilistic Matrix Factorization and Generative Modeling}
Matrix factorization provides a useful low-dimensional representation of gene expression, but it does not by itself specify how expression measurements are generated, how uncertainty should be quantified, or how prior biological knowledge should be incorporated in a principled way. Probabilistic matrix factorization \cite{mnih2007probabilistic} addresses these limitations by reframing factorization as a generative model. Rather than fitting $U$ and $V$ solely by minimizing a reconstruction objective for $W \approx UV^{\top}$, it specifies a likelihood for the observed matrix $W$ together with priors over latent variables such as cell-specific activities and gene-specific regulatory effects. This perspective makes it possible to encode biological assumptions through prior distributions, to model measurement noise explicitly, and to infer a posterior distribution over unobserved regulatory structure rather than a single point estimate.

As in the previous subsections, we let $W \in \mathbb{R}^{N \times M}$ denote an observed gene expression matrix, with $N$ cells and $M$ genes. In probabilistic matrix factorization, we introduce latent matrices $U \in \mathbb{R}^{N \times K}$ and $V \in \mathbb{R}^{M \times K}$ and define a likelihood for $W$ conditioned on these latent factors. The defining assumption is that the expected expression is given by the matrix product $UV^{\top}$, while the observations deviate from this expectation due to measurement noise. We can formulate this with equation \eqref{eq:pmf-likelihood},
\begin{equation} \label{eq:pmf-likelihood}
    p(W \mid U, V, \sigma_{obs}) = \prod_{n=1}^N \prod_{m=1}^M p(W_{n,m} \mid (UV^{\top})_{n,m}, \sigma_{obs}),
\end{equation}
where $\sigma_{obs}$ is an observation noise parameter and the conditional distribution $p(\cdot)$ is chosen to reflect the expression measurement type. Here, the likelihood specifies an explicit stochastic relationship between the latent regulatory variables and observed expression, separating the biological signal captured by $UV^{\top}$ from expression measurement variability.

The probabilistic formulation is completed by specifying prior distributions over the latent variables,
\begin{equation}
    p(W \mid U, V, \sigma_{obs}) = p(U)p(V)p(\sigma_{obs}),
\end{equation}
where the choice of priors determine both regularization and biological structure. TF-specific priors can incorporate external evidence about which TF-gene interactions are plausible, thereby anchoring the latent dimensions to interpretable TF identities as discussed in Section \ref{sec:identifiability-background}. 

We can formulate the joint distribution over observed and latent variables with equation \eqref{eq:joint-dist},
\begin{equation} \label{eq:joint-dist}
    p(W, U, V, \sigma_{obs}) = p(W \mid U, V, \sigma_{obs})p(U)p(V)p(\sigma_{obs}),
\end{equation}
which defines a complete generative story. Here, we first draw latent variables from their priors, then generate expression measurements from the likelihood. Under this model, GRN inference corresponds to computing the posterior distribution, 
\begin{equation}
    p(U, V, \sigma_{obs} \mid W),
\end{equation}
and using the inferred regulatory component of V to characterize TF-gene relationships. In practice, exact posterior inference is intractable for most choices of priors and likelihoods, as we do not know the true data generating distribution to which $W$ belongs. This motivates the following section (Section \ref{sec:variational-inference-background}), which describes approximate inference and optimization approaches.

Overall, probabilistic matrix factorization provides a general statistical scaffold for GRN inference. It retains the dimensionality-reduction benefits of matrix factorization while enabling principled incorporation of prior knowledge, explicit modeling of noise, and posterior inference over regulatory structure with quantified uncertainty.

\subsubsection{Variational Inference}
\label{sec:variational-inference-background}
The probabilistic matrix factorization model defined in the previous section specifies a generative story for how an observed expression matrix $W$ could arise from latent variables such as $U$, $V$, and $\sigma_{obs}$. Once a generative model is specified, the central computational task becomes posterior inference. Given the observed data $W$, we would like to infer which latent configurations are plausible under the model, i.e., compute $p(U, V, \sigma_{obs} \mid W)$. This posterior is the mathematical object that encodes uncertainty about latent regulatory structure, and it is what enables uncertainty-aware GRN reconstruction rather than relying on a single point estimate.

For notational convenience, we collect all latent variables into a single variable $z$, where $z = U, V, \sigma_{obs}$, so that the joint distribution defined by the generative model can be written completely as $p(W,z)$ and the posterior of interest becomes $p(z \mid W)$.

In most reliable models, however, the posterior cannot be computed exactly. The obstacle is the normalizing constant in Bayes' rule, the marginal likelihood (also referred to as the evidence), $p(W) = \int p(W, z) dz$. This integral couples together all latent variables and is rarely tractable in high dimensions. Variational inference (VI) \cite{blei2017variational} addresses this by replacing exact inference with an optimization problem. Instead of computing $p(z \mid W)$ directly, we choose a simple family of distributions $\mathcal{Q}$ and search within that family for an approximation $q(z)$ that is close to the true posterior.

\paragraph{KL Divergence}\mbox{}\\
In order to estimate a precise approximate posterior, VI typically uses the Kullback-Leibler (KL) divergence as a distance measurement between the true and approximate posterior distributions. The variational objective is 
\begin{equation} \label{eq:variational-objective}
    q^{*}(z) = \arg\min_{q(z)\in\mathcal{Q}} D_{\mathrm{KL}}\!\big(q(z)\,\|\,p(z\mid W)\big),
\end{equation}
where
\begin{equation} \label{eq:variational-objective-full}
    D_{\mathrm{KL}}\!\big(q(z)\,\|\,p(z\mid W)\big) = \mathbb{E}_{z\sim q(z)}\!\left[\log\frac{q(z)}{p(z\mid W)}\right].
\end{equation}
The formulation in \eqref{eq:variational-objective} captures the intuition of VI where we select the best approximation to the posterior from a tractable family $\mathcal{Q}$. The remaining problem is practical: the objective still involves $p(z\mid W)$, which depends on the intractable evidence $p(W)$.

\paragraph{Evidence Lower Bound}\mbox{}\\
The standard derivation of the evidence lower bound (ELBO) follows from rewriting the KL divergence in terms of quantities we can evaluate. The key identity is Bayes' rule, $p(z \mid W) = \frac{p(W, z)}{p(W)}$. Substituting this into the definition of KL divergence \eqref{eq:variational-objective-full} yields 
\begin{align}\label{eq:DKl}
D_{\mathrm{KL}}\!\big(q(z)\,\|\,p(z\mid W)\big)
&=
\mathbb{E}_{z\sim q(z)}\!\left[\log\frac{q(z)}{p(W,z)/p(W)}\right] \nonumber \\
&=
\mathbb{E}_{z\sim q(z)}\!\left[\log\frac{q(z)}{p(W,z)}\right]
\;+\; \log p(W). 
\end{align}
This step involves the main trick, in which we replace the posterior $p(z \mid W)$ with the joint distribution $p(W,z)$ and the evidence $p(W)$. The joint $p(W,z)$ is exactly as defined in the previous section through the generative model (likelihood times priors), so it is accessible. The evidence $\log p(W)$ in \eqref{eq:DKl} is still intractable, but it is also a constant with respect to $q(z)$.

We can now collect all terms that depend on $q$ into a single objective:
\begin{equation}\label{def:ELBO}
    \mathcal{L}(q) \;=\; \mathbb{E}_{z\sim q(z)}\big[\log p(W,z) - \log q(z)\big].
\end{equation}
Using definition \eqref{def:ELBO}, the previous equation \eqref{eq:DKl} becomes the following decomposition
\begin{equation}
    D_{\mathrm{KL}}\!\big(q(z)\,\|\,p(z\mid W)\big) \;=\; -\mathcal{L}(q) + \log p(W).
\end{equation}
This relationship demonstrates how optimizing the ELBO solves the original variational problem \cite{jordan1999introduction}. Since $\log p(W)$ does not depend on $q(z)$, minimizing the KL divergence is equivalent to maximizing $\mathcal{L}(q)$:
\begin{equation}
    \arg\min_{q(z)\in\mathcal{Q}} D_{\mathrm{KL}}\!\big(q(z)\,\|\,p(z\mid W)\big) \;\equiv\; \arg\max_{q(z)\in\mathcal{Q}} \mathcal{L}(q).
\end{equation}
The term, evidence lower bound, follows from one final rearrangement:
\begin{equation}
    \log p(W) \;=\; \mathcal{L}(q) + D_{\mathrm{KL}}\!\big(q(z)\,\|\,p(z\mid W)\big).
\end{equation}
Because the KL divergence is always nonnegative, this implies $\mathcal{L}(q) \le \log p(W)$, so $\mathcal{L}(q)$ is the lower bound on the evidence. Maximizing the ELBO therefore has a clear interpretation. It derives $q(z)$ toward the true posterior while also implicitly increasing the model's ability to explain the observed data under the generative story.

Expanding the joint $p(W,z) = p(W \mid z)p(z)$ yields a more interpretable form of the ELBO:
\begin{equation} \label{eq:ELBO-2}
    \mathcal{L}(q) = \mathbb{E}_{z\sim q(z)}\big[\log p(W\mid z)\big] + \mathbb{E}_{z\sim q(z)}\big[\log p(z)\big] - \mathbb{E}_{z\sim q(z)}\big[\log q(z)\big].
\end{equation}
The first term in \ref{eq:ELBO-2} encourages latent variables that reconstruct the observed expression matrix well under the likelihood (data fit). The second term encourages latent variables that remain plausible under priors, while the final term is the entropy of $q(z)$, which discourages overly confident approximations when the posterior is uncertain.

In the context of probabilistic matrix factorization for GRN inference, VI provides a practical route from a generative model to uncertainty-aware estimates of latent regulatory structure. Rather than attempting to compute $p(U,V,\sigma_{obs}\mid W)$ exactly, we optimize a tractable approximation $q(U,V,\sigma_{obs})$ by maximizing the ELBO. This yields posterior summaries (e.g., means and variances) for regulatory parameters, enabling downstream GRN analyses that explicitly reflect uncertainty in inferred TF activities and TF-target gene relationships.

\subsubsection{Sequence Language Models}

Sequence-based priors naturally complement the latent-variable and probabilistic modeling framework described above by providing TF-specific structure that expression data alone cannot resolve. The earlier subsections highlighted two recurring challenges in GRN inference. First, expression measurements are noisy, partial snapshots of regulatory activity, and second, latent factor models are not identifiable without external information that anchors latent dimensions to specific TFs. Molecular sequence offers a direct source of such anchoring information. TF binding and regulatory specificity are ultimately encoded in sequence, through properties of the TF sequence that shape its DNA-binding preferences, and through cis-regulatory DNA sequence that determines how genes respond to regulatory inputs. Sequence language models build on this premise by learning reusable representations of biological sequences from large corpora, enabling the construction of informative priors even when experimentally validated TF-gene interactions are sparse or incomplete.

A sequence language model is a neural network trained to learn general-purpose representations of biological sequences from large-scale data. In biology, the alphabet consists of nucleotides \texttt{A,T,C,G } for DNA, \texttt{A,C,G,U } for RNA, or amino acids for proteins, and a sequences can be treated analogously to a sentence. The most widely used family of models for this purpose is based on the Transformer \cite{vaswani2017attention, devlin2019bert}, which maps an input sequence to contextual token representations. These representations capture dependencies between positions across the sequence, allowing the model to represent both local patterns, such as motifs, and broader context that can modulate binding and regulatory function. 

Pretraining provides the main advantage of sequence language models in settings where labeled biological data are limited. During pretraining, the model is optimized on large unlabeled sequence collections using a self-supervised objective, most commonly masked language modeling. Under this objective, a subset of tokens is hidden and the model learns to predict them from context, encouraging it to discover recurring sequence regularities. The resulting encoder can convert an input sequence into a latent representation, known as an embedding, that summarizes sequence content in a form useful for downstream tasks. Embeddings can be used in feature-based approaches, where the pretrained encoder is frozen and embeddings serve as inputs to a task-specific model, or in a fine-tuning approach, where the encoder is further trained end-to-end, allowing embedding representations to become specialized for a particular prediction problem.

In GRN inference, a recurring goal is to score candidate TF-gene relationships in a way that can be used as prior evidence. Sequence language models support this by producing TF- and gene-specific representations that can be combined into an interaction score. When repeated over many TF-gene pairs, these scores form a TF-by-gene matrix that can be interpreted as a prior network. This provides an additional mechanism for addressing the identifiability issue described earlier, where this sequence-derived prior can introduce TF-specific asymmetry across genes and therefore can help anchor latent dimensions to named TFs in downstream probabilistic inference. 

The role of sequence-derived priors is best understood as a way to constrain and structure uncertainty, rather than as a replacement for expression-based modeling. A probabilistic factor model operating on $W$ can quantify uncertainty about latent regulatory effects and cell-specific activities, but it requires a biologically meaningful inductive bias to map latent dimensions to TF identities. Sequence models supply this inductive bias by proposing which TF-gene edges are plausible apriori. Downstream inference can then refine, reweight, or prune these edges using the observed expression matrix, yielding a regulatory network that is both biologically grounded and adapted to the cellular context capture in $W$. This division of labor mirrors the broader theme of this chapter: priors reduce the search space of plausible regulatory explanations, while probabilistic inference provides a mechanism to combine priors with noisy observations and propagate uncertainty into inferred regulatory structure. 

Sequence language models also introduce limitations and modeling choices that matter for GRN applications. First, sequence alone does not encode all determinants of regulation, including chromatin state, TF post-transcriptional modifications, cofactors, and 3D genome organization. Second, regulatory sequences can also be long, requiring practical implementations to require selecting specific regions (e.g., promoter windows, gene bodies, or putative regulatory elements), and choosing how to represent them. Further, a model pretrained on broad corpora may capture general biochemical constraints but still require task-specific fine-tuning to predict regulatory interactions accurately in a particular organism or cell type. These considerations motivate careful dataset construction and evaluation when using language-model predictions as priors. 

Overall, sequence language models provide a principled way to transform raw biological sequence into informative priors for GRN inference. Sequence models complement probabilistic latent-variable models by supplying TF-specific structure that improves identifiability and interpretability, while enabling prior construction at genome scale even when direct experimental regulatory annotations are sparse.

\subsubsection{Transformers and Encoders}
Sequence language models in genomics are most commonly implemented using Transformer encoders \cite{ji2021dnabert, dalla2024nucleotide}. The starting point is a biological sequence (DNA, RNA, or protein), which is first broken into a sequence of discrete tokens. These tokens may be individual bases or amino acids, or short subsequences such as $k$-mers. After tokenization, the input can be written as a length-$L$ sequence $x = (x_1 \dots, x_L)$, where each $x_i$ is a token identity drawn from a finite vocabulary. As token identities are just integers, the model first maps each token to a vector representation using an embedding layer. This produces a matrix of token embeddings with one vector per position in the sequence. A positional signal is then added so the model can distinguish token order, yielding an initial representation,
\begin{equation}
    H^{(0)} = \mathrm{Embed}(x) + \mathrm{PosEnc}(x).
\end{equation}
Intuitively, $H^{(0)}$ encodes both what each token is and where it occurs in the sequence.

The Transformer encoder then applies a stack of $L_{enc}$ layers that repeatedly update these token vectors by incorporating contexts from the rest of the sequence,
\begin{equation}
    H^{(\ell)} = f_{\ell}\!\left(H^{(\ell-1)}\right), \quad \ell = 1,\dots,L_{enc}.
\end{equation}
Each layer uses self-attention to decide, for every position $i$, which other positions in the sequence are most informative for updating its representation. In other words, the representation of token $x$, is refined by taking a learned, weighted combination of information from tokens throughout the sequence. Repeating this process across layers yields contextualized token embeddings $H^{(L_{enc})} \in \mathbb{R}^{L \times d}$, where each row is a $d$-dimensional vector summarizing a position in the sequence in the context of its surrounding and distal sequence. Self-attention is particularly relevant in regulatory genomics because the regulatory impact of a motif often depends on its surrounding context and on other motifs located elsewhere in the sequence.

For many downstream tasks, the encoder output must be converted into a fixed-dimensional representation. A common approach is to prepend a special classification token, often denoted \texttt{<cls>}, and use its final hidden state $h_{\texttt{cls}} \in \mathbb{R}^d$ as a summary of the full input. Alternative pooling strategies include mean pooling over token embeddings or attention pooling. In Chapter \ref{sec:GLM-Prior-Chapter}, we leverage this encoder-to-representation mapping to construct sequence-derived features that can be used to score candidate TF-gene interactions and to define prior structure for downstream probabilistic inference.

\subsubsection{Binary Classification and Cross-Entropy Loss}
The encoder described above produces a fixed-dimensional representation of an input sequence (or sequence pair) that can be used as input to a downstream predictor. In the GRN setting, our goal is to assign a score to a candidate TF-gene pair and interpret this score as evidence for whether a regulatory interaction is present. This can be formalized as a binary classification problem. Given an input representation $h \in \mathbb{R}^d$ (for example, the pooled encoder embedding for a TF-gene pair), predict a label $y \in \{0, 1\}$ indicating the absence $(y=0)$ or presence $(y=1)$ of regulation.

The simplest classifier maps the representation $h$ to a scalar logit $s \in \mathbb{R}$ using a linear layer,
\begin{equation}
    s = w^{\top}h + b,
\end{equation}
where $w \in \mathbb{R}^d$ and $b \in \mathbb{R}$ are learnable parameters. The logit is converted into a probability using the logistic sigmoid,
\begin{equation}
    p(y=1 | h ) = \sigma(s) = \frac{1}{1 + \mathrm{exp}(-s)}.
\end{equation}
Equivalently, one can predict a two-dimensional logit vector $z \in \mathbb{R^2}$ with components $(z_+, z_-)$ and apply a softmax,
\begin{equation}
    p(y=1 |h) = \frac{\mathrm{exp}(z_+)}{\mathrm{exp}(z_+) + \mathrm{exp}(z_-)}.
\end{equation}
Both parameterizations produce a scalar probability $p \in [0, 1]$ that can be used as an interaction score.

Model parameters are typically learned by minimizing the cross-entropy loss. For a single labeled example $(h,y)$ with predicted probability $p = p(y=1 | h)$, the binary cross-entropy loss is,
\begin{equation}
    \mathcal{L} = -y \log p - (1 - y) \log (1-p).
\end{equation}
In many TF-gene interaction datasets, positive regulatory edges are substantially rarer than negatives. A common adjustment is to use a class-weighted variant of cross-entropy to increase the penalty for mistakes on the minority class,
\begin{equation}
    \mathcal{L} = -w_+y \log p - w_- (1 - y) \log (1-p),
\end{equation}
where $w_+, w_- \geq 0$ are user-specified weights. This loss retains the same probabilistic interpretations as standard cross-entropy while providing a simple mechanism to control the precision-recall tradeoff under class imbalance.

In summary, the components in this section define a coherent pipeline from data to regulatory structure. We begin by introducing latent variable models, which provide a compact representation of expression that separates cell-specific regulatory states from gene-specific responses. Building on this view, probabilistic matrix factorization reframes factorization as a generative model, allowing us to quantify uncertainty and incorporate prior knowledge in a principled way. Since exact posterior inference under these models is typically intractable, we then turn to variational inference, which casts inference as an optimization problem and yields practical posterior approximations. Finally, we introduce sequence language models and standard binary classification objectives as an approach for prior knowledge construction. Here we allow learned sequence representations to be converted into TF-gene interaction scores and then assembled into a prior network that anchors downstream inference. Together, these tools support two main methodological threads of this thesis, combining uncertainty-aware probabilistic inference from noisy expression data, with the construction of informative priors to constrain and stabilize this inference.

\subsection{Evaluating GRN Inference}

Evaluating GRN inference methods requires comparing an inferred set of TF-target gene edges against a reference network that acts as a proxy ground truth. In practice, these reference networks are typically assembled from experimentally supported interactions curated in databases, aggregated from the literature, or derived from targeted assays such as ChIP-seq, perturbation experiments, or reporter measurements. Given an inferred network that assigns each candidate TF-gene pair either a binary prediction or continuous score, evaluation proceeds by aligning the predicted edges to the reference set and measuring how well high-confidence predictions recover known interactions while avoiding edges not supported by the reference.

\subsubsection{Metrics}

Most GRN inference methods produce ranked edge scores rather than a single hard network, allowing evaluation to be framed as a binary classification problem over all candidate TF-gene pairs. Let $y \in \{0, 1 \}$ denote the reference label for a candidate edge, where $1$ defines an edge as present in the reference network and $0$ defines an edge is not present. Further, let $\hat{s}$ denote the inferred score for that edge. Choosing a threshold $\tau$ converts scores into binary predictions $\hat{y}(\tau) = \mathbb{I}[\hat{s} \geq \tau]$. For any fixed threshold, predictions can be summarized by the confusion matrix counts for true positives (TP), false positives (FP), false negative (FN), and true negatives (TN). These quantities form the basis of common evaluation metrics.

GRN inference is typically a highly imbalanced prediction problem, where true regulatory edges are rare relative to the number of possible TF-gene pairs. Metrics that explicitly account for class imbalance and focus on the quality of positive predictions are therefore especially informative. We start by defining two useful metrics, precision and recall \eqref{eq:precision-recall}. 
\begin{equation} \label{eq:precision-recall}
    \mathrm{Precision} = \frac{\mathrm{TP}}{\mathrm{TP + FP}}, \quad \mathrm{Recall} = \frac{\mathrm{TP}}{\mathrm{TP + FN}}.
\end{equation}
Precision measures how reliable predicted edges are by determining how many predicted positives are supported by the reference, while recall measures how completely the method recovers reference edges. To combine these two quantities into a single summary of performance, the F1 score \ref{eq:f1-score} summarizes the precision-recall tradeoff as their harmonic mean,
\begin{align} \label{eq:f1-score}
    \mathrm{F1} &= 2 \cdot \frac{\mathrm{Precision} \cdot \mathrm{Recall}}{\mathrm{Precision} + \mathrm{Recall}} \\
    &= \frac{2\mathrm{TP}}{2\mathrm{TP} + \mathrm{FP} + \mathrm{FN}}.
\end{align}
When methods output continuous scores, precision and recall vary with the threshold $\tau$. This motivates evaluation across thresholds using the precision-recall (PR) curve and its area, the area under the PR curve (AUPRC). AUPRC is widely used for GRN inference because it emphasizes performance on the positive class and remains informative under severe imbalance, where many other metrics may appear deceptively high.

In addition to positive-class performance, it is sometimes useful to report class-specific error rates. The negative analogs are,
\begin{align} \label{eq:neg-metrics}
    \text{Negative Precision} = \frac{\mathrm{TN}}{\mathrm{TN + FN}}, \quad \text{Negative Recall} = \frac{\mathrm{TN}}{\mathrm{TN + FP}}.
\end{align}
These metrics (\ref{eq:neg-metrics}) quantify, respectively, the reliability of predicted negatives and the ability to avoid false positive edges.

A single threshold summary that uses all four confusion matrix entries is the Matthews correlation coefficient (MCC),
\begin{equation} \label{eq:MCC}
    \mathrm{MCC} = \frac{\mathrm{TP} \cdot \mathrm{TN} - \mathrm{FP} \cdot \mathrm{FN}}{\sqrt{(\mathrm{TP} + \mathrm{FP})(\mathrm{TP} + \mathrm{FN})(\mathrm{TN} + \mathrm{FP})(\mathrm{TN} + \mathrm{FN})}}.
\end{equation}
MCC (\ref{eq:MCC}) can be interpreted as a correlation between predicted and true labels, and it is often preferred over accuracy in imbalanced settings because it penalizes trivial solutions, such as predicting all edges as negative.

Another metric commonly used for GRN inference evaluation is the receiver operating characteristic (ROC) curve, which plots the true positive rate (TPR) against the false positive rate (FPR) as the threshold varies,
\begin{equation} \label{eq: ROC}
    \mathrm{TPR} = \frac{\mathrm{TP}}{\mathrm{TP} + \mathrm{FN}}, \quad \mathrm{FPR} = \frac{\mathrm{FP}}{\mathrm{FP} + \mathrm{TN}}.
\end{equation}
The area under the ROC curve (AUC-ROC) summaries ranking performance across thresholds and has a probabilistic interpretation. It represents the probability that a randomly chosen positive edge is scored higher than a randomly chosen negative edge. However, because the negative class is typically very large in GRN inference, AUC-ROC can remain high even when precision is low. This further motivates using AUPRC instead for a more discriminative metric when comparing methods in practice.

An important consideration is that evaluating GRNs is a fundamentally challenging task. First, reference networks are imperfect proxies for truth. Second, curated databases and literature-derived references are incomplete, with many true regulatory edges missing because they have not been tested or reported, creating apparent false positives when methods predict real interactions not present in the reference. Conversely, reference networks can contain false positives due to context mismatch, for example, evidence from a different cell type or condition. Further, negative edges are rarely experimentally established. Edges not present in a reference are often better viewed as unknown rather than truly absent. These issues are especially pronounced outside model organisms, where gold standards are sparse, uneven across TFs and genes, and biased towards well studied regulators. 

For these reasons, quantitative metrics such as AUPRC, F1 score (\ref{eq:f1-score}), precision and recall (\ref{eq:precision-recall}), MCC (\ref{eq:MCC}), and AUROC are best interpreted as measures of agreement with an available reference, not definitive measures of biological truth. They remain essential for benchmarking, providing a consistent way to compare methods, training paradigms, and priors, while the limitations of reference networks motivate complementary analyses when drawing biological conclusions.  

\subsubsection{Uncertainty Quantification and Calibration}

A key advantage of probabilistic GRN inference is that it can provide not only a point estimate for each TF-target gene interaction, but also the model's uncertainty in that estimate. In this setting, uncertainty reflects how strongly the observed expression matrix $W$ constrains a candidate regulatory effect under the assumed generative model. Probabilistic matrix factorization infers a posterior distributions over latent variables and interaction parameters. Under variational inference, this posterior is approximated by a tractable distribution $q(z)$, which yields both posterior means and posterior variances. For each TF-gene interaction parameter, such as an element of the regulatory component of $V$ or a derived edge weight, we summarize the interaction by a point estimate $\hat{\theta}_{t,g} = \mathbb{E}_{q}[\theta_{t,g}]$ and quantify uncertainty with the corresponding posterior variance, 
\begin{equation}
\mathrm{Var}_{q}(\theta_{t,g}) = \mathbb{E}_{q}[\theta_{t,g}^2] - \big(\mathbb{E}_{q}[\theta_{t,g}]\big)^2.
\end{equation}
Low posterior variance indicates that the interaction is consistently supported across plausible latent configurations under the model, while high variance indicates that the data permit a wider range of interaction strengths.

To assess whether posterior uncertainty is meaningful for prioritization, calibration can be evaluated by testing whether low-uncertainty interactions are more likely to agree with a reference network than high uncertainty interactions. For example, TF-gene interactions can be ranked by their posterior variances to construct $10$ cumulative bins corresponding to the lowest $10\%$, $20\%$, $30\%$, and so on of variance. For each bin, an overlap AUPRC can be computed using only those interactions that $(i)$ fall within the bin, and $(ii)$ are present in the gold standard, ensuring that the evaluation is performed on edges that can be scored against the reference. Under good calibration, bins containing lower posterior variance interactions should exhibit higher overlap AUPRC than bins that include progressively more uncertain interactions, indicating that posterior uncertainty provides an informative ranking of interaction reliability. 

\chapter{Probabilistic Matrix Factorization for Gene Regulatory Network Inference}
\label{chp-1}
\label{sec:PMF-GRN-Chapter}
This chapter is a reprint of the published paper "PMF-GRN: a variational inference approach to single-cell gene regulatory network inference using probabilistic matrix factorization". \\
\textbf{Claudia Skok Gibbs}, Omar Mahmood, Richard Bonneau, and Kyunghyun Cho. \\
\textit{Genome Biology}, \textbf{2024}. 
\section{Abstract}

Inferring gene regulatory networks (GRNs) from single-cell data is challenging due to heuristic limitations. Existing methods also lack estimates of uncertainty.  Here we present Probabilistic Matrix Factorization for Gene Regulatory Network Inference (PMF-GRN). Using single-cell expression data, PMF-GRN infers latent factors capturing transcription factor activity and regulatory relationships. Using variational inference allows hyperparameter search for principled model selection and direct comparison to other generative models. We extensively test and benchmark our method using real single-cell datasets, and synthetic data. We show that PMF-GRN infers GRNs more accurately than current state-of-the-art single-cell GRN inference methods, offering well-calibrated uncertainty estimates.

\section{Introduction}

An essential problem in systems biology is to extract information from genome wide sequencing data to unravel the mechanisms controlling cellular processes within heterogeneous populations \cite{hecker2009gene}. Gene regulatory networks (GRNs) that annotate regulatory relationships between transcription factors (TFs) and their target genes \cite{chai2014review} have proven to be useful models for stratifying functional differences between cells \cite{karlebach2008modelling, aijo2009learning, nachman2004inferring, burdziak2019nonparametric} that can arise during normal development \cite{allaway2021genetic}, responses to environmental signals \cite{jackson2020gene} and dysregulation in the context of disease \cite{ciofani2012validated, ji2019inflammatory, yosef2013dynamic}.

GRNs cannot be directly measured with current sequencing technology. Instead, methods must be developed to piece together snapshots of transcriptional processes in order to reconstruct a cell's regulatory landscape \cite{mercatelli2020gene}. Initial approaches to GRN inference relied on Microarray technology \cite{huynh2010inferring, wang2006inferring, chang2008fast}, a hybridization-based method to measure the expression of thousands of genes simultaneously \cite{dufva2009introduction}. This technology was biased as it was limited to only those genes that were annotated at the time, which in turn presented challenges for inferring the complete regulatory landscape \cite{hecker2009gene}. Subsequently, the high-throughput sequencing method RNA-seq provided a genome wide readout of transcriptional output, allowing for the detection of novel transcripts \cite{wang2009rna} and thus improving GRN inference potential. More recently, single-cell RNA-seq technology has enabled the characterization of gene expression profiles within heterogeneous populations \cite{saliba2014single}, vastly increasing the potential for GRN inference algorithms \cite{akers2021gene, lahnemann2020eleven}. In contrast to bulk RNA experiments (Microarray and RNA-seq) that average measurements of gene expression across heterogenous cell populations, GRNs inferred from single-cell data have the advantage of unmasking biological signal in individual cells \cite{chen2019single}. 

Several matrix factorization approaches have been proposed to overcome the limitations of reconstructing GRNs from Microarray data \cite{ochs2012matrix}. These include use of statistical techniques such as Singular Value Decomposition and Principal Component Analysis \cite{alter2000singular}, Bayesian Decomposition \cite{moloshok2002application}, and Non-negative Matrix Factorization \cite{kim2003subsystem, brunet2004metagenes, gao2005improving}. More recently, matrix factorization approaches have been applied to integrative analysis of DNA methylation and miRNA expression data \cite{yang2016non}, as well as single-cell RNA-seq and single-cell ATAC-seq data \cite{duren2018integrative}. However, to the best of our knowledge, these matrix factorization approaches have not yet been used to infer GRNs from single-cell gene expression data. Meanwhile, several regression-based methods have been proposed to learn GRNs from single-cell RNA-seq and single-cell ATAC-seq to capture regulatory relationships at single-cell resolution \cite{hu2020integration}. So far, these integrative approaches to GRN inference have been successfully implemented using regularized regression \cite{skok2022high}, self-organizing maps \cite{jansen2019building}, tree-based regression \cite{van2020scalable}, and Bayesian Ridge regression \cite{kamimoto2023dissecting}. 

Although regression-based methods for inferring GRNs from single-cell data are available, they still suffer from significant limitations \cite{aijo2016biophysically}. Firstly, these methods are designed for specific input datasets, such as bulk or single-cell RNA-seq, causing issues when new data becomes available or new assumptions are required in the model. This can result in inaccurate predictions if the new data or assumptions are not well integrated into the existing model, leading to the need for a complete re-design of the algorithm, which can be costly and time-consuming. Additionally, these methods typically focus on inferring a single GRN that explains the available data, without performing hyperparameter search to determine the optimal model. This can lead to heuristic model selection, with no justification for the approach taken or evidence that the best possible model has been selected. Conversely, hyperparameter search ensures the accuracy of the GRN inference algorithm by finding the optimal model that fits the data well while avoiding overfitting. Regression-based GRN inference algorithms that do not perform hyperparameter search may miss important data features or overemphasize irrelevant ones, leading to inaccurate or incomplete models. 
Moreover, these methods do not provide an indication of their uncertainty about the predictions that they make.
Finally, several regression-based GRN inference algorithms struggle to scale optimally to the size of typical single-cell datasets, limiting inference to small subsets of data or requiring enormous amounts of computational time.

In this study, we introduce PMF-GRN, a novel approach that uses probabilistic matrix factorization \cite{mnih2007probabilistic} to infer gene regulatory networks from single-cell gene expression and chromatin accessibility information. This approach extends previous methods that applied matrix factorization for GRN inference with Microarray data, to address the current limitations in regression-based single-cell GRN inference. We implement our approach in a probabilistic setting with variational inference, which provides a flexible framework to incorporate new assumptions or biological data as required, without changing the way the GRN is inferred. We also use a principled hyperparameter selection process, which optimizes the parameters of our probabilistic model for automatic model selection. In this way, we replace heuristic model selection by comparing a variety of generative models and hyperparameter configurations before selecting the optimal parameters with which to infer a final GRN. Our probabilistic approach provides uncertainty estimates for each predicted regulatory interaction, serving as a proxy for the model confidence in each predicted interaction. Uncertainty estimates can be useful in the situation where there are limited validated interactions or a gold standard is incomplete. By using stochastic gradient descent (SGD), we perform GRN inference on a GPU, allowing us to easily scale to a large number of observations in a typical single-cell gene expression dataset. Unlike many existing methods, PMF-GRN is not limited by pre-defined organism restrictions, making it widely applicable for GRN inference. 

To demonstrate the novelty and advantages of PMF-GRN, we apply our method to datasets from \textit{Sacchromyces cerevisiae}, human Peripheral Blood Mononuclear Cells (PBMCs) and BEELINE.
In our first experiment, we apply our method to two single cell gene expression datasets for the model organism \textit{S. cerevisiae}. We evaluate our model's performance in a normal inference setting, as well as with cross-validation and noisy data. To assess the accuracy of predicted regulatory interactions, we evaluate all regulatory predictions using Area Under the Precision Recall Curve (AUPRC) against database derived gold standards. Our findings show that the uncertainty estimates are well-calibrated for inferred TF-target gene interactions, as the accuracy of predictions increases when the associated uncertainty decreases. Here, in comparison to three state-of-the-art regression-based methods for inferring single cell GRNs, namely the Inferelator \cite{skok2022high}, Scenic \cite{van2020scalable}, and Cell Oracle \cite{kamimoto2023dissecting}, our method demonstrates an overall improved performance in recovering the true underlying GRN. 
Additionally, we apply our method to a PBMC dataset and explore the inferred TFA profiles in the context of annotated cell types and specific immune TFs. We investigate regulatory edges in our inferred GRN and find compelling support for our predictions.
Lastly, we benchmark our method using six synthetic datasets generated from BEELINE \cite{pratapa2020benchmarking} and demonstrate consistent outperformance of PMF-GRN compared to the baseline.

\section{Results}

\subsection{The PMF-GRN Model}

\begin{figure}[htbp]
    \centering
    \includegraphics[width=0.95\textwidth]{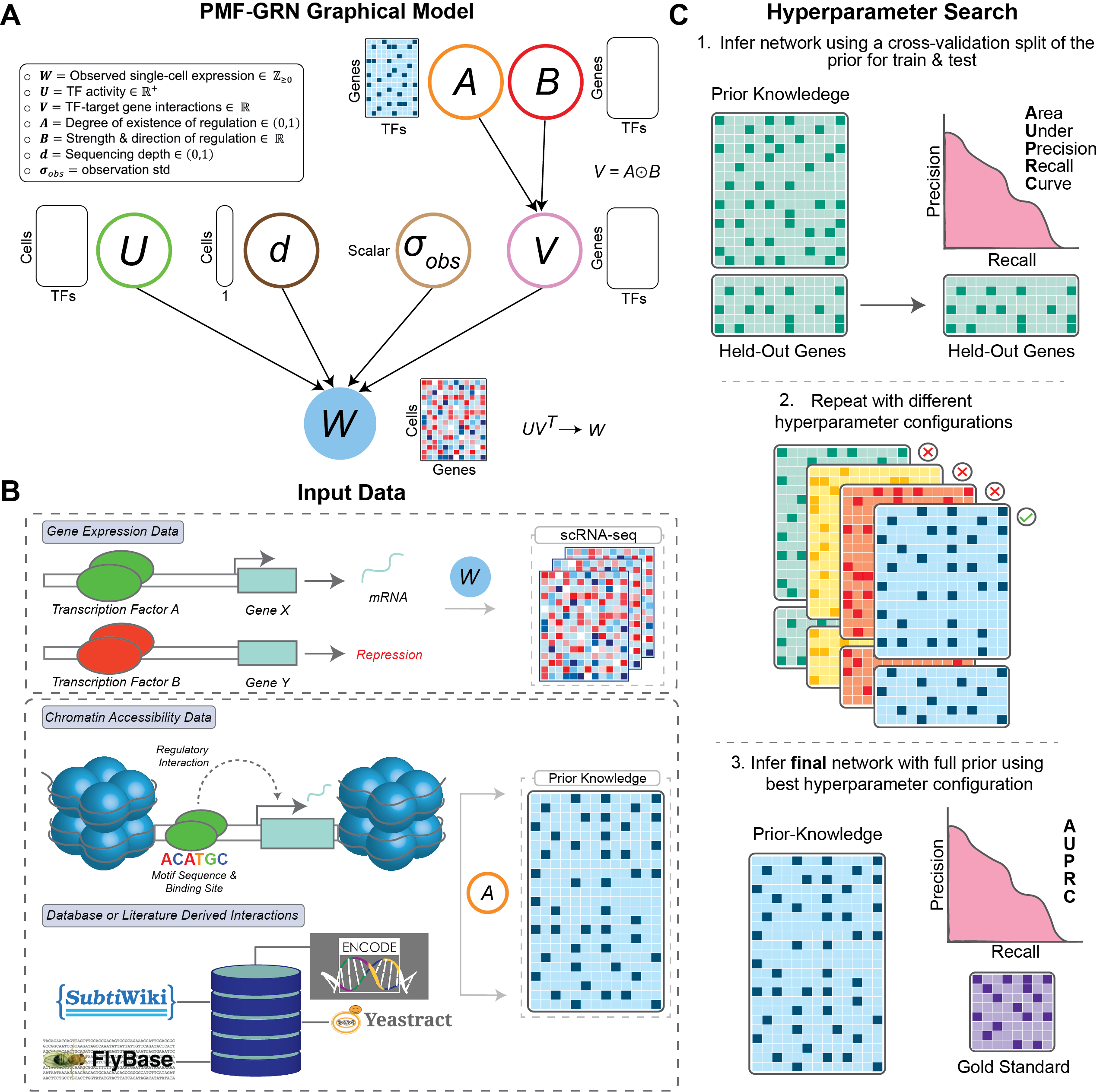}
    \caption{({\bf A}) PMF-GRN graphical model overview. Input single-cell gene expression $W$ is decomposed into several latent factors. Information obtained from chromatin accessibility data or genomics databases is incorporated into the prior distribution for $A$. ({\bf B}) Input experimental data for PMF-GRN includes single-cell RNA-seq gene expression data. Prior-known TF-target gene interactions can be obtained using chromatin accessibility in parallel with known TF motifs, or through databases or literature derived interactions.({\bf C}) Hyperparameter selection process is performed for optimal model selection. The provided prior-known network is split into a train and validation dataset. 80\% of the prior-known information is used to infer a GRN, while the remaining 20\% is used for validation by computing AUPRC. This process is repeated multiple times, using different hyperparameter configurations in order to determine the optimal hyperparameters for the GRN inference task at hand. Finally, using the  optimal hyperparameters, a final network is inferred using the full prior and evaluated using an independent gold standard.}
    \label{fig:figure-1}
\end{figure}

The goal of our probabilistic matrix factorization approach is to decompose observed gene expression into latent factors, representing TF activity (TFA) and regulatory interactions between TFs and their target genes. These latent factors, which represent the underlying GRN, cannot be measured experimentally, unlike gene expression. We model an observed gene expression matrix $W \in \mathbb{R}^{N \times M}$ using a TFA matrix $U \in \mathbb{R}_{>0}^{N \times K}$, a TF-target gene interaction matrix $V \in \mathbb{R}^{M \times K}$, observation noise $\sigma_{obs} \in (0, \infty)$ and sequencing depth $d \in (0,1)^N$, where $N$ is the number of cells, $M$ is the number of genes and $K$ is the number of TFs. We rewrite $V$ as the product of a matrix $A \in (0, 1)^{M \times K}$, representing the degree of existence of an interaction, and a matrix $B \in \mathbb{R}^{M \times K}$ representing the interaction strength and its direction:
\begin{align*}
    V = A \odot B,
\end{align*}
where $\odot$ denotes element-wise multiplication. An overview of the graphical model is shown in Figure \ref{fig:figure-1}A.

These latent variables are mutually independent {\it a priori}, i.e., 
\begin{equation*}
    p(U, A, B, \sigma_{obs}, d) = p(U)p(A)p(B)p(\sigma_{obs})p(d).
\end{equation*} 

For the matrix $A$, prior hyperparameters represent an initial guess of the interaction between each TF and target gene which need to be provided by a user. These can be derived from genomic databases or obtained by analyzing other data types, such as the measurement of chromosomal accessibility, TF motif databases, and direct measurement of TF-binding along the chromosome, as shown in Figure \ref{fig:figure-1}B (see \nameref{sec:pmf-methods} section for details). 

The observations $W$ result from a matrix product $UV^\top$. We assume noisy observations by defining a distribution over the observations with the level of noise $\sigma_{obs}$, i.e., $p(W|U, V= A \odot B, \sigma_{obs}, d)$.

Given this generative model, we perform posterior inference over all the unobserved latent variables; $U$, $A$, $B$, $d$ and $\sigma_{obs}$, and use the posterior over $A$ to investigate TF-target gene interactions.
Exact posterior inference with an arbitrary choice of prior and observation probability distributions is, however, intractable. We address this issue by using variational inference \cite{blei2017variational, ranganath2014black}, where we approximate the true posterior distributions with tractable, approximate (variational) posterior distributions.

We minimize the KL-divergence $D_{KL}(q\|p)$ between the two distributions with respect to the parameters of the variational distribution $q$, where $p$ is the true posterior distribution. This allows us to find an approximate posterior distribution $q$ that closely resembles $p$. This is equivalent to maximizing the evidence lower bound (ELBO) i.e. a lower bound to the marginal log likelihood of the observations $W$:

\begin{align*}
    \log p(W) \geq \mathbb{E}_{U, A, B, \sigma_{obs}, d \sim q(U, A, B, \sigma_{obs}, d)} [ & \log p(W|U, V = A \odot B,  \sigma_{obs}, d)\\
    & + \log p(U, A, B, \sigma_{obs}, d)\\
    & - \log q(U, A, B, \sigma_{obs}, d)]
\end{align*}
The mean and variance of the approximate posterior over each entry of $A$, obtained from maximizing the ELBO, are then used as the degree of existence of an interaction between a TF and a target gene and its uncertainty, respectively.

It is important to note that matrix factorization based GRN inference is only identifiable up to a latent factor (column) permutation. 
In the absence of prior information, the probability that the user assigns TF names to the columns of $U$ and $V$ in the same order that the inference algorithm implicitly assigns TFs to these columns is $\frac{1}{K!}$, is essentially 0 for any reasonable value of $K$. Incorporating prior-knowledge of TF-target gene interactions into the prior distribution over $A$ is therefore essential in order to provide the inference algorithm with the information of which column corresponds to which TF.

With this identifiability issue in mind, we design an inference procedure that can be used on any prior-knowledge edge matrices, described in Figure \ref{fig:figure-1}C. The first step is to randomly hold out prior information for some percentage of the genes in $p(A)$ (we choose $20\%$) by leaving the rows corresponding to these genes in $A$ but setting the prior logistic normal means for all entries in these rows to be the same low number.

The second step is to carry out a hyperparameter search using this modified prior-knowledge matrix. The early stopping and model selection criteria are both the `validation' AUPRC of the posterior point estimates of $A$, corresponding to the held out genes, against the entries for these genes in the full prior hyperparameter matrix. This step is motivated by the idea that inference using the selected hyperparameter configuration should yield a GRN whose columns correspond to the TF names that the user has assigned to these columns.

The third step is to choose the hyperparameter configuration corresponding to the highest validation AUPRC and perform inference using this configuration with the full prior. An importance weighted estimate of the marginal log likelihood is used as the early stopping criterion for this step. The resulting approximate posterior provides the final posterior estimate of $A$.

\subsection{Advantages of PMF-GRN}

Existing methods almost always couple the description of the data generating process with the inference procedure used to obtain the final estimated GRN \cite{skok2022high, kamimoto2023dissecting, van2020scalable}. Designing a new model thus requires designing a new inference procedure specifically for that model, which makes it difficult to compare results across different models due to the discrepancies in their associated inference algorithms. 
Furthermore, this {\it ad hoc} nature of model building and inference algorithm design often leads to the lack of a coherent objective function that can be used for proper hyperparameter search as well as model selection and comparison,
as evident in \cite{skok2022high}. Heuristic model selection in available GRN inference methods presents the challenge of determining and selecting the optimal model in a given setting.

The proposed PMF-GRN framework decouples the generative model from the inference procedure. Instead of requiring a new inference procedure for each generative model, it enables a single inference procedure through (stochastic) gradient descent with the ELBO objective function, across a diverse set of generative models. Inference can easily be performed in the same way for each model. Through this framework, it is possible to define the prior and likelihood distributions as desired with the following mild restrictions: we must be able to evaluate the joint distribution of the observations and the latent variables, the variational distribution and the gradient of the log of the variational distribution. 

The use of stochastic gradient descent in variational inference comes with a significant computational advantage. As each step of inference can be done with a small subset of observations, we can run GRN inference on a very large dataset without any constraint on the number of observations. This procedure is further sped up by using modern hardware, such as GPUs. 

Under this probabilistic framework, we carry out model selection, such as choosing distributions and their corresponding hyperparameters, in a principled and unified way. Hyperparameters can be tuned with regard to a predefined objective, such as the marginal likelihood of the data or the posterior predictive probability of held out parts of the observations. We can further compare and choose the best generative model using the same procedure. 

This framework allows us to encode any prior knowledge via the prior distributions of latent variables. For instance, we incorporate prior knowledge about TF-gene interactions as hyperparameters that govern the prior distribution over the matrix $A$. If prior knowledge about TFA is available, this can be similarly incorporated into the model via the hyperparameters of the prior distribution over $U$. 

Because our approach is probabilistic by construction, inference also estimates uncertainty without any separate external mechanism. 
These uncertainty estimates can be used to assess the reliability of the predictions, i.e., more trust can be placed in interactions that are associated with less uncertainty. We verify this correlation between the degree of uncertainty and the accuracy of interactions in the experiments.

Overall, the proposed approach of probabilistic matrix factorization for GRN inference is scalable, generalizable and aware of uncertainty, which makes its use much more advantageous compared to most existing methods. 

\subsection{PMF-GRN Recovers True Interactions in Simple Eukaryotes}

To evaluate PMF-GRN's ability to infer informative and robust GRNs, we leverage two single-cell RNA-seq datasets from the model organism \textit{Saccharomyces cerevisiae} \cite{jackson2020gene, jariani2020new}. This eukaryote, being relatively simple and extensively studied, provides a reliable gold standard \cite{tchourine2018condition} for assessing the performance of different GRN inference methods. We conduct three experiments to compare the performance of three state-of-the-art GRN inference methods, the Inferelator (AMuSR, BBSR, and StARS) \cite{skok2022high}, SCENIC \cite{van2020scalable}, and CellOracle \cite{kamimoto2023dissecting}. Throughout these experiments, each method is provided with the exact same single-cell RNA-seq datasets (GSE125162 \cite{jackson2020gene}: N cells $= 38,225$, GSE144820 \cite{jariani2020new}: N cells $= 6,118$, combined: N cells $= 44,343$ by M genes $= 6,763$), prior-knowledge (M genes $= 6,885$ by K TFs $= 220$) and gold standard (M genes $= 993$ by K TFs $= 98$).

\begin{figure}[htbp]
    \centering
    \includegraphics[width=0.95\textwidth]{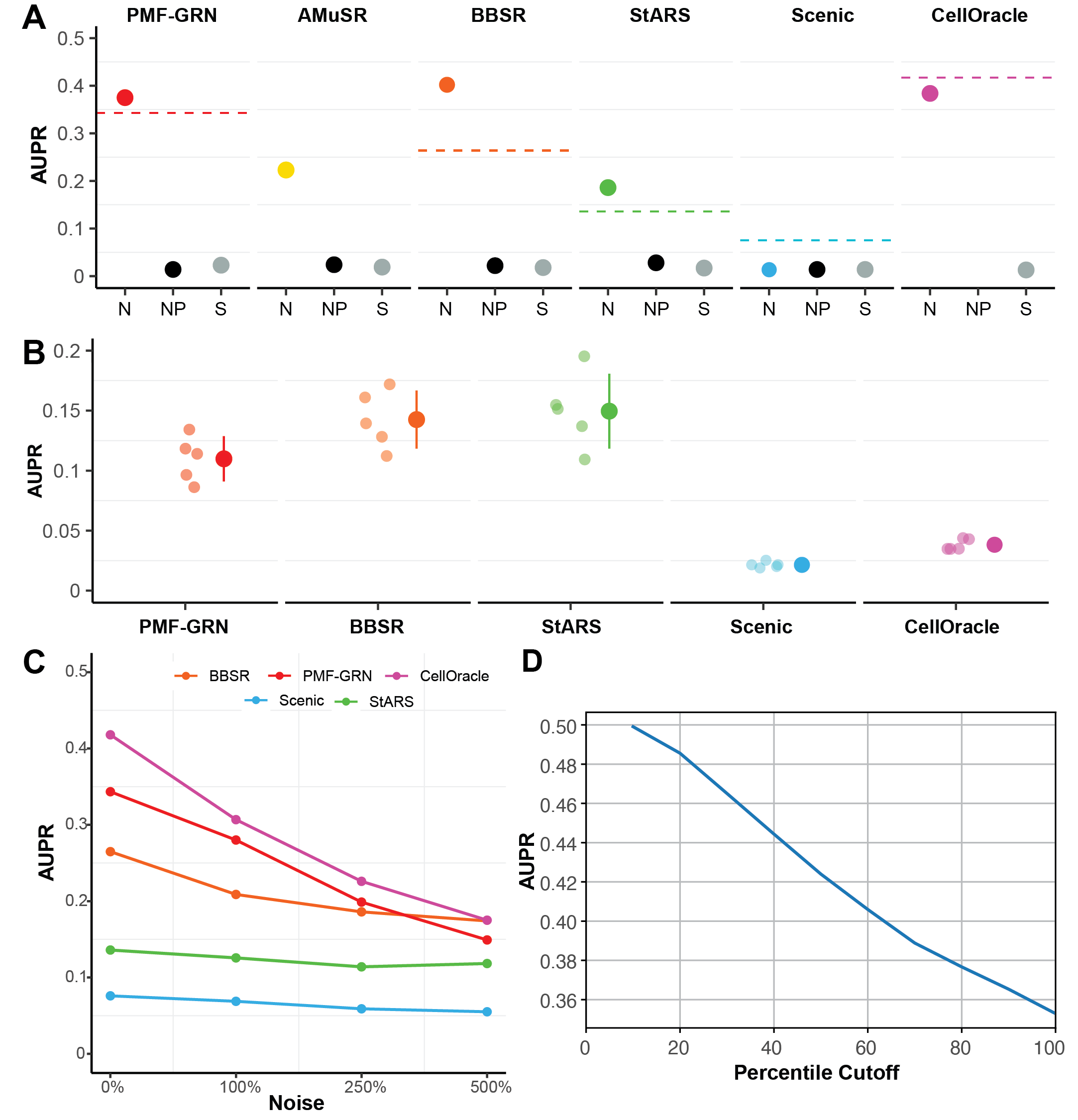}
    \caption{GRN inference in \textit{S. cerevisiae}. ({\bf A}) Consensus Network AUPR with a normal prior-knowledge matrix (N): PMF-GRN (red) performance compared to Inferelator algorithms (AMuSR in yellow, BBSR in orange, StARS in green), SCENIC (blue), and CellOracle (purple). Dashed line represents the baseline if expression data is combined. Negative controls: no prior information (NP - black) and shuffled prior information (S - gray). ({\bf B}) 5-Fold Cross-Validation Baseline: Each dot with low opacity represents one of the five experiments. Colored dots and lines depict the mean AUPR $\pm$ standard deviation for each GRN inference method. ({\bf C}) GRNs inferred with increasing amounts of noise added to the prior. ({\bf D}) Calibration results on \textit{S.cerevisiae} (GSE125162 only) dataset. Posterior means are cumulatively placed in bins based on their posterior variances. AUPRC for each of these bins is computed against the gold standard (see \nameref{sec:pmf-methods} section for details)} 
    \label{fig:auprc-yeast}
\end{figure}

In the first experiment, we infer GRNs for each of the two yeast datasets and average the posterior means of $A$ to simulate a "multi-task" GRN inference approach. Using AUPRC, we demonstrate that PMF-GRN outperforms AMuSR, StARS, and SCENIC, while performing competitively with BBSR and CellOracle (Figure \ref{fig:auprc-yeast}A). We next combine the two expression datasets into one observation to test whether each method can discern the overall GRN accurately when data is not cleanly organized into tasks. This experiment reveals a substantial performance decrease for BBSR, indicating its dependence on organized gene expression tasks. This finding suggests potential challenges for BBSR in more complex organisms with less well-defined cell types or conditions. For benchmarking purposes we provide two negative controls for each method, a GRN inferred without prior information (No Prior), and a GRN inferred using shuffled prior information (Shuffled Prior). For all methods, these negative controls achieve an expected low AUPRC. It is essential to note that for CellOracle, an experiment with no prior information could not be performed. This is due to the fact that by design, CellOracle cannot learn regulatory edges that are not included in the prior information.

In our comparitive GRN inference analysis, we assess the number of edges predicted in common by each algorithm, on the individual \textit{S. cerevisiae} datasets. We do so by computing the Intersection over Union (IoU) score, filtering each GRN to the top $25\%$ of interactions to remove noisy predictions. Notably, PMF-GRN obtains an IoU score of $15.69\%$, outperforming alternative algorithms such as SCENIC ($3.17\%$), AMuSR ($12.46\%$), BBSR ($14.56\%$), and StARS ($11.78\%$). The superior performance of PMF-GRN can be attributed to an ability to discern meaningful regulatory interactions, thereby enriching the consensus among predictions. Importantly, our findings underscore a limitation of CellOracle, which achieves an IoU score of $30.28\%$. This algorithm, while proficient, can only ascertain edges present in the prior-knowledge matrix. Consequently, the two yeast GRNs inferred display high similarity, reflecting an inherent constraint. This characteristic imparts a degree of predictability to CellOracle, limiting its capacity to discover novel interactions beyond the established prior-knowledge. In contrast, PMF-GRNs IoU score is indicative of a more diverse and comprehensive set of common edges. This highlights PMF-GRNs capability to capture nuanced regulatory relationships as a robust and versitle tool for GRN inference.

In a second experiment, we implement a 5-fold cross-validation approach to establish a baseline for each model. Cross-validation is crucial for evaluating the generalization ability of machine learning models like PMF-GRN, particularly in predicting TF-target gene interactions with limited data, a common scenario in experimental settings. To streamline the analysis, we combine the two \textit{S. cerevisiae} single-cell RNA-seq datasets into a single observation matrix. The cross-validation process involves an $80\%-20\%$ split of the gold standard, where a network is inferred using $80\%$ as "prior-known information" and evaluated using the remaining $20\%$. This process is iterated five times with different random splits to yield meaningful results.  We observe that PMF-GRN outperforms SCENIC and CellOracle, while achieving similar performance to BBSR and StARS (Figure \ref{fig:auprc-yeast}B). We note that for this experiment, we are unable to implement the AMuSR algorithm as it is a multi-task inference approach that requires more than one task (dataset). 

In a third experiment, we evaluate the robustness of each GRN inference method in the presence of noisy prior information. We conduct GRN inference with increasing levels of noise introduced into the prior knowledge. Specifically, the prior information begins with $1\%$ non-zero edges, and we systematically introduce noise to observe the performance of each method. The noise levels are varied from zero noise (original prior, $1\%$ non-zero edges), to $100\%$ noise (resulting in $2\%$ non-zero edges), $250\%$ noise ($3.5\%$ non-zero edges), and $500\%$ noise ($6\%$ non-zero edges). Our findings, illustrated in Figure \ref{fig:auprc-yeast}C, reveal that as the noise in the prior information increases, PMF-GRNs AUPRC experiences a slow decline, mirroring the behavior observed in CellOracle. Notably, PMF-GRN consistently outperforms BBSR, StARS, and SCENIC under these noise conditions, showcasing its robustness in accurately inferring GRNs from noisy priors. These results underscore PMF-GRN as one of the most robust approaches in the face of noisy prior information, thereby emphasizing its utility in practical applications.

\begin{figure}[htbp]
    \centering
    \includegraphics[width=0.70\textwidth]{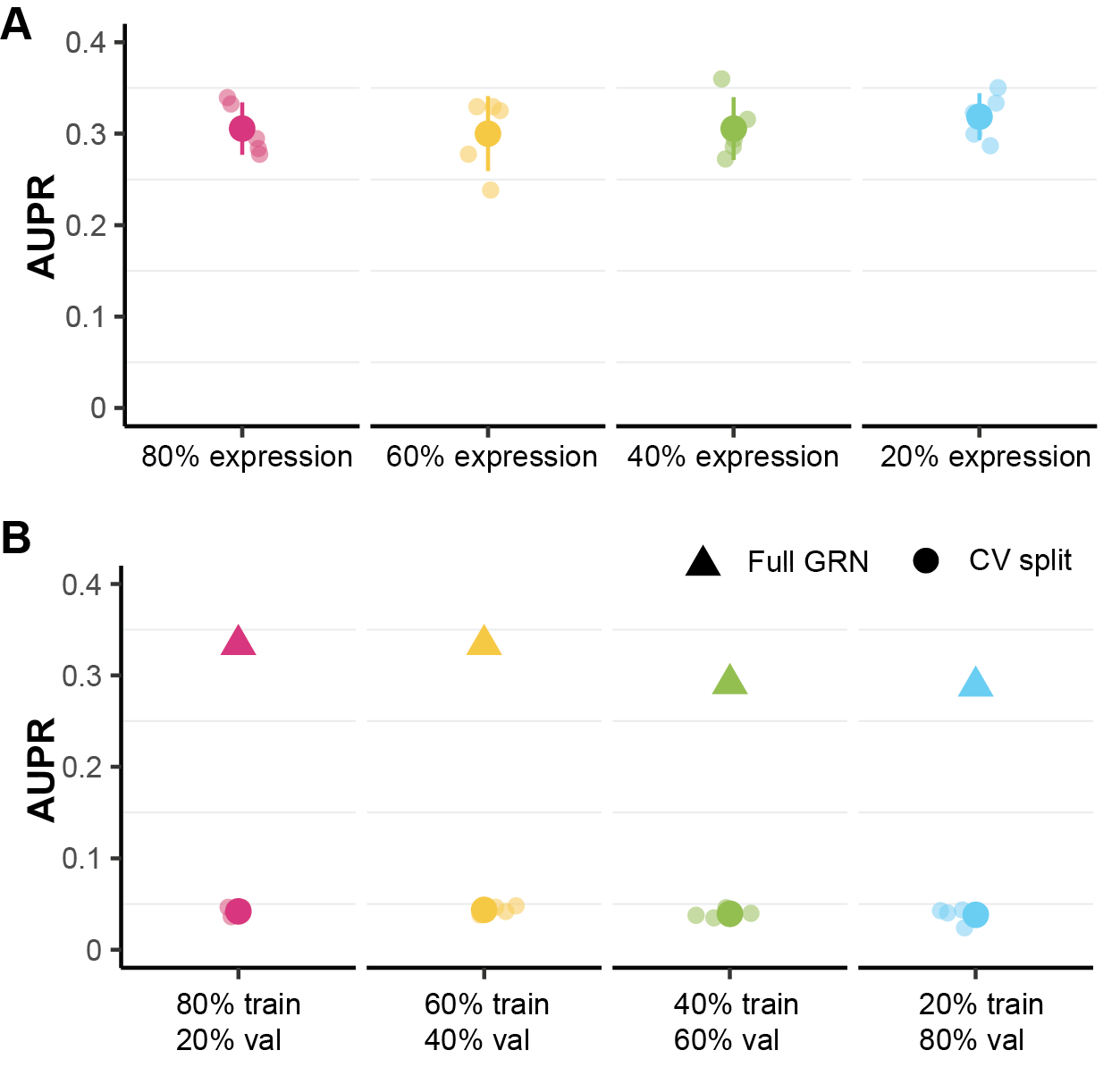}
    \caption{({\bf A}) GRNs inferred by downsampling \textit{S. cerevisiae} expression data. ({\bf B}) Hyperparameter search performed on 4 different ratios of cross-validation. Dots represent validation AUPRC from hyperparameter search during cross-validation, triangle represents AUPRC from a GRN learned using the most optimal hyperparameters for each ratio. } 
    \label{fig:yeast-downsample-cv-experiment}
\end{figure}

To further emphasize PMF-GRN's robustness in a diverse number of settings, we perform the following two experiments. In the first experiment, we examine the performance of PMF-GRN using different sizes of downsampled yeast expression (Figure \ref{fig:yeast-downsample-cv-experiment}A). The downsampling procedure involved reducing the expression data to sizes of $80\%$, $60\%$, $40\%$, and $20\%$, with each size undergoing random sampling five times to generate five distinct datasets per sample size. Remarkably, the AUPRC performance exhibits noteworthy stability across the downsampling variations. Despite the reduction in dataset size, PMF-GRN consistently demonstrates an ability to learn accurate GRNs as evidenced by the sustained AUPRC performance. These findings underscore the robustness of PMF-GRN, suggesting its reliability even under conditions of diminished dataset sizes, a critical consideration for practical applications where data availability may be limited.

In a subsequent experiment, we explore the impact of different cross-validation split sizes on hyperparameter tuning for PMF-GRN using the \textit{S. cerevisiae} prior-knowledge (Figure \ref{fig:yeast-downsample-cv-experiment}B). Four distinct cross-validation splits, ranging from $80\%$ training and $20\%$ validation, to  $20\%$ training and $80\%$ validation, were employed. For each split, we conducted a hyperparameter search across five samples, selecting the optimal hyperparameters based on the highest validation AUPRC. We then selected the best overall hyperparameters from each split to learn a GRN on the full dataset, in order to demonstrate the downstream effect of cross validation split choice on GRN inference. Surprisingly, our results revealed that the choice of cross-validation split size had a marginal impact on the overall performance of the inferred GRN. Specifically, the AUPRC values for the full GRN remained nearly unchanged regardless of whether an $80\%$ train and $20\%$ validation or $60\%$ train and $40\%$ validation split where employed. Even with more disparate splits, such as $40\%$ train and $60\%$ validation, or $20\%$ train and $80\%$ validation, the decrease in AUPRC was only minor. This implies that PMF-GRN exhibits robustness in hyperparameter selection, with the algorithm consistently converging to optimal settings across varying cross-validation scenarios.

From our experiments on \textit{S. cerevisiae} data, several key observations emerge. First, PMF-GRN consistently outperforms the Inferelator in recovering true GRNs, surpassing two Inferelator algorithms (AMuSR and StARS) and performing similarly to BBSR. Notably, when expression data is not separated into tasks, PMF-GRN outperforms BBSR. In comparison to CellOracle, PMF-GRN demonstrates competitive performance during normal inference and significantly outperforms CellOracle in cross-validation. However, PMF-GRN, in contrast to CellOracle, is not constrained to predicting edges solely within the confines of the prior-knowledge matrix. Furthermore, PMF-GRN consistently outperforms SCENIC across all experiments.

A second key observation is that our approach addresses the high variance associated with heuristic model selection among different inference algorithms. When implementing the Inferelator on \textit{S. cerevisiae} datasets under normal conditions, AUPRCs fall within the range of $0.2$ to $0.4$, showcasing significant variability without a priori information to guide algorithm selection. This diversity among Inferelator algorithms constitutes heuristic model selection, as one cannot predict a priori which algorithm will perform better or discern the reasons behind their divergent performances. In contrast, our method offers reliable results grounded in a principled objective function, delivering competitive performance akin to the best-performing Inferelator algorithm (BBSR) and CellOracle. This underscores the importance of a consistent and robust approach in the face of uncertainty associated with heuristic model selection among disparate algorithms.

To underscore the identifiability issue and affirm the utility of prior-known information, we showcase PMF-GRN's performance when prior information is unused (e.g., all prior logistic normal means of $A$ set to the same low number). This process is replicated for other GRN inference algorithms by providing an empty prior. Additionally, we assess PMF-GRN's performance when prior-known TF-target gene interaction hyperparameters are randomly shuffled before building the prior distribution for $A$. The results, along with those for the Inferelator and CellOracle, indicate the capability of these approaches to accommodate such prior information effectively. 

\subsection{PMF-GRN Provides Well-Calibrated Uncertainty Estimates}

Through our inference procedure, we obtain a posterior variance for each element of $A$, in addition to the posterior mean. We interpret each variance as a proxy for the uncertainty associated with the corresponding posterior point estimate of the relationship between a TF and a gene. Due to our use of variational inference as the inference procedure, our uncertainty estimates are likely to be underestimates. However, these uncertainty estimates still provide useful information as to the confidence the model places in its point estimate for each interaction. We expect posterior estimates associated with lower variances (uncertainties) to be more reliable than those with higher variances.

In order to determine whether this holds for our posterior estimates, we cumulatively bin the posterior means of $A$ according to their variances, from low to high. We then calculate the AUPRC for each bin as shown for the GSE125162 \cite{jackson2020gene} \textit{S.cerevisiae} dataset in Figure \ref{fig:auprc-yeast}D. We observe that the AUPRC decreases as the posterior variance increases. In other words, inferred interactions associated with lower uncertainty are more likely to be accurate than those associated with higher uncertainty. This is in line with our expectations as the more certain the model is about the degree of existence of a regulatory interaction, the more accurate it is likely to be, indicating that our model is well-calibrated. 

\subsection{PMF-GRN Integrates Single Cell Multi-Omic Data for GRN and TFA Inference in Human PBMCs}

We next evaluate PMF-GRN's ability to learn informative GRNs in a human cell line by focusing on Peripheral Blood Mononuclear Cells (PBMCs). PBMCs represent an essential component of the human immune system, and consist of Lymphocytes (CD4 and CD8 T cells, B cells, and Natural Killer cells), Monocytes and Dendritic cells. Unraveling the distinct regulatory landscape of PBMCs is an essential task to provide insight into how these immune cells interact, as well as coordinate to maintain homeostasis and respond effectively to infections. 

To infer an informative and comprehensive PBMC GRN, we harness information from a large, paired single cell RNA and ATAC-seq multi-omic dataset \cite{hao2021integrated}. We adopt a prior-knowledge matrix of TF-target gene interactions (M genes $=18,557$ by K TFs $=860$) as previously constructed by \cite{tjarnberg2023structure} for GRN inference with this multi-omic dataset. In this work, the ATAC-seq data was used as a regulatory mask for ENCODE-derived TF ChIP-seq peaks. Regulatory associations were established through the Inferelator-Prior package based on the proximity of TFs to their potential target genes within 50kb upstream and 2kb downstream of the gene transcription start site. We integrate this prior knowledge with the raw expression profiles of $11,909$ PBMCs from a healthy donor to infer a global PBMC GRN and analyze the TFA profiles of eight annotated cell types and several families of immune TFs within this cell line.

\begin{figure}[htbp]
    \centering
    \includegraphics[width=0.85\textwidth]{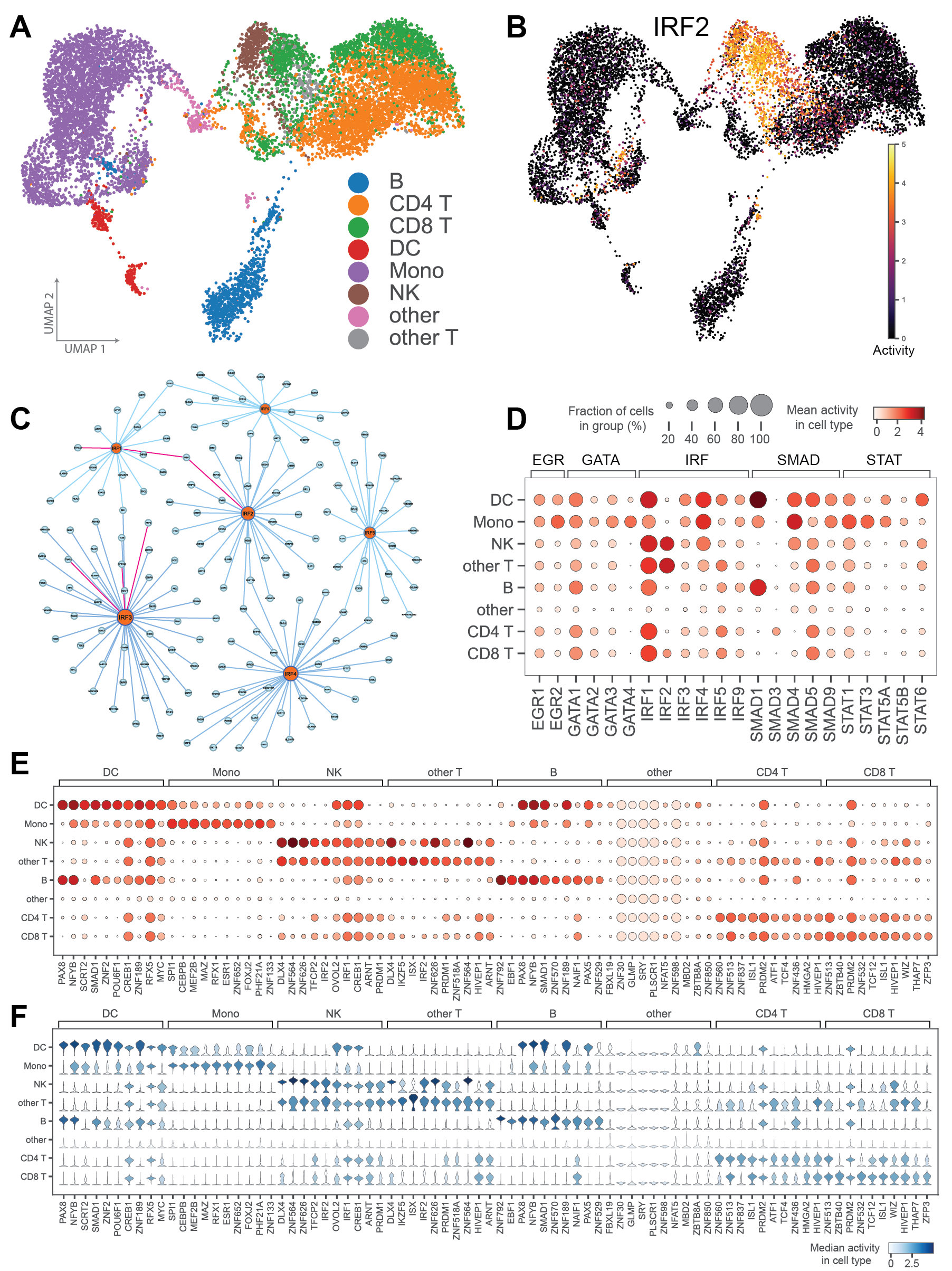}
    \caption{GRN and TFA inference in PBMC. ({\bf A}) UMAP projection of predicted TFA for each annotated PBMC cell type. ({\bf B}) Predicted IRF2 TFA demonstrates high activity in NK and CD8 T cells. ({\bf C}) Heat-map dot-plot depicting TFA of selected immune TFs across annotated PBMC cell types. ({\bf D}) GRN between IRF TFs and their targets. Pink edges indicate literature support for interaction. ({\bf E}) Heat-map dot-plot indicates ten most highly active TFs for each PBMC cell type. ({\bf F}) Violin plot demonstrates corresponding distribution of TFA profiles for ten most highly active TFs.} 
    \label{fig:pbmc-panel}
\end{figure}

We first investigate whether our predicted TFA clusters into distinct cell-type groups, as annotated by \cite{hao2021integrated}. Using UMAP dimensionality reduction, we are able to determine a near clear distinction between each cell type within PBMCs (Figure \ref{fig:pbmc-panel}A). Interestingly, the TFA profiles for each of the T cell sub-types (CD4 T, CD8 T, and other T cells) are closely grouped together, suggesting that these cell types may have a similar lineage or TFA patterns, and may share common transcriptional programs or regulatory networks. 

We next explore the activity profiles of specific immune TF families, starting with the family of TFs belonging to IRF. In PBMCs, IRF contributes to the activation of immune cells that modulate antiviral immunity. Notably, the UMAP projection for IRF2 indicates a high activity pattern within Natural Killer cells and CD8 T cells (Figure \ref{fig:pbmc-panel}B). Indeed, IRF2 is essential for the development and maturation of natural killer cells \cite{persyn2022irf2}, and acts as a CD8 T cell nexus to translate signals from inflammatory tumor microenvironments \cite{lukhele2022transcription}.  

In order to support our predicted TFA for the family of IRF TFs, we additionally investigate the regulatory interactions inferred by PMF-GRN (Figure \ref{fig:pbmc-panel}C). To do this within a reasonable scale, we first threshold our predicted GRN interactions (described in detail in \nameref{sec:pmf-methods}). Within our thresholded GRN, we predict regulatory edges between IRF1 and the target genes B2M and BTN3A1. IF1 has been documented as a transactivator of B2M \cite{gobin2003regulation}, while BTN3A1, a defense-related gene, has been found to be upregulated via the IRF1 pathway \cite{pietz2017immunopathology}. Further, we predict that IRF2 also regulates B2M. Supporting evidence demonstrates that IRF2 has been shown to directly bind to genes linked to the interferon response and MHC Class I antigen presentation, including B2M \cite{mercado2019irf2}. Finally, we predict regulatory edges between IRF3 and GPR108, RNF5, and TRAF2. GPR108 has been shown to be a regulator of type I interferon responses by targeting IRF3 \cite{zhao2023golgi}. Evidence supporting the interaction between IRF3 and RNF5 indicates that RNF5 has an inhibitory effect on the activation of IRF3 \cite{zhong2009ubiquitin}. Lastly, TRAF3 has been shown to be a critical component in the activation of IRF3 during the innate immune response to viral infections \cite{yu2023identification}.

In addition to the IRF TFs, several other families of TFs, such as SMAD, STAT, GATA, and EGR, collectively play pivotal roles in PBMCs. These roles contribute to a wide spectrum of functions, including antiviral responses (IRF), fine-tuning immune responses (SMAD), immune cell development (GATA), immediate early responses to signals (EGR), and central regulation of T cells, B cells, and Natural Killer cells (STAT). Their coordinated activities orchestrate the complex interplay of immune cells, enabling PBMCs to effectively respond to diverse stimuli and maintain immune homeostasis. 

Similarly to IRF, we also explore edges in our thresholded PBMC GRN for these immune TFs to identify regulatory edges supported by literature. Of the five families of immune TFs that we investigate, we find supporting literature for $60$ regulatory edges predicted by PMF-GRN. We provide these literature supported edges, along with their supporting references in Section \nameref{sec:Appendix-pmf-grn} Table \ref{table:pbmc-interactions-1a} and \ref{table:pbmc-interactions-1b}. Additionally, we provide a graph representation of each immune TF GRN in Figure \ref{fig:immune-GRNs}.

We next explore the TFA profiles of each of these immune TFs within the eight PBMC cell types. In Figure \ref{fig:pbmc-panel}D, a heat-map dot-plot provides a visual representation of TFA for each immune TF family across the different PBMC cell types. In particular, we observe that within the IRF family, IRF1 is highly active in CD4 T cells. Previous studies have confirmed the pivotal role of IRF1 in CD4+ T cells, where it is essential for promoting the development of TH1 cells through the activation of the Il12rb1 gene \cite{kano2008contribution}. Additionally, SMAD5 is predicted as highly active in B cells. SMAD5 is a key component of the TGF-$\beta$ signaling pathway, and has been shown to play a crucial role in maintaining immune homeostasis in B cells \cite{malhotra2013smad}. We provide a UMAP of the TFA profiles for each of these immune TFs Section \nameref{sec:Appendix-pmf-grn} Figure \ref{fig:immune-UMAP}.

We further explore our predicted TFA profiles from our global PBMC GRN and calculate the ten most active TFs across the eight distinct cell types. For this experiment, we provide a heat-map dot-plot demonstrating the mean TFA value for each of the top TFs, as well as a corresponding violin plot depicting the distributions of these TFA profiles (Figure \ref{fig:pbmc-panel}E and \ref{fig:pbmc-panel}F). Visualizing these distinct activity profiles provides a concise and informative snapshot of the predominant TFs contributing significant transcriptional activity within each cell population. For example, within B cells we observe high activity for the TF PAX5. PAX5 is known to play a crucial role in B cell development by guiding the commitment of lymphoid progenitors to the B lymphocyte lineage while simultaneously repressing inappropriate genes and activating B lineage-specific genes \cite{cobaleda2007pax5}. 

For each annotated cell-type in the PBMC dataset, a set of marker genes were provided. From our ten most active TFs per cell-type analysis combined with their edges to target genes from our thresholded GRN, we find that several of these TFs are predicted to regulate marker genes. For example, within Dendritic cells, the marker gene HLA-DQA1 is predicted to be regulated by the TFs SMAD1 and RFX5; the marker gene HLA-DPA1 is predicted to be regulated by ZNF2 and RFX5; and the marker gene HLA-DRB1 is predicted to be regulated by RFX5. Within CD4 T cells, the marker gene LTB is predicted to be regulated by the TF ZNF436. Within Natural Killer cells, the marker gene PRF1 is predicted to be regulated by ZNF626. Finally, within B cells, the marker gene BANK1 is predicted to be regulated by the TFs ZNF792, EBF1, PAX8, and PAX5; and the marker gene HLA-DQA1 is predicted to be regulated by the TF SMAD1. 

From the predicted edges between a snapshot of highly active TFs and annotated marker genes, we find the following supporting evidence. The regulatory relationship between RFX5 and HLA-DQA1 involves the inability of RFX5 to bind to the proximal promoter region of HLA-DQA1, potentially due to DNA methylation, hindering the assembly of active regulatory regions \cite{majumder2011dna}. Additionally, EBF1 orchestrates direct transcriptional regulation of BANK1, leading to the observed downregulation of BANK1 expression \cite{treiber2010early}.

Pairing the intensity (dot-plot) with the distribution (violin plot) of TFA offers a comprehensive view of the key TFs guiding our regulatory networks. This approach illuminates the variability in their activity levels across diverse immune cell populations, providing a nuanced understanding of the transcriptional dynamics in PBMCs. This information can be used to guide insights into the functional specialization and diversity of immune cells within PBMCs. Further, this comparison provides a sound starting point for exploring the commonalities and differences in the transcriptional regulation of various immune cell populations.

\subsection{Evaluating PMF-GRN with BEELINE Synthetic Data}

We next evaluated PMF-GRN using synthetic datasets curated from the BEELINE benchmark \cite{pratapa2020benchmarking}. This benchmark provides six synthetic networks, linear (LI), linear long (LL), cycle (CY), bifurcating (BF), trifurcating (TF), and bifurcating converging (BFC). In repetitions of ten, expression datasets of increasing cell sizes (e.g., $n=100, 200, 500, 2000$ and $5000$) were generated by sampling. Using these generated expression datasets, as well as the provided reference GRNs, we inferred $300$ GRNs using PMF-GRN (Figure \ref{fig:auprc-beeline}A). For each of the six synthetic datasets, PMF-GRN outperforms the BEELINE baseline, represented in Figure \ref{fig:auprc-beeline}A with a black dashed line. 

\begin{figure}[htbp]
    \centering
    \includegraphics[width=0.95\textwidth]{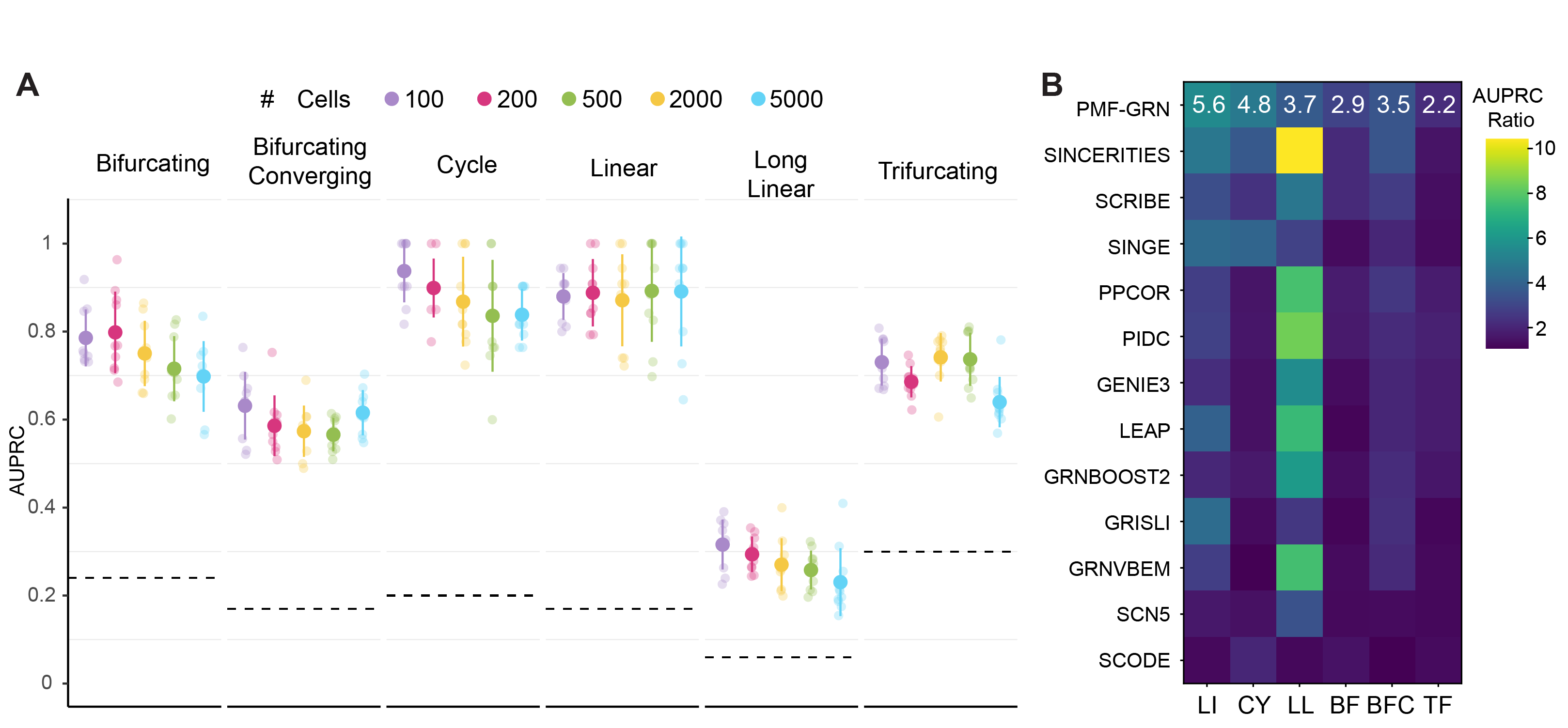}
    \caption{PMF-GRN performance on BEELINE synthetic GRN data ({\bf A}) PMF-GRN inference performance with half of the ground truth provided as prior network information and the remaining half provided as a gold standard for evaluation. Dashed lines are the expected baseline of a random predictor. ({\bf B}) AUPRC ratio over the baseline random predictor for PMF-GRN in comparison to each of the GRN inference methods used in the original BEELINE benchmark.}
    \label{fig:auprc-beeline}
\end{figure}

To further evaluate PMF-GRN, we calculate the AUPRC ratio of PMF-GRN over the baseline random predictor to compare to the similarly computed ratios in the original BEELINE paper (Figure \ref{fig:auprc-beeline}B). We observe that for the linear, cycle, and bifurcating converging, PMF-GRN achieves competitive AUPRC ratios in comparison to the original methods used in the BEELINE benchmark. Interestingly, PMF-GRN does not perform competitively on long linear. This could be due to a number of factors, such as the larger number of intermediate genes introducing additional complexity which PMF-GRN struggles to capture. Alternatively, the extended trajectory introduces a higher-dimensional space, which could present a challenge for our matrix factorization based approach to effectively decompose the data into meaningful latent factors. This presents an interesting avenue of consideration when developing future probabilistic matrix factorization approaches for GRN inference. 

\section{Methods}
\label{sec:pmf-methods}

\subsection{Model Details}

We index cells, genes and TFs using $n \in \{ 1, \cdots, N \}$, $m \in \{ 1, \cdots, M \}$ and $k \in \{ 1, \cdots, K \}$, respectively. We treat each cell's expression profile $W_n$ as a random variable, with local latent variables $U_n$ and $d_n$, and global latent variables (that are shared among all cells) $\sigma_{obs}$ and $V = A \odot B$. We use the following likelihood for each of our observations:
\begin{equation*}
    p_W(W_n \mid U,V, \sigma_{obs}, d) = \mathcal{N}(d_n * U_n V^\top, \sigma_{obs}^2).
\end{equation*}
We assume that $U$, $A$, $B$, $\sigma_{obs}$ and $d$ are independent i.e., 
\begin{equation*}
    p(U, A, B, \sigma_{obs}, d) = p_U(U)p_A(A)p_B(B)p_{\sigma}(\sigma_{obs})p_d(d).
\end{equation*} 
In addition to our i.i.d assumption over the rows of $U$  and $d$, we also assume that the entries of $U_n$ are mutually independent, and that all entries of $A$ and $B$ are mutually independent. We choose a lognormal distribution for our prior over $U$ and a logistic Normal distribution for our prior over $d$:
\begin{equation*}
    p_U(\log(U_{nk})) = \mathcal{N}(\mu_u, \sigma_u^2),
\end{equation*}
\begin{equation*}
    p_d(\text{logit}(d_n)) = \mathcal{N}(0, 9)
\end{equation*}
where $\mu_u \in \mathbb{R}$ and $\sigma_u \in \mathbb{R}^+$.

We use a logistic Normal distribution for our prior over $A$, a Normal distribution for our prior over $B$ and a logistic Normal distribution for our prior over $\sigma_{obs}$:
\begin{equation*}
    p_A(\text{logit}(A_{mk})) = \mathcal{N}(\text{logit}(\text{clip}(\bar{A}_{mk}, a_{\max}, a_{\min}) ), \sigma_a^2),
\end{equation*}
\begin{equation*}
    p_B(B_{mk}) = \mathcal{N}(0, \sigma_b^2).
\end{equation*}
\begin{equation*}
    p_{\sigma}(\log(\sigma_{obs})) = \mathcal{N}(0, 1),
\end{equation*}
where
\begin{align*}
    \bar{A}_{mk} &\in \{0, 1\}, \\
    a_{\max}, a_{\min} &\in (0, 1), \\
    \sigma_a, \sigma_b &\in \mathbb{R}_{>0},
\end{align*}
and
\begin{align*}
    \operatorname{clip}(\bar{A}_{mk}, a_{\max}, a_{\min})
    = \max\bigl(\min(\bar{A}_{mk}, a_{\max}), a_{\min}\bigr).
\end{align*}
Here, $\bar{A}_{mk}$ is provided by a prior-knowledge pipeline also used by methods such as the Inferelator. The pipeline leverages ATAC-seq and TF binding motif data to provide binary initial guesses of gene-TF interactions. $a_{\max}$ and $a_{\min}$ are hyperparameters that determine how we clip these binary values before transforming them to the logit space.

For our approximate posterior distribution, we enforce independence as follows:
\begin{equation*}
q(U, A, B, \sigma_{obs}, d) = q_U(U)q_A(A)q_B(B)q_{\sigma}(\sigma_{obs})q_d(d).
\end{equation*}
We impose the same independence assumptions on each approximate posterior as we do for its corresponding prior. Specifically, we use the following distributions:
\begin{equation*}
    q_U(\log(U_{nk})) = \mathcal{N}(\tilde{U}_{nk}, \tilde{\sigma}_{U_{nk}}^2)
\end{equation*}
\begin{equation*}
    q_d(\text{logit}(d_n)) = \mathcal{N}(\tilde{d}_n, \tilde{\sigma}_{d_n}^2)
\end{equation*}
\begin{equation*}
    q_A(\text{logit}(A_{mk})) = \mathcal{N}(\tilde{A}_{mk}, \tilde{\sigma}_{A_{mk}}^2)
\end{equation*}
\begin{equation*}
    q_B(B_{mk}) = \mathcal{N}(\tilde{B}_{mk}, \tilde{\sigma}_{B_{mk}}^2)
\end{equation*}
\begin{equation*}
    q_{\sigma}(\log(\sigma_{obs})) = \mathcal{N}(\tilde{o}, \tilde{\sigma}_o^2),
\end{equation*}
where the parameters on the right hand sides of the equations are called variational parameters:
$\tilde{U}_{nk}$, $\tilde{d}_n$, $\tilde{A}_{mk}$, $\tilde{B}_{mk}$, $\tilde{o}$ $\in \mathbb{R}$ and $\tilde{\sigma}_{U_{nk}}$, $\tilde{\sigma}_{d_n}$, $\tilde{\sigma}_{A_{mk}}$, $\tilde{\sigma}_{B_{mk}}$, $\tilde{\sigma}_o$ $\in \mathbb{R^+}$. To avoid numerical issues during optimization, we place constraints on several of these variational parameters.

\subsection{Inference}

We perform inference on our model by optimizing the variational parameters to maximize the ELBo. In doing so, we minimise the KL-divergence between the true posterior and the variational posterior. In practice, to help with addressing the latent factor identifiability issue, we use a modified version of the ELBo where the prior and posterior terms are weighted by a constant $\beta \geq 1$ \cite{higgins2017beta}:
\begin{align*}
    \mathbb{E}_{U, A, B, \sigma_{obs}, d \sim q(U, A, B, \sigma_{obs}, d)} [ & \log p(W|U, V = A \odot B,  \sigma_{obs}, d)\\
    & + \beta ( \log p(U, A, B, \sigma_{obs}, d) - \log q(U, A, B, \sigma_{obs}, d) ) ]
\end{align*}
Inference is carried out using the Adam optimizer with learning rate 0.1 and beta values of 0.9 and 0.99. We clip gradient norms at a value of 0.0001. We set $a_{\min}=0.005$, $a_{\max}=0.995$, $\sigma_b^2 = 1$ and $\mu_u = 0$. We vary $\sigma_a$ and $\sigma_u$ as hyperparameters that control the strengths of the priors over $A$ and $U$, respectively. We also vary $\beta$ as a hyperparameter.

We choose a hyperparameter configuration using validation AUPRC as the objective function as well as the early stopping metric. We hold out hyperparameters for $p(A)$ for a fraction of the genes. We do this by setting $\bar{A}_{mk} = 0$ for $m$ corresponding to these genes for all $k$. During inference we regularly obtain posterior point estimates for these entries and measure the AUPRC against the original values of these entries as given in the full prior. This quantity is known as the validation AUPRC.

Once we have picked the hyperparameter configuration corresponding to the best validation AUPRC, we perform inference with this model using the full prior without holding out any information. We use an importance weighted estimate of the marginal log likelihood as our early stopping criterion:
\begin{align*}
    \log p(W) &= \log \left( \mathbb{E}_{U, A, B, \sigma_{obs}, d \sim q(U, A, B, \sigma_{obs}, d)} \left[ \frac{p(W|U, A, B,  \sigma_{obs}, d) p(U, A, B, \sigma_{obs}, d)}{q(U, A, B, \sigma_{obs}, d)}\right] \right),
\end{align*}
where the expectation is computed using simple Monte Carlo and the $\log$-$\sum$-$\exp$ trick is used to avoid numerical issues.

\subsection{Computing Summary Statistics for the Posterior}

After training the model, we use $\tilde{A}$ and $\tilde{\sigma}_{A}$, the variational parameters of $q(A)$, to obtain a mean and a variance for each entry of $A$. Since $q(A)$ is logistic normal, it admits no closed form solution for the mean and variance. We therefore use Simple Monte Carlo i.e. we sample each entry of $A$ several times from its posterior distribution and then compute the sample mean and sample variance from these samples. We use each mean as a posterior point estimate of the probability of interaction between a TF and a gene, and its associated variance as a proxy for the uncertainty associated with this estimate.

\subsection{Calculating AUPRC}
\label{subsec:calc-auprc}

The gold standards for the datasets used in this paper do not necessarily perfectly overlap with the genes and TFs that make up the rows and columns of $A$ as defined by the prior hyperparameters i.e. there may be genes and TFs in the gold standard with a recorded interaction or lack of interaction, that do not appear in our model at all because they are not present in the prior. The reverse is also true: the prior may contain genes and TFs that are not in the gold standard. For this reason, we compute the AUPRC using one of two methods: `keep all gold standard' or `overlap', which correspond to evaluating only interactions that are present in the gold standard or only interactions that are present in both the gold standard and the prior/posterior. We present results with `keep all gold standard' AUPRC as the evaluation metric when comparing our model to the Inferelator in Figure \ref{fig:auprc-yeast}. For our evaluation of uncertainty calibration (Figure \ref{fig:auprc-yeast}D), we use the overlap AUPRC so that bins containing a lower number of posterior means do not have artificially deflated AUPRCs (see the \nameref{sec:methods-calibration} part of the \nameref{sec:pmf-methods} Section for further information).

\subsection{Evaluating Calibration of Posterior Uncertainty}
\label{sec:methods-calibration}
We create 10 bins, corresponding to the lowest 10\%, 20\%, 30\% and so on of posterior variances. We place the posterior point estimates of TF-gene interactions associated with these variances into these bins and then calculate the `overlap AUPRC' for each bin using the corresponding gold standard. The AUPRC for each bin is calculated using those interactions that are in the gold standard and also in the bin. We use such a cumulative binning scheme because using a non-cumulative scheme could result in some bins having very small numbers of posterior interactions that are present in the gold standard, which would lead to noisier estimates of the AUPRC.

\subsection{Inference and Evaluation on Multiple Observations of W}
The Inferelator method applies two scRNA-seq experiments separately on \textit{S. cerevisiae}, with each resulting in a distinct model. These models are used to infer TF-gene interaction matrices, which are then sparsified. The final matrix is obtained by taking the intersection of the two matrices and retaining only the entries that are non-zero in both matrices. In our approach, we also train a separate model on each expression matrix, and obtain a posterior mean matrix for $A$ for each of them. To obtain the final posterior mean matrix for $A$, we average the posterior mean matrices from each model. While this approach works well, future research could focus on explicitly modeling separate expression matrices within the model, as mentioned in the \nameref{sec:disc} section.

\subsection{Measuring the Impact of Prior Hyperparameters}
We evaluate the utility of each of the prior hyperparameter matrices used in our experiments. In Figures \ref{fig:auprc-yeast}A and Section \nameref{sec:Appendix-pmf-grn} Figure \ref{fig:auprc-bsubtilis_normalization}, we present with grey dots the AUPRCs achieved when performing inference using shuffled prior hyperparameters for $A$. This corresponds to randomly assigning to each row (gene) of $A$, the prior hyperparameters that correspond to a different row of $A$. Shuffling the hyperparameters should lead to worse performance, as the posterior estimates should then also be shuffled, whereas the row/column labels for the posterior will remain unshuffled. For the `no prior' setting, shown with black dots in the figures, we set $\bar{A}_{mk} = 0 \  \forall \ m, k$. The difference in AUPRC achieved using the unshuffled vs shuffled or no hyperparameters measures the usefulness of the provided hyperparameters for the inference task on the dataset in question.

\subsection{Cross-Validation}
For \textit{S. cerevisiae}, we perform a five-fold cross validation experiment (Figure \ref{fig:auprc-yeast}B). Cross-validation is performed by partitioning the gold standard into an 80\% - 20\% split, where 80\% of the data represents prior-known information to be used as a prior for $p(A)$, and the remaining 20\% is treated as the gold standard for evaluation. This process is repeated five times to generate five random splits of the data in order to robustly evaluate GRN inference. It is important to note that PMF-GRN performs hyperparameter search before inferring a final GRN within each cross-validation split. For each of the five partitioned cross-validation folds the 80\%, or prior portion, is further split into 80\% train and 20\% test for hyperparameter search and evaluation. Once the optimal hyperparameters have been determined, the initial 80\% split is treated as the training data, while the remaining 20\%, which was not seen during hyperparameter selection, is used for evaluation.

\subsection{Intersection over Union}
Intersection over Union (IoU) scores were computed using the GRN learned by each algorithm for the two \textit{S. cerevisiae} expression datasets. For each GRN, we calculate and retain the top $25\%$ of predicted edges in order to obtain the best estimates for each algorithm and elimate noisier predictions. For each algorithm, we compute both the intersection and the union of the GRN interactions predicted from the two \textit{S. cerevisiae} datasets. Dividing the Intersection by the Union allows us to obtain a score indicating how similar the two inferred GRNs are for each algorithm.

\subsection{Downsampling Expression}
For \textit{S. cerevisiae}, in repetitions of five, we randomly sample the \textit{S. cerevisiae} expression matrix on the cell axis to obtain downsampled expression dataset sizes of $80\%$, $60\%$, $40\%$, and $20\%$. We perform a hyperparameter search, using an $80\%$ training - $20\%$ validation split of the prior-knowledge matrix, on each of these five expression matrices for each sample size. Using these hyperparameters, we infer GRNs for each repetition within each split to obtain our final downsampled GRNs.

\subsection{Exploring the Effect of Cross-Validation Ratios on Hyperparamater Selection}
To effectively explore the influence of cross-validation split size on obtaining optimal hyperparameters for GRN inference, we methodically separate our \textit{S. cerevisiae} prior-knowledge into 4 different split sizes. These splits consist of $80\%$ training - $20\%$ validation, $60\%$ training - $40\%$ validation, $40\%$ training - $60\%$ validation, and $20\%$ training - $80\%$ validation. For each split size, we obtain 5 random training and validation splits to ensure robust results. We then perform hyperparameter search across each 5 random splits for each split size. Using the best overall hyperparameters for each split size, we infer a final GRN to demonstrate the impact each particular split had on obtaining the optimal hyperparameters for the final GRN.

\subsection{Datasets and Preprocessing}

We inferred each GRN using a single-cell RNA-seq expression matrix, a TF-target gene connectivity matrix, and a gold standard for bench-marking purposes. We modeled the single-cell expression matrices based on the raw UMI counts obtained from sequencing for the \textit{S. cerevisiae} and PBMC datasets, which were therefore not normalized for the purpose of this work. For the two \textit{B. subtilis} datasets used in this work, we demonstrate the effect of different normalization and scaling techniques, and convert all data used to integers in order to create a single-cell-like dataset. We further obtained binary TF-gene matrices representing prior-known interactions, which served as prior hyperparameters over {\bf A}, and were derived from the YEASTRACT and subtiwiki databases, as well as from \cite{tjarnberg2023structure} for PBMC. We acquired a gold standard for \textit{S. cerevisiae} our datasets from independent work which is detailed below.

\subsubsection{Saccharomyces cerevisiae}

We used two raw UMI count expression matrices for the organism \textit{S. cerevisiae} obtained from NCBI GEO (GSE125162 \cite{jackson2020gene} and GSE144820 \cite{jariani2020new}). For this well studied organism, we employed the YEASTRACT \cite{monteiro2020yeastract+, teixeira2018yeastract} literature derived network of TF-target gene interactions to be used as a prior over $A$ in both \textit{S. cerevisiae} networks. A gold standard for \textit{S. cerevisiae} was additionally obtained from a previously defined network \cite{tchourine2018condition} and used for bench-marking our posterior network predictions.
We note that the gold standard is roughly a reliable subset of the YEASTRACT prior. Additional interactions in the prior can still be considered to be true but have less supportive evidence than those in the gold standard.

\subsubsection{Peripheral Blood Mononuclear Cells}
\label{pbmc-processing}
We used a paired multi-omic single cell RNA-seq and ATAC-seq dataset for PBMC obtained from \cite{hao2021integrated}. The single-cell expression matrix contained 11,909 cells. The prior-knowledge matrix was constructed using the ATAC-seq data from this multi-omic dataset, constructed and described in detail by \cite{tjarnberg2023structure}. The prior-knowledge matrix is $18,557$ genes by $860$ TFs, and contains $0.5\% $ non-zero edges.

Due to the complex and dynamic nature of PBMCs, a gold standard is currently unavailable for this cell line. To evaluate our inferred network, we implement a 5-fold cross-validation procedure where our chromatin accessibility-based prior is split into 5 random sets, where $80\%$ is used as prior knowledge and $20\%$ is used as the gold standard for evaluation. We then took the intersection of the regulatory edges inferred across each of the 5 fold cross-validation experiments, and filtered to retain the highest quality edges, obtaining a prediction probability of $90\%$ or higher. 

\subsubsection{BEELINE Synthetic Datasets}
\label{BEELINE-processing}
We used the BEELINE synthetic expression datasets \cite{pratapa2020benchmarking} without modification. Reference GRNs were transformed into cross-tab matrices in order to use this information for prior-knowledge and gold standard evaluation. We used $50\%$ of the reference GRN as the prior and the remaining $50\%$ as the gold standard, as was similarly done in \cite{skok2022high}.

\section{Discussion}
\label{sec:disc}
In this chapter, we introduce a robust framework for probabilistic matrix factorization, optimized through automatic variational inference, to infer GRNs from single-cell gene expression data. A distinctive feature of our approach is the decoupling of the data generation model from the inference procedure, providing unprecedented flexibility. This decoupling allows for modifications to the latent variables and their distributions, without altering the inference process. Such flexibility facilitates the seamless integration of diverse sequencing datasets and modeling assumptions. Unlike previous methods, our framework eliminates the need to define a new inference procedure for each specific dataset or biological context when building new models. 

PMF-GRN not only offers a flexible and unified approach to GRN inference but also provides a principled methodology for model selection and hyperparameter configuration. The use of a consistent objective function and inference procedure across all generative models streamlines the process of hyperparameter search, reducing ambiguity present in methods like the Inferelator. By conducting hyperparameter search across different generative models, we identify configurations corresponding to optimal values of our objective function, minimizing the reliance on heuristic model selection.

To validate the effectiveness of our approach, we applied PMF-GRN to infer GRNs from single cell \textit{S. cerevisiae} gene expression, comparing results with state-of-the-art single cell GRN inference methods such as the Inferelator, SCENIC and CellOracle. Our method demonstrates competitive, if not superior, performance in terms of AUPRC, in each experiment performed. Here, PMF-GRN provides a stable and reliable inferred GRN without the need for heuristic model selection or data separation into tasks. 

Cross-validation experiments further support the robustness of PMF-GRN, BBSR, and StARS, indicating their ability to generalize well to new data without overfitting. In contrast, SCENIC and CellOracle exhibited poor performance during cross-validation, suggesting potential issues with generalizability. Notably, we assessed the robustness of each algorithm against increasing noise in the prior-knowledge, identifying PMF-GRN and CellOracle as the most resilient to noisy priors. This resilience ensures the reliability of inferred GRNs even in the presence of uncertain prior knowledge.  

Our model uniquely provides well-calibrated uncertainty estimates alongside point estimates for each interaction in the final GRN. The evaluation of uncertainty estimates demonstrated that as the posterior variance decreases, the AUPRC increases, indicating that the model is well-calibrated. Biologists can leverage these uncertainty estimates for downstream experimental validation, placing more trust in estimates with lower posterior variance. Finally, the linear scalability of our models computational cost with the number of cells enables its application to single-cell RNA-seq datasets of any size.

Our investigation into PMF-GRN's application to human PBMCs provides insightful findings into the regulatory landscape of these essential immune cells. Leveraging a comprehensive multi-omic dataset, we demonstrate that our approach integrates single cell RNA and well-curated prior knowledge derived from ATAC-seq data. The resulting global PBMC GRN unveils distinct TFA profiles for eight annotated cell types and various immune TF families. Through UMAP dimensionality reduction, we observe clear clustering of TFA profiles. Focusing on the IRF family, we identify specific TF-target gene interactions supported by literature, shedding light on regulatory relationships critical for immune responses. Extension to other immune TF families reveals their orchestrated activities within PBMCs, contributing to antiviral responses, immune cell development, and the regulation of T cells, B cells, and Natural Killer cells. By exploring predicted edges between active TFs and marker genes, we establish connections between regulatory networks and cellular functions. The combined dot-plot and violin plot visualization strategy provides a nuanced understanding of TF activities, offering a valuable resource for deciphering the intricate transcriptional dynamics in PBMCs. This detailed exploration sets the stage for further investigations into the functional specialization and diversity of immune cells within the PBMC population, with implications for advancing our understanding of immune responses and disease mechanisms.

In the context of synthetic datasets curated from the BEELINE benchmark, PMF-GRN demonstrates robust performance across various network structures. Outperforming the BEELINE baseline across different synthetic networks, PMF-GRN consistently achieves competitive AUPRC ratios compared to the original methods used in the BEELINE benchmark. Notably, PMF-GRN's competitive performance is observed in linear, cycle, and bifurcating converging structures. However, challenges arise in the long linear structured synthetic data, suggesting potential limitations in capturing the complex dynamics of extended trajectories. Factors such as the increased number of intermediate genes and a higher-dimensional space may contribute to this limitation. This observation opens avenues for future development of probabilistic matrix factorization approaches, encouraging exploration of methods better suited for intricate network structures. The overall success of PMF-GRN in diverse synthetic network scenarios underscores its versatility and effectiveness in inferring GRNs, promising broad applicability in deciphering complex biological systems and regulatory interactions.

\section{Conclusion}
\label{sec:pmf-conclusion}

In conclusion, the PMF-GRN framework provides a flexible and principled approach for inferring GRNs from single-cell gene expression data. By decoupling the model and inference procedure, PMF-GRN enables easy integration of new and various sequencing datasets as well as modeling assumptions without the need for defining a new inference procedure. Additionally, PMF-GRN provides a principled approach for model selection through hyperparameter search, reducing the need for heuristic model selection. Overall, PMF-GRN consistently yields high-performing competitive results compared to other state-of-the-art single cell GRN inference methods with a reliable gold standard, and is robust to cross validation, noisy priors and downsampling. Further, PMF-GRN provides well-calibrated uncertainty estimation, enabling a reliable set of results for downstream experimental validation. 

We envision many possible directions for future work to design a better algorithm for inferring GRNs under our framework. This framework could be extended to explicitly model multiple expression matrices and their batch effects. We could probabilistically model prior information for $A$ obtained from ATAC-seq and TF motif databases, and include this as part of the probabilistic model over which we carry out inference. Evaluating the posterior estimates of the direction of transcriptional regulation, provided by the matrix $B$, could provide a useful benchmark for the computational estimation of TF activation and repression. Research could also be carried out on improved self-supervised objectives for hyperparameter selection.

Future work could also focus on how to use results from our framework to guide experimental wet-lab work. For example, the uncertainty quantification provided by our model could open up new research directions in active learning for GRN inference. Highly ranked, uncertain interactions could be experimentally tested and the results fed back into the prior hyperparameter matrix for $A$. Inference with this updated matrix would ideally yield a better posterior GRN estimate. Posterior estimates of TFA provided by our model could be useful to wet lab scientists, as this quantity provides information about possible post-transcriptional modifications, which are currently challenging to measure experimentally.

Most importantly, the study of GRN inference is far from complete. GRN inference approaches have thus far required new computational models and assumptions in order to keep up with relevant sequencing technologies. It is thus essential to develop a model that can be easily adapted to new biological datasets as they become available, without having to completely re-build each model. We have therefore proposed PMF-GRN as a modular, principled, probabilistic approach that can be easily adapted to both new and different biological data without having to design a new GRN inference method.

\section{Retrospective}
\paragraph{Perspective since publication in 2024}\mbox{}\\
\label{sec:pmfgrn-retrospective}

PMF-GRN was developed under the pragmatic premise that single-cell expression measurements are informative about regulation, yet they are only indirectly related to the regulatory mechanisms of interest. The PMF-GRN model therefore treats TF activity and TF-gene regulatory effects as latent quantities, and relies on prior knowledge to anchor these latent factors to biologically interpretable TF identities. With the benefit of subsequent work in this thesis, the strongest and most durable contribution of PMF-GRN is not a single likelihood choice or factorization variant, but the framing of GRN inference as a modular probabilistic program in which prior information, uncertainty, and model comparison are explicit, integral components of the modeling framework. The decoupling of the generative model from an automatic variational inference procedure remains the core design decision that enables extensibility, principled hyperparameter selection, and uncertainty estimates at the level of individual TF-gene interactions.

Several developments since PMF-GRN's publication in 2024 have clarified how this framing fits into the broader trajectory of the field. Evaluation has shifted further toward perturbation-grounded benchmarks, motivated by the long recognized limitation that curated reference networks are incomplete, context-dependent, and often aggregate heterogeneous evidence. Resources such as CausalBench \cite{chevalley2025large} explicitly foreground single-cell perturbation data as a basis for assessing network inference methods in settings that are closer to causal intervention than correlation-based reconstruction. This shift does not diminish the role of probabilistic modeling; rather, it strengthens the case for models that expose calibrated uncertainty and make their assumptions explicit, as perturbation benchmarks often highlight failure modes that are difficult to detect when evaluation relies only on incomplete reference networks. 

At the same time, GRN inference has continued moving toward multi-omic formulations in which chromatin accessibility is not merely an optional source of prior edges but part of the primary observational data \cite{yuan2025inferring}. Methods designed around paired RNA and ATAC measurements, including neural approaches that integrate atlas-scale external data, reflect an emerging default, where regulatory network reconstruction increasingly relies on combining transcriptional state with information about regulatory potential and binding. In retrospect, PMF-GRN's emphasis on priors derived from chromatin accessibility anticipated this direction, while also making a clear limitation that remains unresolved in many pipelines: accessibility and motif evidence constrain what regulation is plausible, but they do not uniquely determine functional influence on transcription. A probabilistic treatment of prior evidence, including uncertainty in peak-to-gene linking, motif matches, and cell-type specificity, is therefore an important next step for reconciling multi-omic evidence with expression-driven inference. 

A third development is the rapid growth of representation-learning approaches for cellular systems. Large cell models, trained on tens of millions of cells, are now explicitly positioned as tools for robust gene network prediction \cite{kalfon2025scprint}. These cell models suggest that a meaningful fraction of regulatory structure can be learned from scale even before introducing experiment-specific priors or mechanistic assumptions. This trend reframes the role of probabilistic GRN inference. One plausible synthesis is that foundation models provide strong, transferable representations or priors, while probabilistic inference provides the mechanism to $(i)$ adapt those priors to a specific dataset, and $(ii)$ quantify what the data can and cannot resolve about regulatory structure. This perspective aligns naturally with the thesis-wide narrative that follows PMF-GRN: improving the quality and transferability of priors can shift the effective bottleneck in GRN reconstruction, while uncertainty estimates remain central for deciding which predicted interactions warrant experimental follow-up.

These changes also sharpen several limitations of PMF-GRN that are important to acknowledge explicitly in this thesis. First, the interpretability of latent factors depends on the availability and quality of prior knowledge, and this dependence is not a defect so much as a statement of identifiability. Expression alone often under-specifies which TF is responsible for a regulatory program, especially when TFs are correlated, when regulation is combinatorial, or when regulatory signals precede the moment of capture. Second, the uncertainty estimates produced by variational inference should be treated as conservative in their scope: they are most reliable as a relative ranking of confidence across edges, while absolute calibration can be affected by posterior approximation. Finally, the BEELINE results suggest regimes where the modeling assumptions are strained (for example, extended trajectories with many intermediate genes), motivating extensions that more directly represent dynamics, time-lagged effects, or state transitions when such information is available.

Viewed from the present, PMF-GRN serves as a foundation in two ways. Methodologically, it establishes a probabilistic scaffold for GRN inference with explicit priors, model comparison, and edge-wise uncertainty. Conceptually, it motivates this thesis's later emphasis on prior construction and transfer-learning, where if expression data primarily refines, contextualizes, and prunes a regulatory scaffold, then learning that scaffold well, either from sequence and other scalable modalities, becomes the central objective. The chapter that follows builds directly on this premise by focusing on how to construct stronger priors and how to integrate them into inference pipelines in ways that preserve interpretability and make uncertainty actionable. 

\section{Supplementary Material for Chapter \ref{chp-1}}
\label{sec:Appendix-pmf-grn}

\subsection{PMF-GRN Recovers True Interactions in Prokaryotes as Evaluated by Cross-Validation}

To demonstrate GRN inference on a forth additional dataset, we carry out experiments using two microarray datasets for the prokaryote \textit{Bacillus Subtilis} (B1 - GSE27219 \cite{nicolas2012condition} and B2 - GSE67023 \cite{arrieta2015experimentally}). Although PMF-GRN is not primarily designed to learn GRNs from microarray data, we show that it is still possible to learn informative GRNs with this data. For our \textit{B. subtilis} experiments, we have access to prior-knowledge derived from the subtiwiki database \cite{michna2016subti, zhu2018subti, pedreira2022current}. Here, we implement a 5 fold cross-validation approach by using five random splits of the subtiwiki database-derived information, where $80\%$ is used as prior knowledge and $20\%$ is used as the gold standard for evaluation.

The two \textit{B. subtilis} datasets were previously normalized after data collection as part of standard microarray processing. However, each dataset was normalized using different approaches (described in \nameref{sec:pmf-methods}). For B1, the expression data underwent no further normalization and was simply converted to integers to simulate single-cell-like data. For B2, the expression data was re-scaled and then converted to integers, in order to contain only positive integers resembling single-cell-like data. The results from our experiments are shown in Figure \ref{fig:auprc-bsubtilis_normalization}, and the numbers used to create this figure are given in Table \ref{table:bsubtilis-B1-auprc} and \ref{table:bsubtilis-B2-auprc}. Using five repeats of cross-validation, we show the performance of GRNs inferred for the two \textit{B. subtilis} datasets (B1 and B2). We remark that the difference in performance between B1 and B2 is likely a result of the chosen microarray processing normalization. To further support this claim, we demonstrate GRN performance after re-scaling the data with min-max scaling (Figure \ref{fig:auprc-bsubtilis_normalization}).

`No Prior' and `Shuffled' results are also shown in Figure~\ref{fig:auprc-bsubtilis_normalization} by black and gray dots respectively. Here, we are able to demonstrate that for B1 and B2, each GRN yields a better performance as compared to negative controls. 

\begin{figure}[!htbp]
    \renewcommand{\figurename}{Figure}
    \setcounter{figure}{0}
    \centering
    \includegraphics[width=0.95\textwidth]{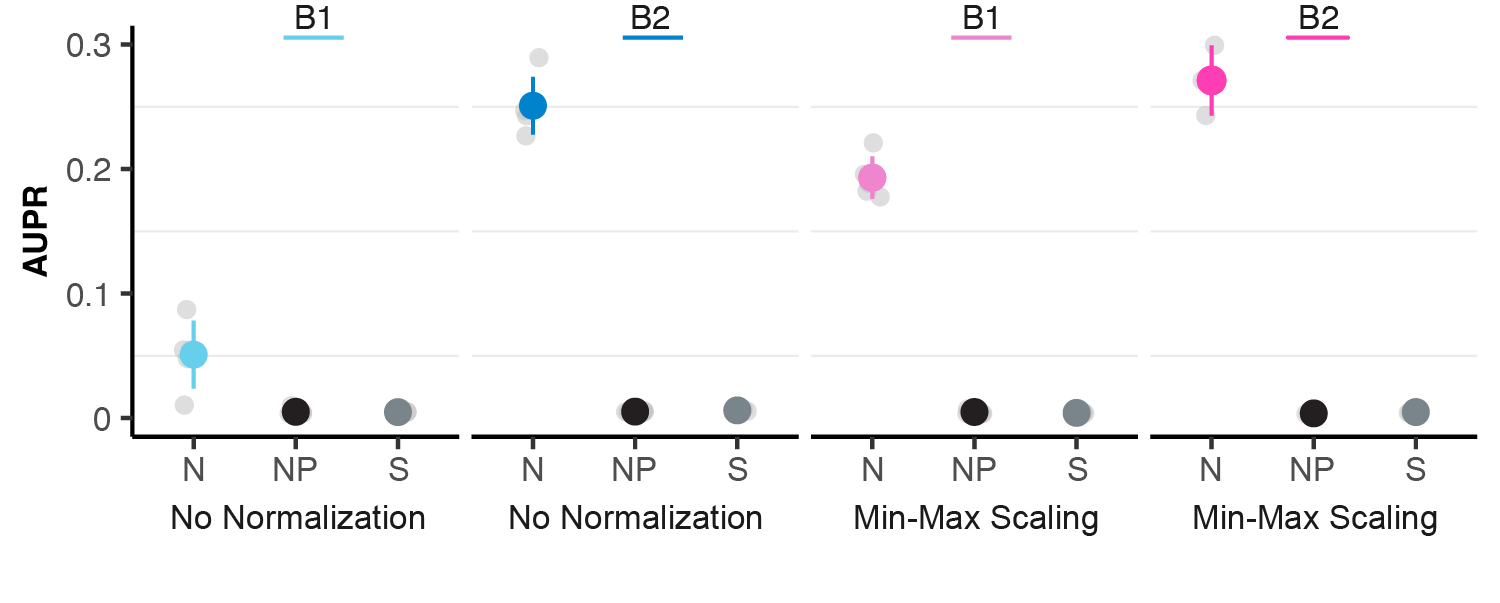}
    \caption{Results for GRNs learned in \textit{B. subtilis} datasets B1 (GSE27219) and B2 (GSE67023) comparing "No Normalization" to "Min-Max Scaling". Colored dots represent the normal (N) GRN, with the line indicating the mean of the cross-validation experiments $\pm$ standard deviation. Negative controls are demonstrated by black dots for GRNs inferred with No Prior (NP) and grey dots for Shuffled Prior (S).}
    \label{fig:auprc-bsubtilis_normalization}
\end{figure}

\begin{table}[!htbp]
    \centering
    \renewcommand{\arraystretch}{1.2}  
    \begin{tabular}{l|ccc}
        \multicolumn{1}{c}{} & \multicolumn{3}{c}{\textbf{\textit{B. subtilis} Cross-Validation Dataset B1}} \\ \cline{2-4}
        \textbf{Method} & \textbf{Regular} & \textbf{No Prior} & \textbf{Shuffled} \\ \hline
        No Normalization & $0.0509 \pm 0.0273$ & $0.0048 \pm 0.0003$ & $0.0050 \pm 0.0028$ \\
        Min-Max Scaling  & $0.1931 \pm 0.0171$ & $0.0042 \pm 0.0003$ & $0.0042 \pm 0.0018$ \\
    \end{tabular}
    \caption{AUPRCs achieved by PMF-GRN on the \textit{B. subtilis} B1 dataset. Results are reported as the mean AUPRC across five cross-validation splits $\pm$ standard deviation.}
    \label{table:bsubtilis-B1-auprc}
\end{table}

\begin{table}[!htbp]
    \centering
    \renewcommand{\arraystretch}{1.2}  
    \begin{tabular}{l|ccc}
        \multicolumn{1}{c}{} & \multicolumn{3}{c}{\textbf{\textit{B. subtilis} Cross-Validation Dataset B2}} \\ \cline{2-4}
        \textbf{Method} & \textbf{Regular} & \textbf{No Prior} & \textbf{Shuffled} \\ \hline
        No Normalization & $0.2508 \pm 0.0232$ & $0.0062 \pm 0.0006$ & $0.0052 \pm 0.0004$ \\
        Min-Max Scaling  & $0.2886 \pm 0.0312$ & $0.0048 \pm 0.0008$ & $0.0038 \pm 0.0005$ \\
    \end{tabular}
    \caption{AUPRCs achieved by PMF-GRN on the \textit{B. subtilis} B2 dataset. Results are reported as the mean AUPRC across five cross-validation splits $\pm$ standard deviation.}
    \label{table:bsubtilis-B2-auprc}
\end{table}

\subsection{Peripheral Blood Mononuclear Cells}
\label{sec:supplement-pbmc}

For GRN inference in PBMC, we perform a hyperparameter search using a 5 fold cross-validation split of the prior-knowledge matrix. Here, $80\%$ of the prior-knowledge is used for training, while the remaining $20\%$ is used for validation. For PBMC, we do not have access to a gold standard network. For this reason, to evaluate our inferred GRNs, we select the network for each of the 5 cross-validation splits that achieves the best training hyperparameters. We then take the intersection of each of these 5 optimal networks, and filter the predicted interactions by those obtaining predicted means (probability) of $> 0.90$ across every split. 

For TFA inference, we select the best overall hyperparameters from the 5 fold cross-validation hyperparemeter search to infer a single network. We do this in order to obtain a single TFA matrix where the entries of this matrix are not affected by averaging across multiple datasets. All predicted TFA values are non-zero by nature of matrix factorization. We thus apply an $l1$ regularization on the matrix with $\lambda = 1$ to push low scoring activity values to 0.

UMAP projections were performed on this regularized TFA matrix using the scanpy package \cite{wolf2018scanpy} with the following parameters: n-neighbors$=10$, n-pcs$=40$. Additional UMAPs for each of the considered PBMC immune TFs are available in Figure \ref{fig:immune-UMAP}. To create the GRN diagrams for PBMC, we used the Gephi Open Graph Viz Platform \cite{bastian2009gephi}. Additional GRNs for each of the PBMC immune TF families are displayed in Figure \ref{fig:immune-GRNs}. PBMC heat-map dot-plots and violin-plot were created using the default scanpy parameters for these functions.

\begin{figure}[!htbp]
    \renewcommand{\figurename}{Figure}
    \setcounter{figure}{1}
    \centering
    \includegraphics[width=0.95\textwidth]{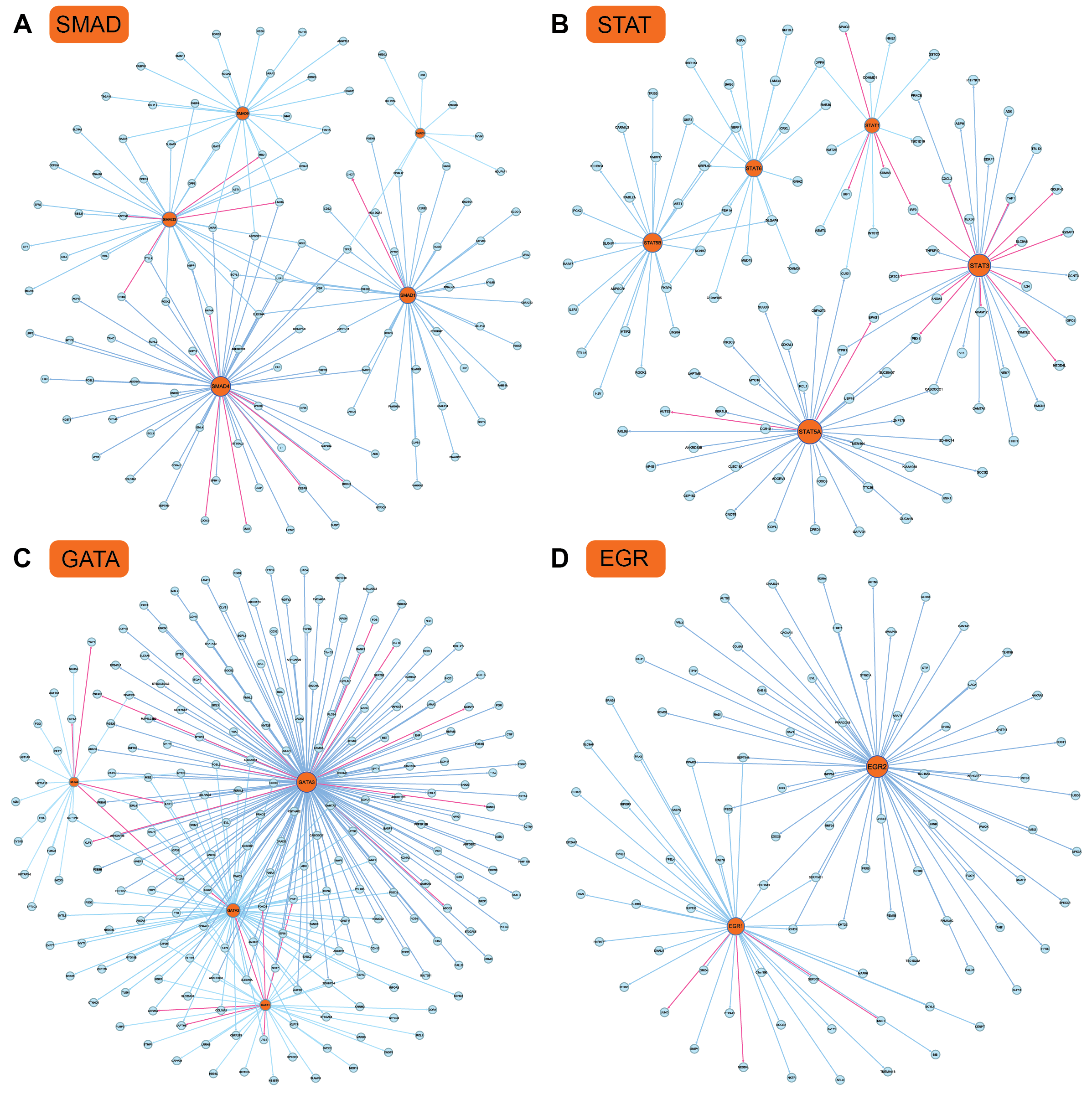}
    \caption{PBMC GRN graphs for the family of TFs belonging to ({\bf A}) SMAD, ({\bf B}) STAT, ({\bf C}) GATA, and ({\bf D}) EGR. TFs for each graph are represented by orange nodes, while target genes are blue nodes. Pink regulatory edges represent interactions with supporting literature. Color scale for blue regulatory edges are scaled from light blue (less out-degree regulation) to darker blue (more out-degree regulation).}
    \label{fig:immune-GRNs}
\end{figure}

\begin{figure}[!htbp]
    \renewcommand{\figurename}{Figure}
    \setcounter{figure}{2}
    \centering
    \includegraphics[width=0.95\textwidth]{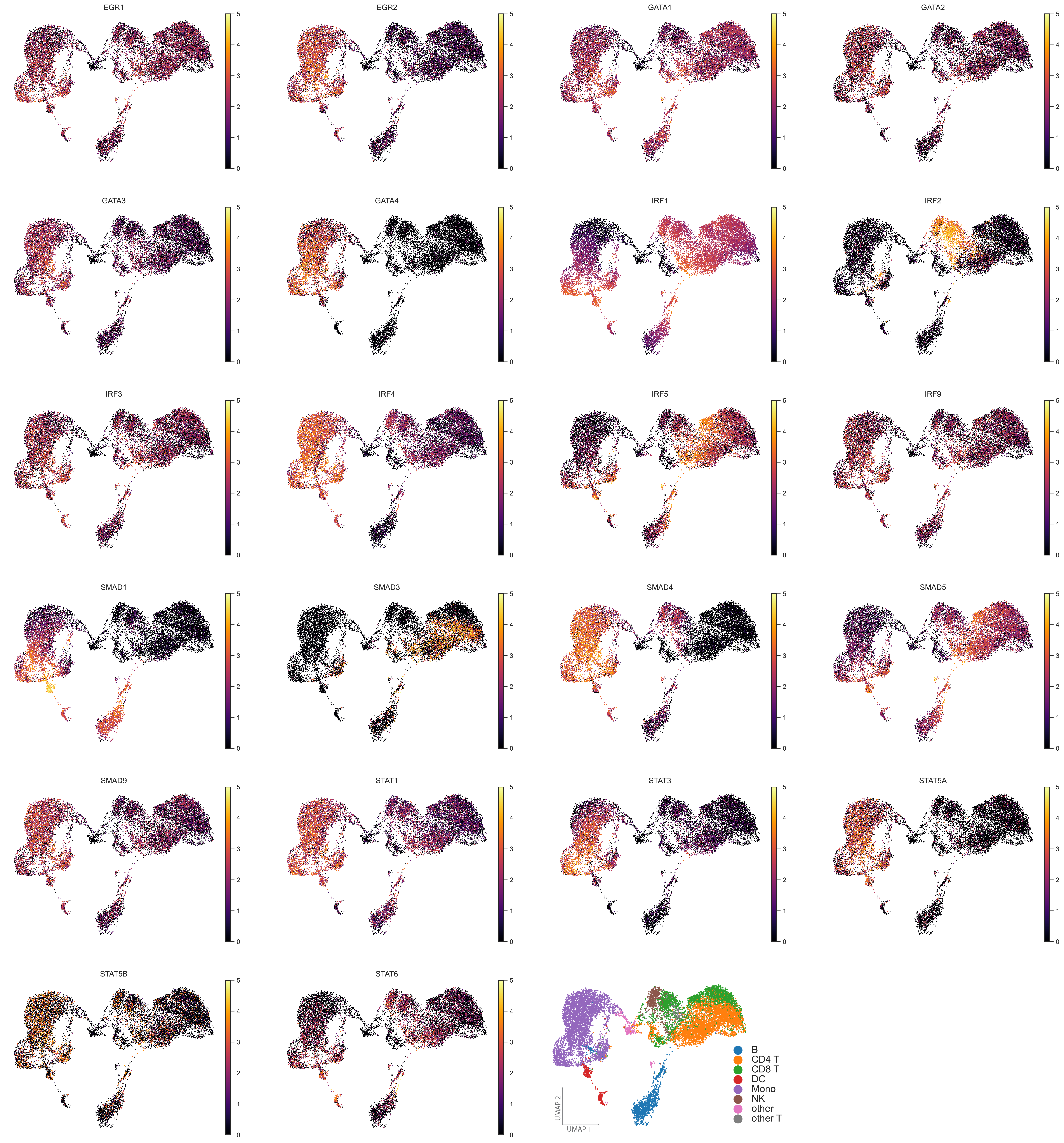}
    \caption{UMAP of predicted PBMC TFA. Each UMAP highlights the specific location of activity for each TF considered from the immune PBMC TFs. The final UMAP serves as a reference to TFA annotated by cell-type. }
    \label{fig:immune-UMAP}
\end{figure}

\begin{table}[!htbp]
\scriptsize
\centering
\renewcommand{\arraystretch}{1.15}
\setlength{\tabcolsep}{4pt}

\begin{threeparttable}
\begin{tabularx}{\textwidth}{@{}>{\raggedright\arraybackslash}p{0.12\textwidth}|  
                            >{\raggedright\arraybackslash}p{0.12\textwidth}
                            >{\raggedright\arraybackslash}p{0.16\textwidth}
                            >{\raggedright\arraybackslash}X@{}}
\toprule
\textbf{TF family} & \textbf{TF} & \textbf{Gene} & \textbf{Citation} \\
\midrule

\multirow{20}{*}{\textbf{STAT}} &
\multirow{4}{*}{STAT1} &
\cellcolor{gray!10}IRF1 & \cellcolor{gray!10}\cite{abou2017properties} \\
& & IRF9 & \cite{au2013transcriptional} \\
& & \cellcolor{gray!10}KDM6B & \cellcolor{gray!10}\cite{johnstone2021dysregulation} \\
& & SPAG9 & \cite{mayumi2021activation} \\
\cmidrule(l){2-4}
& \multirow{14}{*}{STAT3} &
\cellcolor{gray!10}IL24 & \cellcolor{gray!10}\cite{kumari2013tumor}, \cite{andoh2009expression} \\
& & IRF9 & \cite{edsbacker2019stat3} \\
& & \cellcolor{gray!10}ADAM12 & \cellcolor{gray!10}\cite{roy2017adam12} \\
& & CRTC3 & \cite{kim2017creb} \\
& & \cellcolor{gray!10}CXCL2 & \cellcolor{gray!10}\cite{nguyen2010stat3} \\
& & GOLPH3 & \cite{wu2018golph3} \\
& & \cellcolor{gray!10}IQGAP1 & \cellcolor{gray!10}\cite{wei2021role} \\
& & NEDD4L & \cite{nie2022paeoniflorin} \\
& & \cellcolor{gray!10}NRL & \cellcolor{gray!10}\cite{keuthan2019stat3} \\
& & PBX1 & \cite{wei2018pbx1} \\
& & \cellcolor{gray!10}SLC9A8 & \cellcolor{gray!10}\cite{liu2023downregulation} \\
& & YAP1 & \cite{shibata2020concurrent} \\
& & \cellcolor{gray!10}IRF9 & \cellcolor{gray!10}\cite{edsbacker2019stat3} \\
& & ANXA4 & \cite{li2020anxa4} \\
\cmidrule(l){2-4}
& \multirow{2}{*}{STAT5} &
\cellcolor{gray!10}AUTS2 & \cellcolor{gray!10}\cite{nagel2016deregulation} \\
& & EPAS1 & \cite{pietz2017immunopathology} \\
\midrule

\multirow{20}{*}{\textbf{GATA}} &
\multirow{6}{*}{GATA1} &
\cellcolor{gray!10}ATP2B4 & \cellcolor{gray!10}\cite{lessard2017erythroid} \\
& & FOXO3 & \cite{katsumura2017gata} \\
& & \cellcolor{gray!10}GATA2 & \cellcolor{gray!10}\cite{gao2015gata} \\
& & LAPTM5 & \cite{zhang2019glcnac} \\
& & \cellcolor{gray!10}LYL1 & \cellcolor{gray!10}\cite{johnson2007friend} \\
& & PBX1 & \cite{chakrabarti2016transcription} \\
\cmidrule(l){2-4}
& \multirow{1}{*}{GATA2} &
\cellcolor{gray!10}CUX1 & \cellcolor{gray!10}\cite{wu2020integrative} \\
\cmidrule(l){2-4}
& \multirow{9}{*}{GATA3} &
\cellcolor{gray!10}ABCC3 & \cellcolor{gray!10}\cite{kobayashi2016wnt} \\
& & EGFR & \cite{kong2022mammaglobin} \\
& & \cellcolor{gray!10}ETS2 & \cellcolor{gray!10}\cite{blumenthal1999regulation} \\
& & FOS & \cite{liu2023loss} \\
& & \cellcolor{gray!10}FOSL2 & \cellcolor{gray!10}\cite{li2020landscape} \\
& & IQGAP1 & \cite{yang2022identification} \\
& & \cellcolor{gray!10}KLF6 & \cellcolor{gray!10}\cite{fitch2020gata3} \\
& & RUNX2 & \cite{liao2017participation} \\
& & \cellcolor{gray!10}ZNF462 & \cellcolor{gray!10}\cite{hintze2017cell} \\
\cmidrule(l){2-4}
& \multirow{4}{*}{GATA4} &
\cellcolor{gray!10}EPAS1 & \cellcolor{gray!10}\cite{arroyo2021gata4} \\
& & HNF4A & \cite{san2015transcription} \\
& & \cellcolor{gray!10}IL1R1 & \cellcolor{gray!10}\cite{yu2021loss} \\
& & YAP1 & \cite{khalid2021gata4} \\
\midrule

\multirow{6}{*}{\textbf{IRF}} &
\multirow{2}{*}{IRF1} &
\cellcolor{gray!10}B2M & \cellcolor{gray!10}\cite{gobin2003regulation} \\
& & BTN3A1 & \cite{pietz2017immunopathology} \\
\cmidrule(l){2-4}
& \multirow{1}{*}{IRF2} &
\cellcolor{gray!10}B2M & \cellcolor{gray!10}\cite{mercado2019irf2} \\
\cmidrule(l){2-4}
& \multirow{3}{*}{IRF3} &
\cellcolor{gray!10}GPR108 & \cellcolor{gray!10}\cite{zhao2023golgi} \\
& & RNF5 & \cite{zhong2009ubiquitin} \\
& & \cellcolor{gray!10}TRAF2 & \cellcolor{gray!10}\cite{yu2023identification} \\
\bottomrule
\end{tabularx}

\caption{Literature supported interactions for PBMC GRNs for STAT, GATA, and IRF immune TF families.}
\label{table:pbmc-interactions-1a}
\end{threeparttable}
\end{table}

\begin{table}[!htbp]
\scriptsize
\centering
\renewcommand{\arraystretch}{1.15}
\setlength{\tabcolsep}{4pt}

\begin{threeparttable}
\begin{tabularx}{\textwidth}{@{}>{\raggedright\arraybackslash}p{0.12\textwidth}|  
                            >{\raggedright\arraybackslash}p{0.12\textwidth}
                            >{\raggedright\arraybackslash}p{0.16\textwidth}
                            >{\raggedright\arraybackslash}X@{}}
\toprule
\textbf{TF family} & \textbf{TF} & \textbf{Gene} & \textbf{Citation} \\
\midrule

\multirow{12}{*}{\textbf{SMAD}} &
\multirow{1}{*}{SMAD1} &
\cellcolor{gray!10}CHD7 & \cellcolor{gray!10}\cite{liu2014chd7} \\
\cmidrule(l){2-4}
& \multirow{4}{*}{SMAD3} &
\cellcolor{gray!10}LAPTM5 & \cellcolor{gray!10}\cite{gao2019laptm5} \\
& & LIN28A & \cite{jung2020lin28a} \\
& & \cellcolor{gray!10}MSL2 & \cellcolor{gray!10}\cite{jiang2019msi2} \\
& & TRIB3 & \cite{hua2011trb3} \\
\cmidrule(l){2-4}
& \multirow{7}{*}{SMAD4} &
\cellcolor{gray!10}CEBPB & \cellcolor{gray!10}\cite{hill2016transcriptional} \\
& & CXXC5 & \cite{wang2013cxxc5} \\
& & \cellcolor{gray!10}GDF15 & \cellcolor{gray!10}\cite{rochette2021gdf15} \\
& & HNF4 & \cite{chen2019reinforcing} \\
& & \cellcolor{gray!10}ROCK2 & \cellcolor{gray!10}\cite{wang2017megakaryocytic} \\
& & ULK1 & \cite{trelford2021canonical} \\
& & \cellcolor{gray!10}WWOX & \cellcolor{gray!10}\cite{hsu2017hyaluronan} \\
\midrule

\multirow{3}{*}{\textbf{EGR}} &
\multirow{3}{*}{EGR1} &
\cellcolor{gray!10}JUND & \cellcolor{gray!10}\cite{chen2010identification} \\
& & NEDD4L & \cite{liu2022dual} \\
& & \cellcolor{gray!10}NME1 & \cellcolor{gray!10}\cite{wong2021ctcf} \\
\bottomrule
\end{tabularx}

\caption{Literature supported interactions for PBMC GRNs for SMAD and EGR immune TF families.}
\label{table:pbmc-interactions-1b}
\end{threeparttable}
\end{table}

\subsubsection{Bacillus subtilis}

We used two microarray datasets for \textit{B. subtilis}, which we label as B1 (GSE27219) and B2 (GSE67023). Both B1 and B2 underwent different normalization as part of standard microarray processing, described in detail in \cite{nicolas2012condition} and \cite{arrieta2015experimentally}. For the experiment "No Normalization", B1 was simply converted to integers, while B2 contained negative numbers and had to be scaled and then converted to integers so that the data represented positive integers similar to single-cell data.  

To demonstrate the importance of scaling microarray data to place independently collected datasets on the same scale, we demonstrate how Min-Max Scaling improves inference in both \textit{B. subtilis} datasets. For "Min-Max Scaling", both B1 and the positive scaled B2 dataset were subsequently normalized using the following logic. Using the observation axis, values were linearly transformed so that the minimum value was mapped to 0 and the maximum value was mapped to 1. Each value was then multiplied by 100 and converted to integers to produce the resulting expression matrix of scaled single-cell-like integers.

\subsection{Inferelator, Scenic, and CellOracle Networks}

\subsubsection{Saccharomyces cerevisiae}
Networks were inferred using the "multitask" workflow setting of the Inferelator for the same single-cell \textit{S.cerevisiae} datasets described in \cite{skok2022high}. For each algorithm, BBSR, StARS, and AMuSR, the following parameters were used: gold\_standard\_filter\_method = \\
"keep\_all\_gold\_standard", num\_bootstraps=5. Aggregated multi-task networks were used for benchmarking, while single-task networks were disregarded for the purpose of this work. To make these networks directly comparable to PMF, we did not make use of  normalization, count minimum, or meta-data options available within the Inferelator workflow.

Networks inferred with Scenic and CellOracle used the same input files, with no additional parameters specified.

\begin{table}[!htbp]
    \centering
    \renewcommand{\arraystretch}{1.2}  
    \begin{tabular}{l|ccc}
        \multicolumn{1}{c}{} & \multicolumn{3}{c}{\textbf{Prior Information}} \\ \cline{2-4}
        \textbf{Method} & \textbf{Regular} & \textbf{None} & \textbf{Shuffled} \\ \hline
        PMF-GRN     & $0.375$ & $0.014$ & $0.023$ \\
        AmUSR       & $0.223$ & $0.024$ & $0.019$ \\
        BBSR        & $0.402$ & $0.022$ & $0.018$ \\
        StARS       & $0.186$ & $0.028$ & $0.017$ \\
        SCENIC      & $0.014$ & $0.014$ & $0.014$ \\
        CellOracle  & $0.383$ & N/A     & $0.013$ \\
    \end{tabular}
    \caption{AUPRCs achieved by PMF-GRN, the Inferelator algorithms (AMuSR, BBSR, and StARS), Scenic and CellOracle on \textit{S. cerevisiae} datasets.}
    \label{table:yeast-auprc}
\end{table}

\begin{table}[!htbp]
    \centering
    \renewcommand{\arraystretch}{1.2}  
    \begin{tabular}{l|ccccc}
        \multicolumn{1}{c}{} & \multicolumn{5}{c}{\textbf{Cross Validation Split}} \\ \cline{2-6}
        \textbf{Method} & \textbf{Split 1} & \textbf{Split 2} & \textbf{Split 3} & \textbf{Split 4} & \textbf{Split 5} \\ \hline
        PMF-GRN     & $0.114$ & $0.096$ & $0.086$ & $0.1342$ & $0.118$ \\
        BBSR        & $0.112$ & $0.128$ & $0.161$ & $0.171$  & $0.139$ \\
        StARS       & $0.109$ & $0.137$ & $0.154$ & $0.195$  & $0.151$ \\
        SCENIC      & $0.020$ & $0.021$ & $0.018$ & $0.025$  & $0.021$ \\
        CellOracle  & $0.034$ & $0.042$ & $0.034$ & $0.043$  & $0.034$ \\
    \end{tabular}
    \caption{AUPRCs achieved by PMF-GRN, the Inferelator algorithms (BBSR, and StARS), Scenic and CellOracle on \textit{S. cerevisiae} datasets using the gold standard for 5-fold cross validation.}
    \label{table:yeast-cv-auprc}
\end{table}

\begin{table}[!htbp]
    \centering
    \renewcommand{\arraystretch}{1.2}  
    \begin{tabular}{l|cccc}
        \multicolumn{1}{c}{} & \multicolumn{4}{c}{\textbf{Noise Added}} \\ \cline{2-5}
        \textbf{Method} & \textbf{No Noise} & \textbf{$100\%$ Noise} & \textbf{$250\%$ Noise} & \textbf{$500\%$ Noise} \\ \hline
        PMF-GRN     & $0.343$ & $0.280$ & $0.198$ & $0.149$ \\
        BBSR        & $0.264$ & $0.208$ & $0.186$ & $0.174$ \\
        StARS       & $0.136$ & $0.125$ & $0.114$ & $0.118$ \\
        SCENIC      & $0.075$ & $0.068$ & $0.059$ & $0.055$ \\
        CellOracle  & $0.417$ & $0.306$ & $0.226$ & $0.175$ \\
    \end{tabular}
    \caption{AUPRCs achieved by PMF-GRN, the Inferelator algorithms (BBSR, and StARS), Scenic and CellOracle on \textit{S. cerevisiae} datasets using increasing amounts of noise added to the prior-knowledge data.}
    \label{table:yeast-noise-auprc}
\end{table}

\begin{table}[!htbp]
    \centering
    \renewcommand{\arraystretch}{1.2}  
    \begin{tabular}{l|c}
        \multicolumn{1}{c}{} & \multicolumn{1}{c}{\textbf{Intersection over Union (IoU)}} \\ \cline{2-2}
        \textbf{Method} & \textbf{IoU} \\ \hline
        PMF-GRN     & $15.69\%$ \\
        BBSR        & $14.56\%$ \\
        AMuSR       & $12.46\%$ \\
        StARS       & $11.78\%$ \\
        SCENIC      & $3.17\%$ \\
        CellOracle  & $30.28\%$ \\
    \end{tabular}
    \caption{Intersection over Union (IoU) scores achieved by PMF-GRN, the Inferelator algorithms (AMuSR, BBSR, and StARS), Scenic and CellOracle for GRNs learned on individual \textit{S. cerevisiae} datasets.}
    \label{table:yeast-iou}
\end{table}

\begin{table}[!htbp]
    \centering
    \renewcommand{\arraystretch}{1.2}  
    \begin{tabular}{l|ccccc}
        \multicolumn{1}{c}{} & \multicolumn{5}{c}{\textbf{Sample Number}} \\ \cline{2-6}
        \textbf{Expression size (\%)} & \textbf{1} & \textbf{2} & \textbf{3} & \textbf{4} & \textbf{5} \\ \hline
        $80\%$ & $0.2838$ & $0.2947$ & $0.3395$ & $0.3325$ & $0.2774$ \\
        $60\%$ & $0.2382$ & $0.3295$ & $0.3252$ & $0.2775$ & $0.3296$ \\
        $40\%$ & $0.3157$ & $0.2723$ & $0.3599$ & $0.2858$ & $0.2939$ \\
        $20\%$ & $0.2995$ & $0.3503$ & $0.3229$ & $0.2868$ & $0.3335$ \\
    \end{tabular}
    \caption{AUPRCs achieved by PMF-GRN across 4 different downsample sizes (80\%, 60\%, 40\%, and 20\%), across 5 samples for each downsample size.}
    \label{table:yeast-downsample}
\end{table}

\begin{table}[!htbp]
    \centering
    \renewcommand{\arraystretch}{1.2}  
    \begin{tabular}{l|cccccc}
        \multicolumn{1}{c}{} & \multicolumn{6}{c}{\textbf{Dataset}} \\ \cline{2-7}
        \textbf{Cross Validation Split} & \textbf{1} & \textbf{2} & \textbf{3} & \textbf{4} & \textbf{5} & \textbf{Full GRN} \\ \hline
        $80\%$ train - $20\%$ val & $0.0460$ & $0.0400$ & $0.0463$ & $0.0361$ & $0.0408$ & $0.3337$ \\
        $60\%$ train - $40\%$ val & $0.0479$ & $0.0378$ & $0.0463$ & $0.0417$ & $0.0437$ & $0.3337$ \\
        $40\%$ train - $60\%$ val & $0.0459$ & $0.0397$ & $0.0375$ & $0.0347$ & $0.0381$ & $0.2909$ \\
        $20\%$ train - $80\%$ val & $0.0401$ & $0.0434$ & $0.0403$ & $0.0427$ & $0.0239$ & $0.2882$ \\
    \end{tabular}
    \caption{AUPRCs achieved by PMF-GRN across 4 different cross-validation splits. 5 hyperparameters searches were performed for each cross-validation split. Full GRN was inferred using the hyperparameters for the best overall AUPRC per cross-validation split.}
    \label{table:yeast-cv-experiment}
\end{table}

\begin{table}[!htbp]
\scriptsize
\centering
\renewcommand{\arraystretch}{1.25}
\setlength{\tabcolsep}{4pt}
\rowcolors{2}{gray!6}{white}

\begin{tabularx}{\textwidth}{>{\raggedright\arraybackslash}p{0.11\textwidth}
                            >{\raggedright\arraybackslash}X
                            >{\raggedright\arraybackslash}p{0.16\textwidth}
                            >{\raggedright\arraybackslash}p{0.22\textwidth}
                            >{\raggedright\arraybackslash}X}
\toprule
\textbf{Method} & \textbf{Input} & \textbf{Output} & \textbf{Methodology} & \textbf{Pipeline} \\
\midrule
PMF-GRN &
(1) one or more scRNA-seq datasets
(2) priors constructed from genomic data such as ATAC-seq or ChIP-seq and TF motifs or literature database derived priors &
(1) individual (and) combined GRN
(2) TFA &
probabilistic matrix factorization, variational inference &
1) hyperparameter search using 80-20 split of input prior
(2) GRN inference with optimal model parameters
(3) evaluate GRNs with AUPRC, MCC and F1 scores and uncertainty calibration \\
\addlinespace[2pt]
SCENIC &
(1) scRNA-seq
(2) a list of TFs ranking databases for motif analysis &
(1) GRN: loom file containing regulons (interactions between TFs and their target genes)
(2) TFA: biological activity of regulons in a given cell &
tree based regression models, gradient boosting machine regression, motif enrichment analysis, and gaussian mixture models for regulon activity binarization &
(1-4) preprocessing steps
(5) network inference
(6) module generation
(7) motif enrichment and TF regulon prediction
(8) cellular enrichment
(9) optional binarization of cellular regulon activity
(10) clustering of cells based on regulon activity \\
\addlinespace[2pt]
CellOracle &
(1) ATAC-seq + TF motifs
(2) scRNA-seq &
(1) GRN
(2) cell-state transition vectors after gene perturbation &
Bayesian Ridge regression &
(1) base GRN construction with scATAC-seq or promoter databases
(2) scRNA-seq preprocessing
(3) context dependent GRN inference
(4) network analysis
(5) simulation of cell identity after TF perturbation
(6) calculation of pseudotime gradient for perturbation scores \\
\addlinespace[2pt]
AMuSR &
(1) two or more scRNA-seq datasets
(2) one or more priors constructed from genomic data and TF motifs or literature database derived priors &
(1) individual and combined GRNs
(2)  TFA &
Regularized linear regression &
(1) estimating TFA
(2) learning regression parameters
(3) model selection with bayesian information criterion
(4) evaluate GRNs with AUPRC, MCC, F1 scores \\
\addlinespace[2pt]
BBSR &
(1) one or more scRNA-seq datasets
(2) one or more priors constructed from genomic data and TF motifs or literature database derived priors &
(1) individual (and) combined GRN
(2) TFA &
Bayesian best subset regression &
(1) estimate TFA
(2) learn regression parameters
(3) model selection
(4) evaluate GRNs with AUPRC, MCC, F1 scores \\
\addlinespace[2pt]
StARS &
(1) one or more scRNA-seq datasets
(2) one or more priors constructed from genomic data and TF motifs or literature database derived priors &
(1) individual (and) combined GRN
(2) TFA &
Least absolute shrinkage and selection operator combined with the Stability Approach to Regularization Selection &
(1) estimate TFA
(2) learn regression parameters
(3) model selection
(4) evaluate GRNs with AUPRC, MCC, F1 scores \\
\bottomrule
\end{tabularx}

\caption{GRN inference method comparison table. Table includes input data, output, methodology and pipeline organization for each of the six GRN inference methods discussed in this work.}
\label{tab:grn-methods}
\end{table}

\clearpage
\subsection{TF Target Gene Connectivity Matrix Generation}

\subsubsection{Saccharomyces cerevisiae}
Datasets were obtained from \cite{skok2022high} without further modification.

\subsubsection{Peripheral Blood Mononuclear Cells}
Datasets were obtained from \cite{tjarnberg2023structure} without further modification.

\subsubsection{BEELINE}
Datasets were obtained from \cite{pratapa2020benchmarking}. TF-target gene matrices were constructed by creating cross-tab matrices from the RefGRN. $50\%$ of this matrix was used as prior-knowledge, the remaining $50\%$ was used as a gold standard for evaluation.

\subsubsection{Bacillus subtilis}
A prior-known TF-target gene interactions matrix was obtained from the Subtiwiki database \cite{faria2016reconstruction} from "regulations" (downloaded 07/21/22). Using the columns "regulator locus" and "gene locus" a cross-tab integer matrix was created, where 1 represents the existence of an interaction and 0 represents no interaction. This matrix was randomly split 5 times in 80\%-20\% proportions along the gene axis to generate independent prior-known information and gold standard matrices.

\chapter{A Genomic Language Model Prior for Gene Regulatory Network Inference}
\label{chp-2}
\label{sec:GLM-Prior-Chapter}
This chapter is adapted from the pre-print "GLM-Prior: a genomic language model for transferable sequence-derived priors in gene regulatory network inference". \\
\textbf{Claudia Skok Gibbs}, Angelica Chen, Richard Bonneau, and Kyunghyun Cho. (\textbf{2026}).

\section{Abstract}
Gene regulatory network inference depends on high-quality prior knowledge, yet curated priors are often incomplete or unavailable across species and cell types. 
We present GLM-Prior, a genomic language model fine-tuned to predict transcription factor–target gene interactions directly from nucleotide sequence. 
We integrate GLM-Prior with PMF-GRN, a probabilistic matrix factorization model, to create a dual-stage pipeline that combines sequence-derived priors with single-cell gene expression data for GRN inference. 
Across six human, mouse, and yeast cell lines, GLM-Prior performance scales with positive label abundance and diverse transcription factor coverage, achieving strong accuracy in well-annotated mammalian contexts.
We evaluate single-species, species-transfer learning, and multi-species training paradigms and show that GLM-Prior generalizes to held-out gene and TF sequences, enabling experiment-agnostic prior construction in previously unprofiled contexts.
Furthermore, comparisons to accessibility-based priors across multiple GRN inference methods show that GLM-Prior provides the most robust priors in mammalian cell lines. 
Together, our results demonstrate that prior construction, rather than the choice of GRN inference algorithm, is the primary determinant of GRN inference performance, and establish GLM-Prior as a framework for building high-quality, experiment-agnostic priors that can be deployed even in understudied or experimentally inaccessible systems.

\section{Introduction}
\label{sec:introduction}
Gene regulatory networks (GRNs) map the transcriptional relationships between transcription factors (TFs) and their target genes, providing a framework for understanding cellular function and gene expression control in cells \cite{badia2023gene, kim2023gene}. Constructing accurate GRNs requires integrating large-scale sequencing datasets, such as single-cell RNA-seq and ATAC-seq, to infer regulatory interactions that cannot be measured experimentally \cite{fiers2018mapping}. However, the accuracy of GRN inference depends heavily on the quality of the prior knowledge used to guide the inference algorithm \cite{stock2025leveraging, mccalla2023identifying}. In the context of GRNs, prior knowledge refers to an initial matrix of putative TF-gene regulatory interactions, which provides structural guidance to the inference algorithm by defining which TF-gene pairs are more likely to represent true regulatory edges. Recent methods have demonstrated that better prior knowledge, which captures biologically meaningful regulatory signals, leads to more accurate and informative GRN reconstruction, highlighting the critical role of prior knowledge in improving GRN inference quality \cite{skok2022high, skok2024pmf, kamimoto2023dissecting, van2020scalable}. 

Prior knowledge for well-studied species is typically constructed using databases of experimentally validated TF-gene interactions \cite{de2012encode, teixeira2018yeastract}. For less-characterized organisms, prior knowledge is inferred by combining structural genomic data with sequencing information \cite{mercatelli2020gene}. The Inferelator 3.0's Inferelator-Prior performs TF motif enrichment within open chromatin regions to define edges between proximally bound TFs to target genes \cite{skok2022high}. CellOracle's base GRN draws edges between TFs bound within promoter regions of accessible target genes \cite{kamimoto2023dissecting}. SCENIC's CisTarget constructs prior knowledge by ranking enriched motifs in accessible regions using genome-wide motif scores and a hidden Markov model to predict TF-gene interactions \cite{van2020scalable}. While these methods demonstrate the importance of prior knowledge in GRN inference, they also expose key limitations. Motif annotation remains incomplete \cite{inukai2017transcription}, chromatin accessibility data is noisy and cell-type specific \cite{buenrostro2015single}, and existing methods often fail to capture long-range regulatory interactions \cite{loers2024single}. Further, these existing approaches cannot create prior knowledge matrices which generalize to other species. These limitations underscore the need for a more generalizable approach to constructing high-quality prior knowledge matrices, particularly for poorly characterized organisms \cite{stock2025leveraging, barbosa2018guide}. 

A promising direction is to incorporate transformer-based genomic foundation models which learn regulatory logic directly from DNA sequence. Architectures such as the Nucleotide Transformer \cite{dalla2024nucleotide} have demonstrated strong performance across genomic tasks, including predicting chromatin accessibility, gene expression and transcription factor binding \cite{avsec2021effective, ji2021dnabert, nguyen2023hyenadna, linder2025predicting}. By learning conserved regulatory patterns directly from sequence data, transformer-based genomic models offer a compelling solution to the limitations of motif-based prior knowledge approaches \cite{consens2025transformers}. By leveraging attention mechanisms, they effectively capture long-range dependencies in DNA sequences \cite{vaswani2017attention}, enabling more accurate modeling of complex regulatory interactions. When trained on large, multi-species genomic datasets, these models encode both species-specific and cross-species features, supporting generalization across cell types \cite{avsec2021effective, cui2024scgpt} and organisms \cite{dalla2024nucleotide, zhou2023dnabert}. This opens the door to a powerful transfer learning paradigm: models trained on species with extensive regulatory data (e.g., \textit{H. Sapiens}) can be adapted to predict interactions in related species (e.g., \textit{M. musculus}), when curated priors are incomplete or unavailable. In this way, genomic language models offer the potential to expand prior knowledge construction to organisms and cell types for which no direct interaction data exist, enabling GRN inference in new and challenging contexts.

In this work, we present GLM-Prior, a novel approach for constructing high-quality prior knowledge to improve downstream GRN inference. GLM-Prior is built by fine-tuning the 250 million parameter Nucleotide Transformer \cite{dalla2024nucleotide}, pretrained on the genomes of 850 species, into a sequence classification model for predicting TF-gene regulatory interactions. Unlike motif-based approaches that rely on the proximity of TF motifs to accessible promoter regions, GLM-Prior directly predicts interactions from paired nucleotide sequences of TF binding motifs and gene bodies. These sequences are jointly encoded by the transformer and passed through a classification head, which outputs a probability of regulatory interaction for each TF-gene pair. To construct a binary prior knowledge matrix suitable for downstream GRN inference, we apply an optimized classification threshold that converts these probabilities into discrete binary interactions. As most TF-gene pairs are non-regulatory, the model is trained using class-weighted loss and balanced sampling to mitigate class imbalance during training. Together with the Transformer's attention mechanisms, which capture long-range dependencies and complex regulatory grammar, these features enable GLM-Prior to generalize across cell types and species more effectively than motif-based approaches. 

We investigate the generalizability of GLM-Prior by evaluating three training paradigms: $(1)$ single-species models trained and evaluated on the same organism; $(2)$ transfer learning, in which models trained in one species are used to predict regulatory interactions in an unseen species; and $(3)$ multi-species training, where a single model is fine-tuned on data from multiple organisms. Across these settings, performance scales with abundance and diversity of positive class labels and TF coverage in the training data. Human and mouse cell lines with rich label sets yield higher AUPRCs than yeast, where extreme class imbalance and sparse labels limit generalization. Transfer learning between evolutionarily related species (human and mouse) maintains or modestly improves performance relative to single-species models, demonstrating that GLM-Prior captures conserved regulatory sequence features that can be reused across lineages. In contrast, transfer learning between mammals and yeast does not exceed near-chance performance, reflecting a deeper divergence in regulatory grammar. Multi-species training on human, mouse, and yeast preserves accuracy comparable to the best single-species or transfer models in most cell lines, with only mild trade-offs in some human contexts, and produces a domain-stable prior that can be applied to new systems without species-specific retraining.

To evaluate the utility of GLM-Prior, we incorporate it into a dual-stage GRN inference pipeline. In the first stage, GLM-Prior generates a prior knowledge matrix of TF-gene interactions from nucleotide sequence. In the second stage, this matrix is used to constrain PMF-GRN \cite{skok2024pmf}, a probabilistic matrix factorization model that infers GRNs from single-cell gene expression data. This framework allows us to isolate and compare the contributions of prior construction and expression-based inference to overall GRN reconstruction performance.

Building on this, we benchmark GLM-Prior against accessibility-based priors (Inferelator-Prior \cite{skok2022high}, and CellOracle's baseGRN \cite{kamimoto2023dissecting}), and integrate each prior with it's downstream GRN inference algorithm (PMF-GRN \cite{skok2024pmf}, the Inferelator 3.0 \cite{skok2022high}, and CellOracle \cite{kamimoto2023dissecting}) across yeast, mouse, and human. GLM-Prior consistently performs at or above chance in all six cell lines and provides the strongest prior in four of six contexts (hESC, mESC, mDC, and mHSC). In contrast, accessibility-based priors remain competitive in settings dominated by proximal regulatory logic, such as yeast, but struggle to generalize above chance performance in complex mammalian cell lines. When these priors are passed into downstream GRN inference, GRN inference typically yields modest gains when the prior is already strong, and larger improvements are observed only when the prior is weak or inaccurate. Critically, the identity of the best-performing GRN in each cell line almost always matches the identity of the best prior, and no GRN inference method is uniformly superior across priors or species. Together, these results indicate that prior construction, rather than the choice of GRN inference algorithm, is the primary driver of GRN reconstruction performance. 

Together, our findings recast GRN inference as a problem primarily limited by the quality of prior knowledge rather than by expression-based modeling. GLM-Prior provides a scalable, species-aware framework for constructing priors, while downstream GRN methods act mainly to refine or contextualize this scaffold. In this work, we characterize GLM-Prior's generalization across cell types and species, quantify its advantages over accessibility-based priors, and dissect how different GRN inference algorithms interact with these priors to shape the final inferred networks.

\section{Results}
\label{sec:results}

\subsection{Dual-stage training pipeline overview}
\label{sec:DST-overview}

We develop a dual-stage training pipeline to improve the accuracy and interpretability of GRN inference by integrating sequence-derived regulatory interactions with probabilistic modeling of single-cell expression data (Figure \ref{fig:DST}). The pipeline is designed to decouple the construction of prior knowledge from the expression-driven GRN inference step, allowing each component to be independently optimized and evaluated. In this framework, the first stage generates a biologically informed prior knowledge matrix by using a fine-tuned genomic language model (GLM-Prior), while the second stage uses this matrix as structural input to the probabilistic matrix factorization model (PMF-GRN \cite{skok2024pmf}) to a infer GRN.

\subsubsection{Stage 1: Constructing sequence-informed prior knowledge with GLM-Prior}
\label{sec:GLM-Prior}
\begin{figure}[!htbp]
    \centering
    \includegraphics[width=1.0\textwidth]{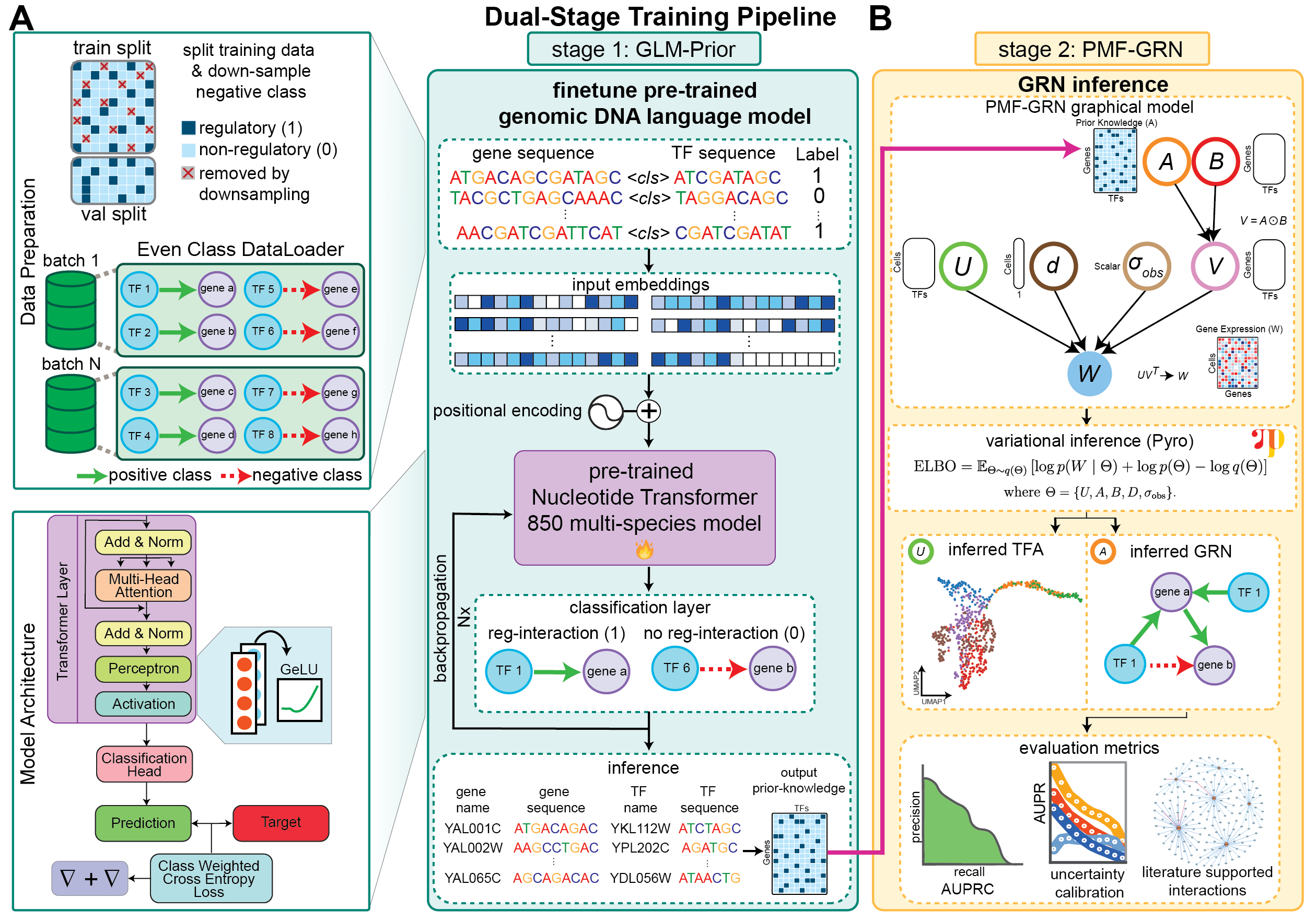} 
    \caption{Schematic of the dual-stage training pipeline. (\textbf{A}) GLM-Prior is trained on concatenated TF binding motifs and gene sequences labeled as interacting or non-interacting. Input data is split into training and validation sets, where the training set is downsampled and placed into even class batches using a custom DataLoader. These input sequences are embedded and encoded by the transformer architecture, and passed through a classification head to predict the probability of a regulatory interaction. After training, inference is performed on all TF-gene pairs to generate a prior knowledge matrix. (\textbf{B}) Inferred prior knowledge is passed to PMF-GRN to provide a structural constraint to the probabilistic graphical model during inference. PMF-GRN infers transcription factor activity and a GRN for the input cell-line dataset. The resulting GRN is then evaluated using AUPRC and uncertainty calibration with an independent gold standard.}
    \label{fig:DST}
\end{figure}

In the first stage of the pipeline, we construct a prior knowledge edge matrix representing predicted positive regulatory ($1$) and negative non-regulatory ($0$) interactions between TFs and their target genes (Figure \ref{fig:DST}A). To generate these predictions, we develop GLM-Prior, a transformer-based sequence classification model built by fine-tuning the 250 million parameter Nucleotide Transformer \cite{dalla2024nucleotide} model, which was pretrained on the genomes of 850 different species. The primary goal of GLM-Prior is to leverage sequence information to accurately predict TF-gene regulatory interactions, in order to construct an informative prior knowledge matrix.

GLM-Prior takes as input paired nucleotide sequences consisting of TF binding motifs and gene body regions, along with labels representing known positive or negative interactions. These sequences are concatenated and encoded into a shared representation using the transformer's embedding and encoding layers. This representation is passed through a classification layer that outputs the probability of a true regulatory interaction. During training, the model evaluates a range of classification thresholds, selecting the threshold that maximizes the F1 score on the validation set. This optimal threshold is used to binarize the predicted probabilities into positive or negative interactions for the final prior knowledge matrix. 

Since GRN datasets are often highly imbalanced, with far fewer positive interactions compared to negative interactions, we apply several strategies to mitigate this imbalance during training. Specifically, we apply a downsampling rate to the negative class and use a class-weighted cross-entropy loss, tuning the negative class weight through hyperparameter search. Additionally, to further ensure stable training, we implement a custom DataLoader to create training batches with equal numbers of positive and negative examples, ensuring balanced gradient updates.

After training, GLM-Prior performs inference across all provided TF-gene pairs, generating a dense probability prior knowledge matrix. The predicted probabilities are then binarized using the optimal F1 score-derived classification threshold, generating a binary prior knowledge matrix that encodes regulatory interactions for a given organism. 
Although we can use the probabilities directly, we binarize the prior knowledge matrix to be in line with existing methods and for more direct comparison.
We then either use this sequence-based prior knowledge as is and evaluate on an independent gold standard, or can pass it into the second stage of the pipeline, enabling a modular information transfer from model-driven sequence inference to data-driven GRN inference.

\subsubsection{Stage 2: Refining GRN inference from single-cell data with PMF-GRN}
\label{sec:PMF-GRN}
In the second stage of the pipeline, the prior knowledge matrix predicted by GLM-Prior serves as a structured input to our previously developed model, PMF-GRN \cite{skok2024pmf} (Figure \ref{fig:DST}B). PMF-GRN leverages this sequence-informed prior knowledge to enhance the inference of GRNs from single-cell gene expression data. It does so by using probabilistic matrix factorization, which decomposes the observed gene expression into latent factors representing transcription factor activity (TFA) and TF-target gene regulatory relationships. 

Concretely, PMF-GRN models the expression matrix $W$ using the likelihood function 
\begin{align*}
    p(W | U, V=A \odot B, \sigma_{obs}, d), 
\end{align*}
where $\mathbb{E} [W] \approx d\odot UV^{\top}$, with per-cell sequencing depth $d$ and observation noise $\sigma_{obs}$. Here, $U$ is a nonnegative cells by TFs matrix of TF activities, and $V$ is a genes by TFs matrix, which is further factorized as $V = A \odot B$. $A \in (0, 1)$ is defined as the degree of existence of a TF-target gene interaction and $B \in \mathbb{R}$ is its signed effect. Each latent variable, $U, A, B, d,$, $\sigma_{obs}$ is assumed independent \textit{a priori}. In this formulation, the GLM-Prior predicted prior knowledge matrix provides the hyperparameters for the logistic-normal prior on $A$, anchoring factors to specific TFs and mitigating column-label identifiability inherent to matrix factorization.

Using variational inference, PMF-GRN estimates the posterior distribution of all latent variables. The posterior mean of $A$ defines the inferred GRN, while the posterior variance provides an edge-level uncertainty score. These uncertainty estimates enable confidence-weighted interpretation of TF-gene edges and are especially valuable when gold standard regulatory annotations are incomplete or unavailable.

The combination of GLM-Prior and PMF-GRN is motivated by their complementary strengths. GLM-Prior uses nucleotide sequence to construct a biologically grounded scaffold of regulatory edges, capturing both local motif-driven interactions and potential long-range dependencies. This scaffold can be generated as a general prior matrix or tailored to specific cell lines or conditions. However, as GLM-Prior has not seen any experimental data, its predictions require refinement using real observations. PMF-GRN provides this refinement by integrating single-cell expression data within a principled generative framework, updating edge probabilities based on observed regulatory patterns while maintaining consistency with the sequence-derived prior. Moreover, like most GRN inference algorithms, PMF-GRN relies on prior knowledge to obtain a meaningful posterior GRN; without an informative prior, factor identifiability issues limit interpretability. GLM-Prior resolves this limitation by anchoring TF-gene relationships, while PMF-GRN tailors the scaffold to the data, quantifies uncertainty, and produces context-specific networks. Together, this integration balances predictive power from large-scale genomic language models with the interpretability and flexibility of probabilistic modeling, yielding GRNs that are both grounded in sequence evidence and informed by experimental observations. 

\subsection{GLM-Prior's generalization scales with data composition across species and cell types}
\label{sec:GLM-Prior-Model-Results}

\begin{figure}[!htbp]
    \centering
    \includegraphics[width=1.0\textwidth]{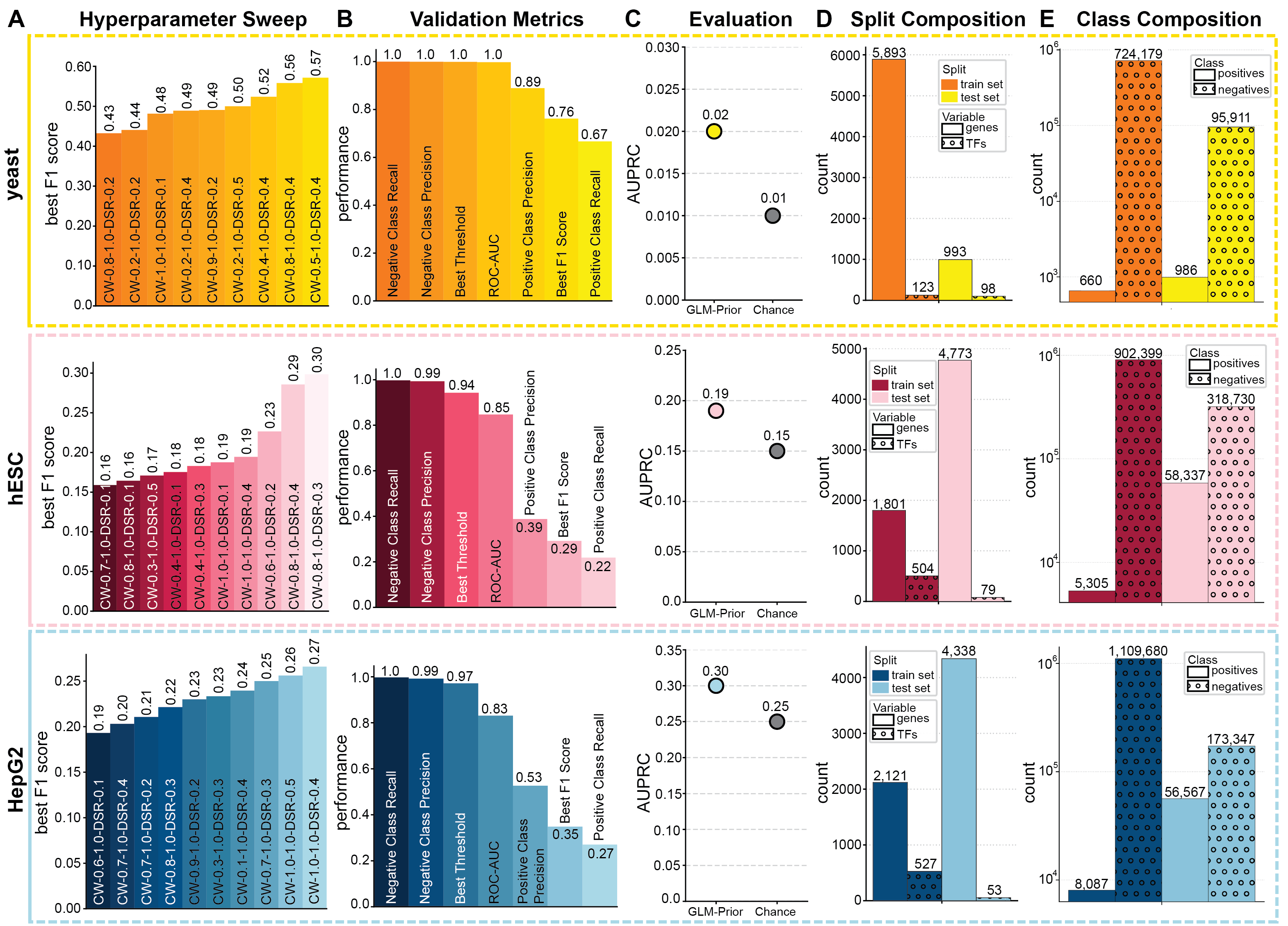} 
    \caption{GLM-Prior model performance across yeast, hESC, and HepG2 cell lines. (\textbf{A}) One-epoch hyperparameter sweep over class-weights and negative-class downsampling rates, evaluated by F1-score. (\textbf{B}) Validation metrics during $10$ epoch training using the best hyperparameter configuration. (\textbf{C}) Test AUPRC on held-out gold standards (with chance performance shown for reference in gray). (\textbf{D}) Train and test split composition (number of genes and TFs). (\textbf{E}) Class composition (number of positive and negative labels) in train and test sets.}
    \label{fig:figure2-a}
\end{figure}

\begin{figure}[!htbp]
    \centering
    \includegraphics[width=1.0\textwidth]{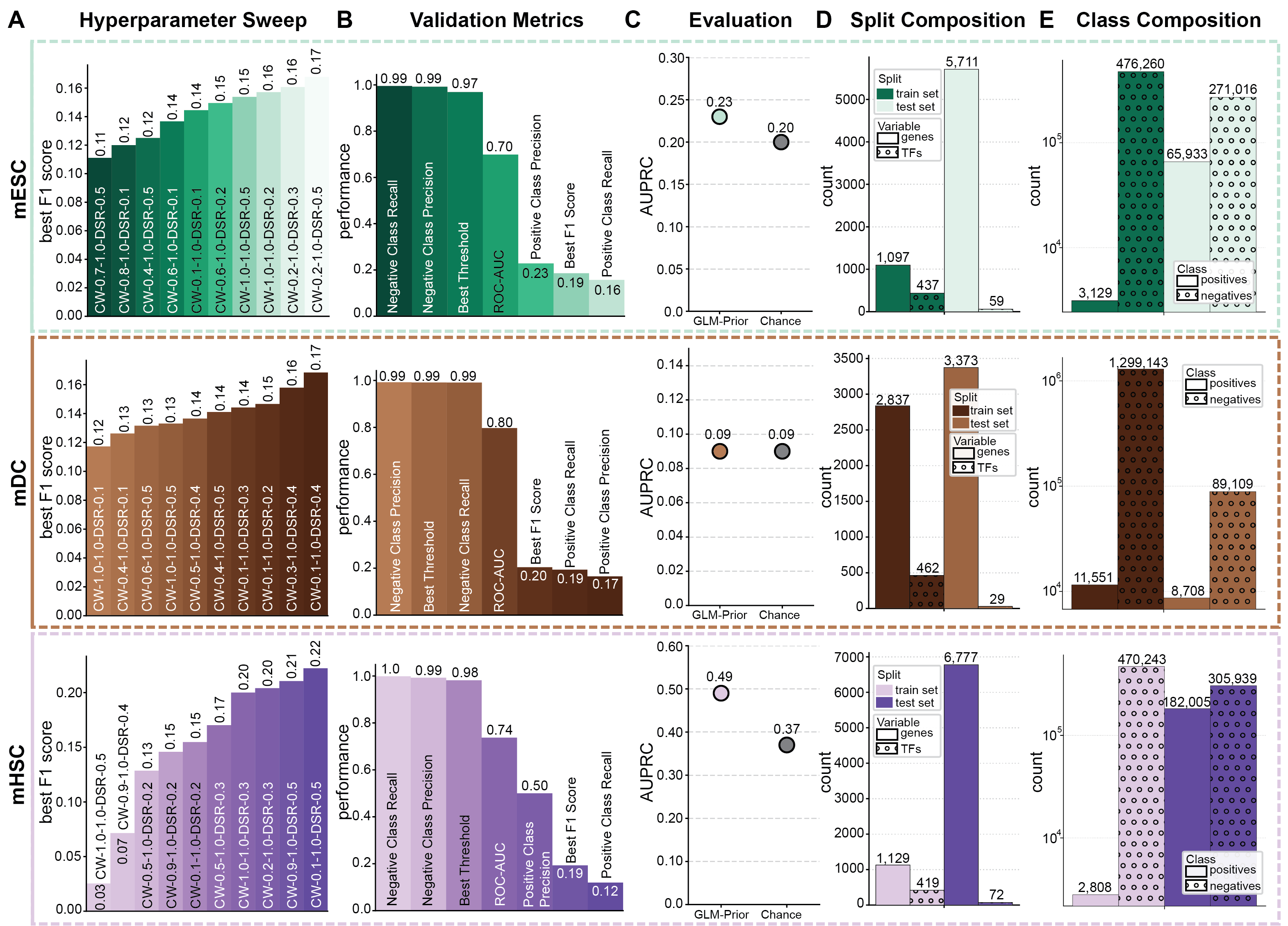} 
    \caption{GLM-Prior model performance across mESC, mDC, and mHSC cell lines. (\textbf{A}) One-epoch hyperparameter sweep over class-weights and negative-class downsampling rates, evaluated by F1-score. (\textbf{B}) Validation metrics during $10$ epoch training using the best hyperparameter configuration. (\textbf{C}) Test AUPRC on held-out gold standards (with chance performance shown for reference in gray). (\textbf{D}) Train and test split composition (number of genes and TFs). (\textbf{E}) Class composition (number of positive and negative labels) in train and test sets.}
    \label{fig:figure2-b}
\end{figure}

We assess GLM-Prior by benchmarking across six cell lines spanning yeast, human, and mouse (Figure \ref{fig:figure2-a} and \ref{fig:figure2-b}). For each organism, we create gene-TF sequence pairs by extracting nucleotide sequences for genes from GTF annotations and TF binding motifs from CisBP \cite{weirauch2014determination}. For each gene-TF pair, we gather available regulatory labels from the YEASTRACT database \cite{teixeira2018yeastract} for yeast, and the STRING \cite{mering2003string, szklarczyk2010string, szklarczyk2021string, szklarczyk2023string} and TRRUST \cite{han2015trrust, han2018trrust} databases for human and mouse. To assess out-of-distribution performance, we evaluate on held-out test sets constructed from well-established reference labels such as the literature curated gold standard from \cite{tchourine2018condition} for yeast, as well as cell-line specific reference networks from BEELINE \cite{pratapa2020benchmarking} for human and mouse. Before training each cell-line specific model, we ensure strict independence between training and test sets at the level of gene-TF sequence pairs and their labels, eliminating sequence and label leakage. This design ensures that performance reflects the model's ability to generalize to unseen sequences and correctly predict their regulatory interactions. 

To evaluate model performance within each cell line, we implement a multi-stage benchmarking procedure (Figure 
\ref{fig:figure2-a} and \ref{fig:figure2-b} A-E). In Figure \ref{fig:figure2-a}A and \ref{fig:figure2-b}A, we perform a one-epoch hyperparameter sweep over combinations of class weights in the loss function and downsampling rates for the negative class. We identify the configuration that yields the highest F1-score and use these parameters for full model training. In Figure \ref{fig:figure2-a}B and \ref{fig:figure2-b}B, we train the model for ten epochs using this optimal configuration and monitor performance on a held-out validation set. Following training, we apply the model to all gene-TF pairs, including those unseen during training, and evaluate our predictions against the corresponding held-out set's gold standard labels (Figure \ref{fig:figure2-a}C and \ref{fig:figure2-b}C). To interpret what drives the resulting performance, we examine the number of genes and TFs represented in the training and test sets for each cell line (Figure \ref{fig:figure2-a}D and \ref{fig:figure2-b}D), providing context for possible domain shifts between splits. Finally, we quantify the class composition of positive and negative examples in the training and test sets (Figure \ref{fig:figure2-a}E and \ref{fig:figure2-b}E), allowing us to relate generalization behavior to the underlying data balance.

Across the six evaluated cell lines, GLM-Prior demonstrates variation in predictive performance, highlighting the influence of data composition, particularly the number of positive regulatory interactions, on generalization (Figure \ref{fig:figure2-a}E and \ref{fig:figure2-b}E). In yeast, GLM-Prior achieves a near-chance AUPRC of $0.02$ compared to a $0.01$ baseline, despite strong within-training metrics (Figure \ref{fig:figure2-a}B). This limited generalization likely stems from the extreme class imbalance in the training data, where only $660$ positive interactions are available among $724,179$ negatives. Although the test set contains a moderate number of positives ($986$), the small number of training examples and high-performing validation metrics suggest that the model overfits the limited regulatory patterns seen during training and fails to extrapolate to novel gene-TF combinations.

In human embryonic stem cells (hESC), performance improves modestly, with an AUPRC of $0.19$ versus a $0.15$ chance baseline. Here, GLM-Prior benefits from a larger pool of positive training examples ($5,305$) and a more balanced test set ($53,337$ positives and $318,730$ negatives). However, TF coverage drops from $504$ in training to $79$ in testing, implying that while model generalizes across genes, reduced TF diversity in the test set limits broader generalization. 

In HepG2 hepatocytes, GLM-Prior achieves a stronger performance (AUPRC $= 0.30$ versus $0.25$ chance) supported by both a higher number of positive training examples ($8,087$) and a well-represented test set ($56,567$ positives). Despite the test set containing fewer TFs ($53$ versus $527$ in training), the model still generalizes effectively, indicating that it learns transferable sequence features that extend across TFs in this cell line. This performance, coupled with the relatively large and balanced training data, suggest that sufficient label diversity allows the model to learn regulatory sequence logic that is not confined to specific TF identities. 

In mouse embryonic stem cells (mESC), GLM-Prior achieves an AUPRC of $0.23$ versus $0.20$ chance, showing moderate but consistent generalization. The training set includes $3,129$ positives among $476,260$ negatives while the test set is considerably more balanced ($65,933$ positives, $271,015$ negatives), covering $5,711$ genes unseen during training. The broad gene coverage supports cross-gene generalization, though limited positive training examples likely cap performance relative to the human cell lines.

In mouse dendritic cells (mDC), performance remains at chance (AUPRC $= 0.09$, despite a substantial number of training examples $11,551)$. The sharp reduction in TF coverage from $462$ in training to only $29$ in testing suggests that the test TFs may regulate distinct, cell-type specific programs absent from training.

Finally, mouse hematopoietic stem cells (mHSC) exhibit the strongest generalization (AUPRC $= 0.49$ versus $0.37$ chance). Although the training set contains a modest number of positives ($2,808$), the test set is both large ($182,005$ positives) and diverse ($6,777$ genes, $72$ TFs), providing a robust evaluation. The high AUPRC gain indicates that GLM-Prior captures sequence-level regulatory features that transfer effectively to unseen gene-TF pairs in this cell line.

Together, these results demonstrate diverse composition of gene-TF relationships in training data is a critical factor behind GLM-Prior's generalization. In particular, generalization is most affected by the abundance and diversity of positive interactions and TFs in training. We interpret the above chance performance gains as the model's ability to learn transferable sequence regulatory patterns when sufficient label diversity is available.

\subsection{GLM-Prior supports successful species-transfer learning and multi-species training}

\begin{figure}[!htbp]
    \centering
    \includegraphics[width=1.0\textwidth]{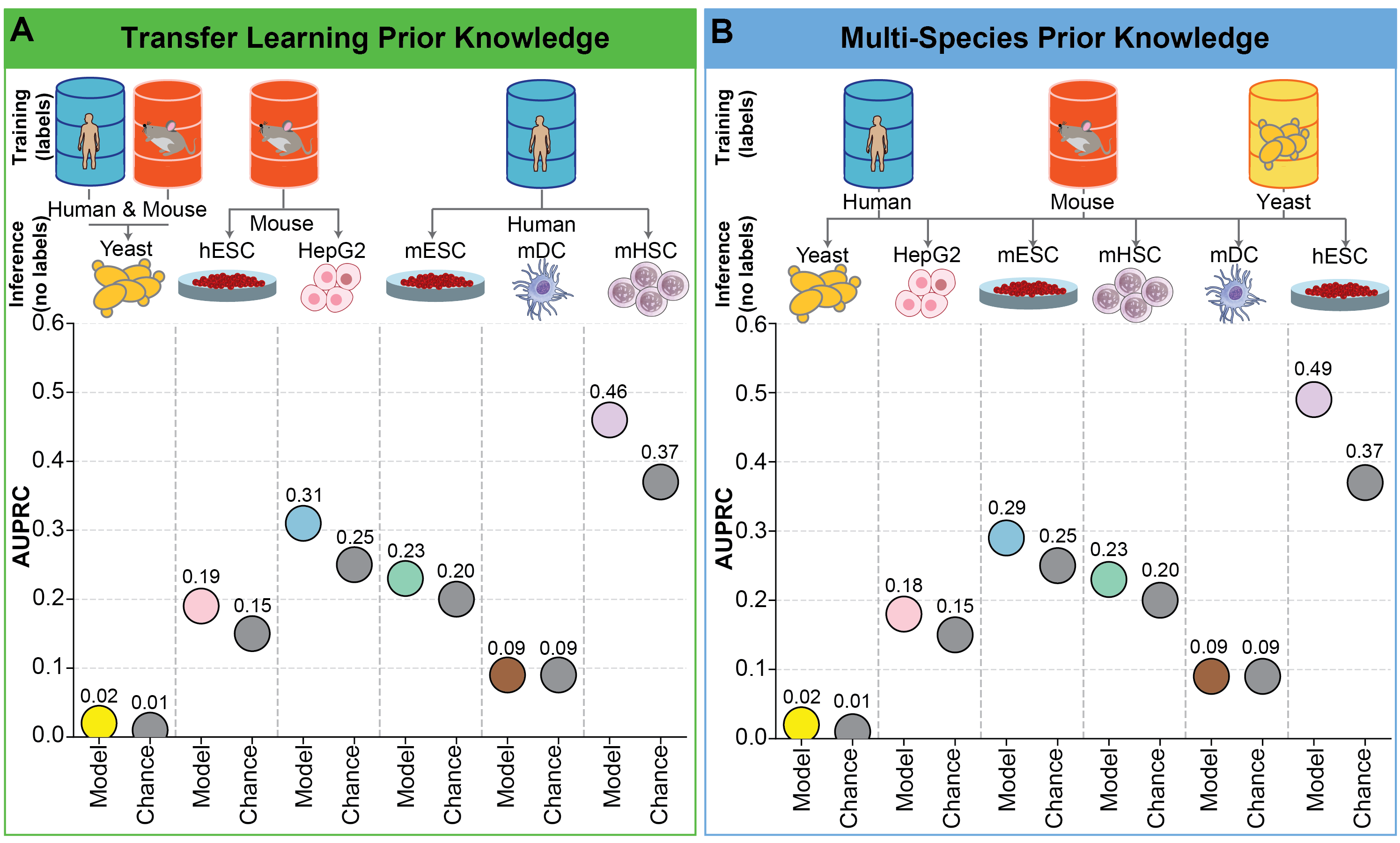} 
    \caption{Transfer learning and multi-species training with GLM-Prior. (\textbf{A}) Schematic and AUPRC results of the species-transfer learning setup, showing three configurations: $(i)$ a model trained jointly on human and mouse data evaluated on yeast, $(ii)$ a mouse-only model evaluated on human cell lines, and $(iii)$ a human-only model evaluated on mouse cell lines. (\textbf{B}) Schematic and AUPRC results of the multi-species model trained sequentially on human, mouse and yeast data, evaluated across six cell lines.}
    \label{fig:figure3-a}
\end{figure}

\begin{figure}[!htbp]
    \centering
    \includegraphics[width=1.0\textwidth]{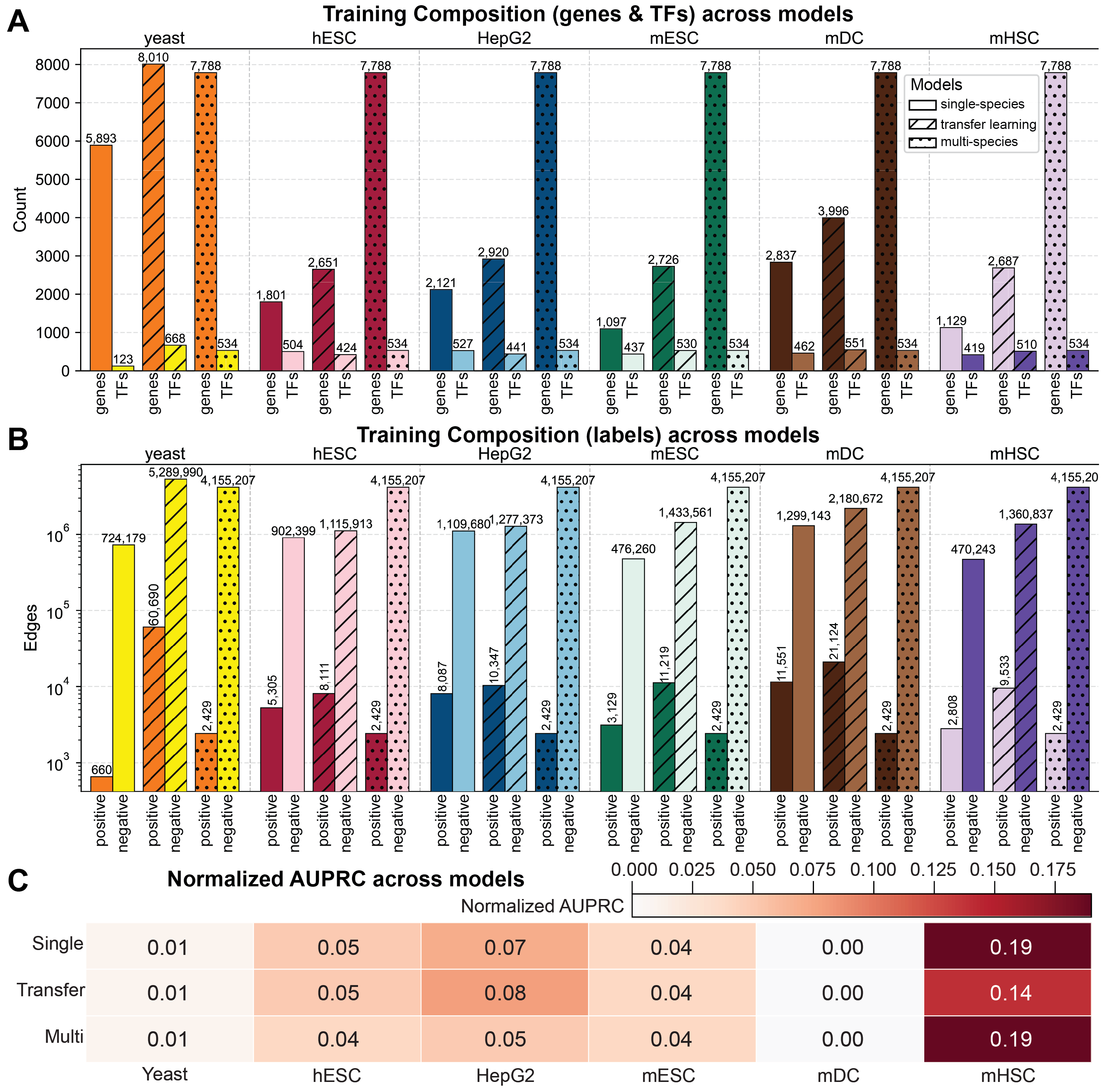} 
    \caption{Transfer learning and multi-species training with GLM-Prior continued. (\textbf{A}) Comparison of training composition (number of genes and TFs) for single-species, species-transfer, and multi-species models. (\textbf{B}) Class composition (positive and negative labels) in the corresponding training sets. (\textbf{C}) Normalized AUPRC across single-species, species-transfer, and multi-species models for each cell line.}
    \label{fig:figure3-b}
\end{figure}

Following our six cell line evaluation of GLM-Prior, we next task the model to predict prior knowledge in a species unseen during training (Figure \ref{fig:figure3-a} and \ref{fig:figure3-b}). Building on the previous section, which established that generalization depends strongly on data composition, we now ask whether the model can transfer learned regulatory logic across species. Such cross-species transfer would represent a major advantage over existing approaches, enabling the construction of informative priors for understudied organisms using models trained on evolutionarily similar species. To test this, we perform two complementary experiments: A) transfer learning, where models trained in one species are applied to another, and B) multi-species training, where a single model is trained sequentially on human, mouse, and yeast data. In all cases, the held-out evaluation sets are fully disjoint from training data, and overlapping gene and TF identifiers are removed to prevent cross-species leakage.

In Figure \ref{fig:figure3-a}A, we provide a schematic for the transfer learning setup. Three transfer learning configurations are evaluated, $(i)$ a model trained jointly on human and mouse data, evaluated on yeast, $(ii)$ a mouse-only model evaluated on human cell lines (hESC and HepG2), and $(iii)$ a human-only model evaluated on mouse cell lines (mESC, mDC, and mHSC). Across all six cell lines, we find that transfer learning maintains or modestly improves performance relative to single-species models. 

In yeast, performance remains near chance (AUPRC $=0.02$ vs $0.01$), consistent with deep evolutionary divergence between eukaryotic and mammalian regulatory syntax. In contrast, cross-species transfer between human and mouse demonstrates clear success, where hESC maintains an AUPRC $= 0.19$ and HepG2 improves to $0.31$. Simiarly, human-trained models applied to mouse cell lines achieve stable performance comparable to single-species results. These results demonstrate that GLM-Prior captures conserved sequence-level features that transfer across closely related species, enabling accurate prior construction without species-specific retraining.

In Figure \ref{fig:figure3-a}B, we present the multi-species model trained sequentially on human, mouse, and yeast data and evaluated on all six cell lines. The multi-species model achieves comparable AUPRCs relative to both single-species and species-transfer settings (yeast = $0.02$, hESC = $0.19$, HepG2 = $0.29$, mESC = $0.23$, mDC = $0.09$, and mHSC = $0.49$). 

To interpret these trends, Figure \ref{fig:figure3-b}A-B summarize the training compositions and normalized AUPRC (Figure \ref{fig:figure3-b}C) across single, transfer, and multi-species setups. Across each cell line, transfer learning consistently expands the number of genes, TFs, and positive labels seen during training. In some settings, such as with HepG2, this broader coverage allows for a modest normalized AUPRC increase ($0.08$ vs $0.07$ single-species). In others, transfer learning maintains single-species performance. In contrast, mHSC shows a slight decrease in performance ($0.14$ species-transfer vs $0.19$ single-species) despite larger training breadth. This drop suggests that the regulatory logic active in mouse HSCs is relatively cell-type and species-specific and therefore not well represented in the context-agnostic human interaction data used for species-transfer training, limiting the model's ability to fully recover mHSC-specific edges. 

Notably, the multi-species model maintains similar performance to the single-species models with only slight performance drops in hESC and HepG2. Here, we demonstrate that multi-species training yields a robust, domain-stable prior that generalizes across lineages. This provides a practical route to build prior knowledge for poorly understood systems or cell lines when a well-diversified training set is desired, offering broad regulatory coverage without assuming a species specific match.

Together, these results demonstrate that GLM-Prior achieves successful cross-species transfer, a property not attainable with accessibility-based prior knowledge approaches which depend on species-specific data. Applying a model trained in one organism directly to another enables scalable, sequence-based prior construction in under-characterized systems. In parallel, multi-species training yields a robust, lineage-agnostic prior that maintains stable performance across contexts. When the target cell line is poorly characterized or its closest species match is uncertain, the multi-species model presents a practical default for constructing reliable prior knowledge.

\subsection{Integrating GLM-Prior into PMF-GRN provides contextual GRN inference and edge refinement}

To evaluate how sequence-derived regulatory prior knowledge interacts with expression-based GRN inference, we next used each cell line-specific GLM-Prior matrix as input to PMF-GRN. Whereas Stage 1 (Section \ref{sec:GLM-Prior}) directly predicts TF-gene interactions from sequence, Stage 2 (Section \ref{sec:PMF-GRN}) treats these predictions as prior knowledge and refines the network using single cell expression data. For each cell line, we paired GLM-Prior with matched single cell datasets, for example, GSE125162 \cite{jackson2020gene} and GSE144820 \cite{jariani2020new} for yeast, and the corresponding expression datasets provided in BEELINE \cite{pratapa2020benchmarking} for the human and mouse cell lines. This experiment is designed to assess not only whether PMF-GRN improves predictive accuracy, but also how much the network must be altered when inference is initialized from an already information-rich prior.

\begin{figure}[!htbp]
    \centering
    \includegraphics[width=1.0\textwidth]{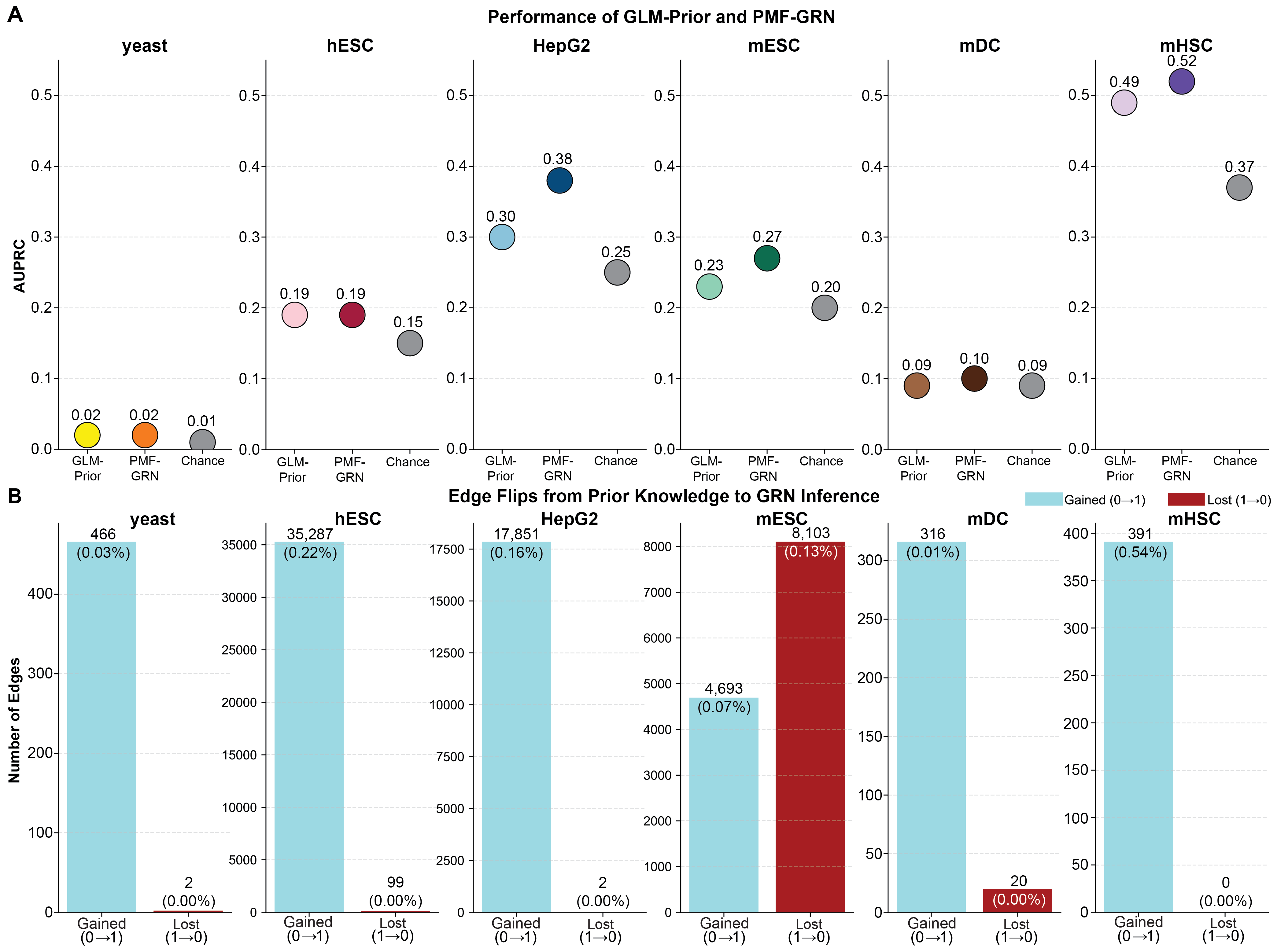} 
    \caption{Integration of GLM-Prior with PMF-GRN for downstream GRN inference. (\textbf{A}) AUPRC comparison of GLM-Prior alone versus integrated with PMF-GRN across six cell line contexts. (\textbf{B}) Edge flip analysis quantifying changes made to the prior knowledge interaction matrix during GRN inference. Blue bars indicate edges gained and red bars indicate edges removed by GRN inference.}
    \label{fig:figure4}
\end{figure}

Across all datasets, integrating GLM-Prior into PMF-GRN results in equal or improved AUPRC relative to the prior alone (Figure \ref{fig:figure4}A). The magnitude of improvement varies by species, where performance remains unchanged in yeast ($0.02$) and hESC ($0.19$), while clear gains are observed in HepG2 ($0.30$ prior $\rightarrow$ $0.38$ GRN), mESC ($0.23$ prior $\rightarrow$ $0.27$ GRN), and mHSC ($0.49$ prior $\rightarrow$ $0.52$ GRN). These improvements occur despite PMF-GRN modifying only a small fraction of the GLM-Prior edge matrix (Figure \ref{fig:figure4}B), with added edges ranging from $0.03\%$ in yeast to $0.54\%$ in mHSC, and edge removals remaining minimal across all contexts. Notably, modifications are strongly biased toward edge additions rather than deletions, indicating that PMF-GRN treats the GLM-Prior as a stable scaffold and makes targeted expansions rather than large-scale refinement.

Interestingly, the extent of structural changes does not strictly track with performance improvement. For example, hESC and mESC exhibit tens of thousands of added or removed edges, yet show minimal differences in AUPRC. This mismatch suggests that many of the changes introduced by PMF-GRN occur outside the held-out evaluation portion of the network, which only cover a subset of TF-gene interactions. Conversely, in yeast, where GLM-Prior is close to chance (AUPRC $=0.02$) and the available training interactions are sparse, PMF-GRN fails to recover meaningful signal despite access to two large single-cell expression datasets, indicating that expression-driven inference alone cannot compensate for a weak or uninformative prior.

These findings indicate that when a sequence-derived prior already captures substantial regulatory signal, the role of expression-based inference shifts from discovery to contextual refinement. Rather than constructing the network from expression patterns alone, PMF-GRN behaves conservatively, preserving high-confidence structure, adding plausible missing edges, and performing limited pruning. In this dual-stage framework, the strength of GLM-Prior largely determines the achievable performance ceiling, while PMF-GRN contributes selective improvements where transcriptomic evidence provides additional support; when the prior is poor, as in yeast, expression-based GRN inference has limited capacity to recover accurate regulatory structure.

\subsection{Comparing prior construction strategies across yeast, mouse, and human cell lines}
\label{sec:prior-method-comparison}
Next, we assess how GLM-Prior compares to widely used accessibility-based prior construction approaches, evaluating performance relative to CellOracle and Inferelator priors across six cell line contexts.  Unlike GLM-Prior, which leverages sequence-level information to predict TF-gene regulatory edges, both CellOracle and Inferelator derive putative interactions from chromatin accessibility and motif enrichment. Importantly, these accessibility-based priors are constructed using ATAC-seq measured in the matched cell line, whereas GLM-Prior is trained on context-agnostic interaction resources (STRING and TRRUST) and evaluated without explicitly conditioning on cell-line specific chromatin state. This comparison therefore evaluates whether sequence-derived priors can match or exceed the performance of accessibility-based approaches across diverse biological settings. 

\begin{figure}[!htbp]
    \centering
    \includegraphics[width=1.0\textwidth]{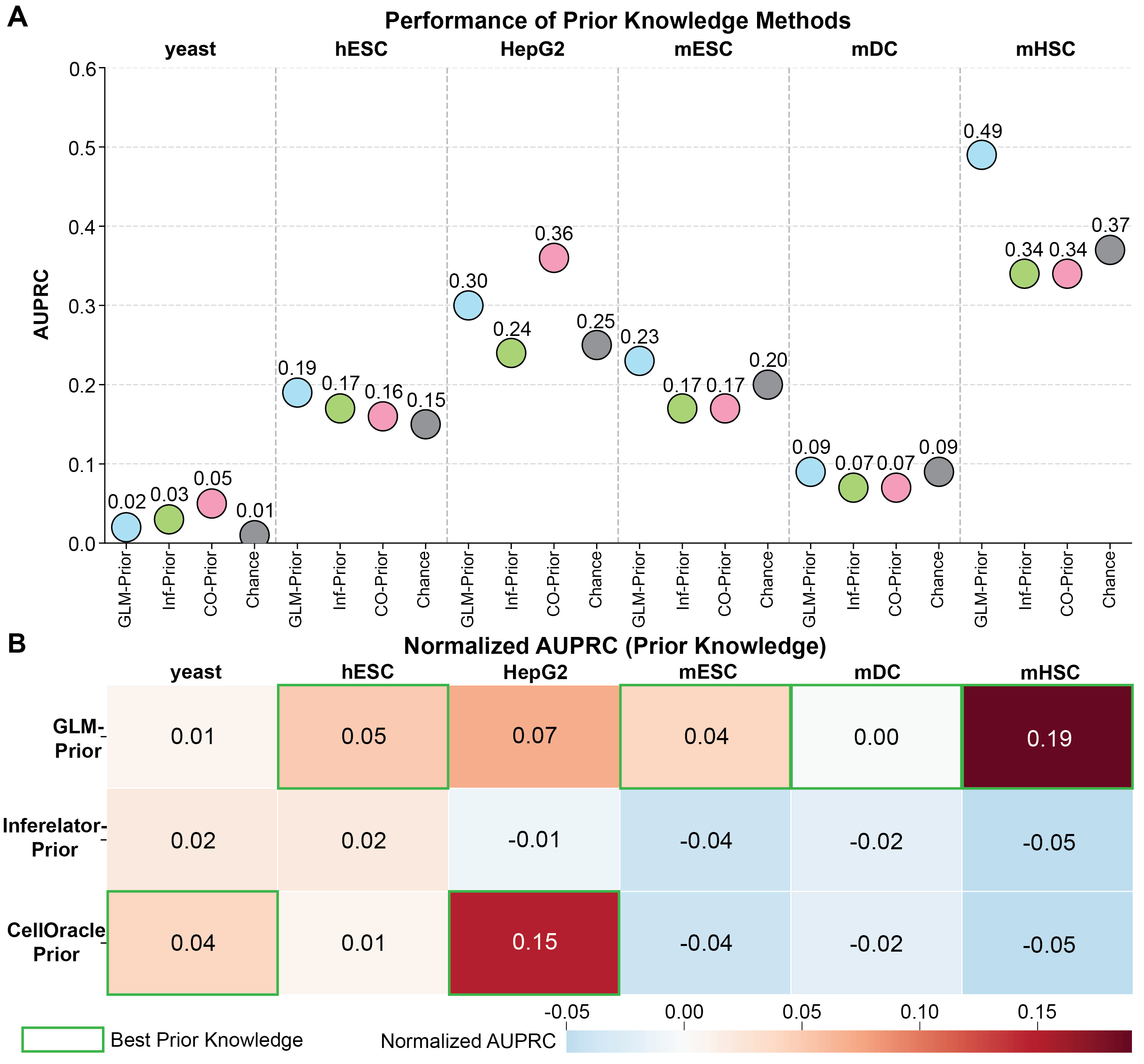} 
    \caption{Comparison of prior construction strategies across species. (\textbf{A}) AUPRC of GLM-Prior, Inferelator-Prior and CellOracle Prior across six cell lines, with chance performance (gray) provided for each context. (\textbf{B}) Normalized AUPRC values with the best-performing prior (green box) highlighted in each cell line.}
    \label{fig:figure5}
\end{figure}

Across all six cell lines, GLM-Prior performs at or above chance, exceeding the chance accuracy in five of the six contexts, demonstrating robust generalization across species and cell types (Figure \ref{fig:figure5}A). In comparison, Inferelator-Prior performs above chance in only two out of six cell lines, and CellOracle's Prior achieves above-chance performance in three out of six, indicating less consistent cross-context behavior. Yeast represents the main exception, where CellOracle produces the highest-performing prior (AUPRC $0.05$ versus $0.03$ for Inferelator-Prior and $0.02$ for GLM-Prior), highlighting the strengths of accessibility-based priors in simple eukaryotic systems where regulatory interactions are predominantly promoter-proximal and well captured by open chromatin-based assignment. In this setting, GLM-Prior is further disadvantaged by the limited number of positive regulatory interactions ($660$) available for model training (as seen in Figure \ref{fig:figure3}D).

A broader comparison across species using normalized AUPRC values further illustrates these patterns (\ref{fig:figure5}B), where normalized AUPRC is defined as $(\text{AUPRC} - \text{chance}) / (1 - \text{chance})$. While GLM-Prior provides the most consistent gains overall, CellOracle's prior outperforms all other approaches in HepG2 ($0.36$ versus $0.30$ for GLM-Prior and $0.24$ for Inferelator-Prior). This comparatively strong performance in a mammalian cell line can be explained by alignment between the construction of CellOracle's prior and the BEELINE reference network used for evaluation. BEELINE assembles reference networks using ChIP-seq supported TF-gene edges curated from ENCODE \cite{de2012encode}, ChIP-ATLAS \cite{zou2024chip} and ESCAPE \cite{xu2013escape} databases, where HepG2 is among the most densely profiled ENCODE cell line. As a result, the HepG2 reference likely contains a large and internally consistent set of TF occupancy-supported edges that are recoverable by accessibility-conditioned, proximity-based linking, precisely the evidence CellOracle prioritize when constructing its prior from motifs in accessible regions. By contrast, Inferelator-Prior follows a similar accessibility and motif-based strategy but imposes stronger sparsity which may remove a substantial number of true TF-gene edges in this densely profiled HepG2 context and thereby reduce its alignment with the BEELINE reference network.

In contrast, GLM-Prior is the top-performing method in the four remaining cell lines: hESC, mESC, mDC, and mHSC. In hESC, all methods exceed chance, with GLM-Prior achieving the highest AUPRC ($0.19$ versus $0.17$ for Inferelator-Prior and $0.16$ for CellOracle's prior). In mESC and mHSC, the advantage of GLM-Prior becomes more pronounced, as accessibility-based priors fall below chance while GLM-Prior remains predictive. These results suggest that in several mammalian contexts, sequence-based priors may better capture distal enhancer regulation and long-range TF-gene interactions that proximity-based methods miss.

These comparisons indicate that accessibility-driven approaches such as CellOracle remain highly effective in contexts dominated by proximal regulatory logic, whereas GLM-Prior provides a more robust and transferable strategy across diverse mammalian systems. The complementary performance profiles across species highlight that no single prior construction approach is universally optimal; instead, GLM-Prior offers a scalable, sequencing technology-independent alternative that is better suited to capturing complex long-range regulatory architecture and transferring information across closely related species and cell lines. 

\subsection{Paired prior-GRN strategies reveal prior-dependent gains from expression-based inference}

\begin{figure}[!htbp]
    \centering
    \includegraphics[width=1.0\textwidth]{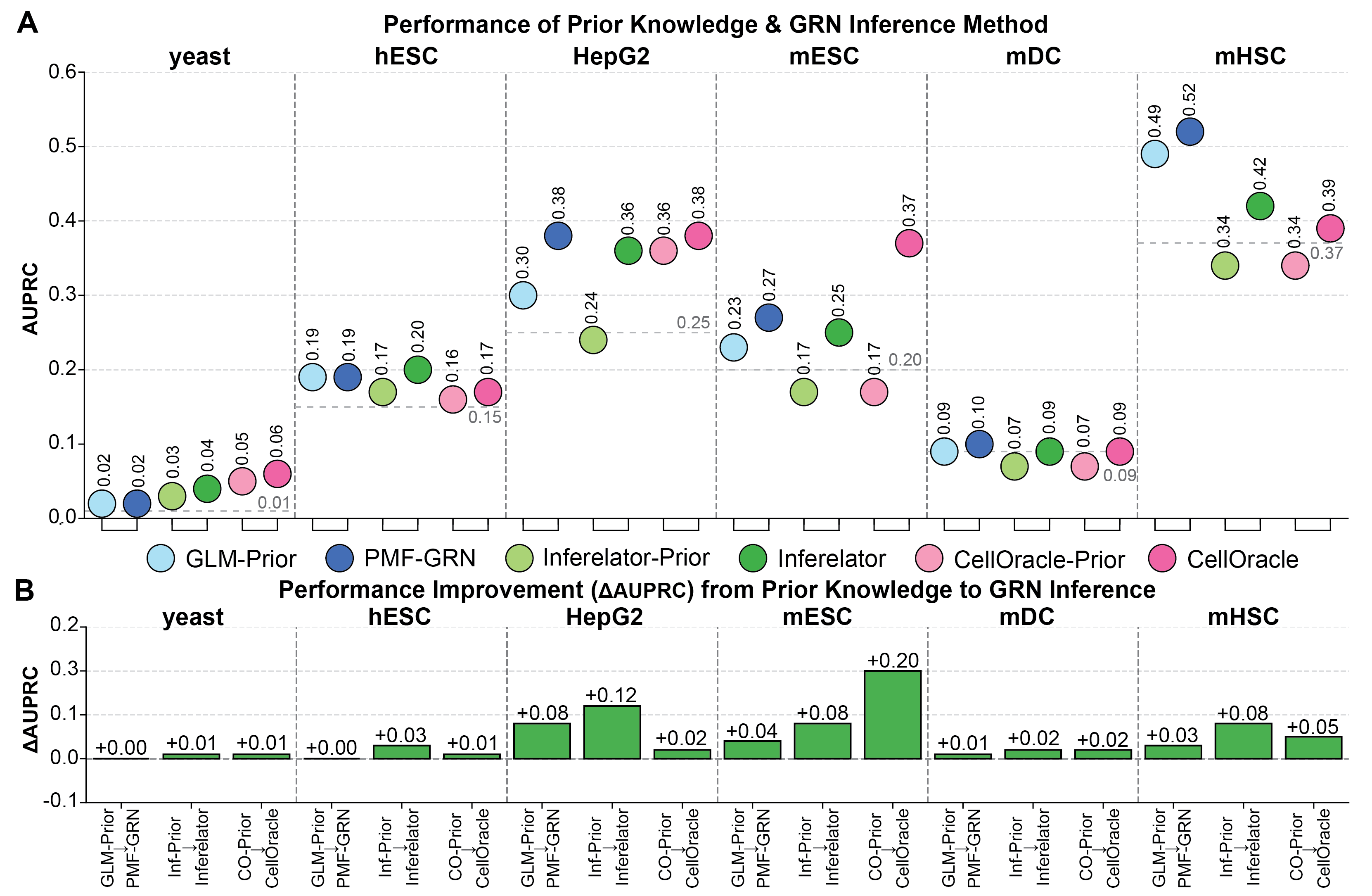} 
    \caption{Paired prior-GRN strategies reveal prior dependent gains from expression-based GRN inference across six cell lines. (\textbf{A}) AUPRC of each prior (GLM-Prior, Inferelator-Prior, and CellOracle's prior) and its corresponding GRN (PMF-GRN, Inferelator, CellOracle) across six cell lines, with gray dashed lines indicating the chance level in each species. (\textbf{B}) Change in AUPRC from prior to GRN ($\Delta$AUPRC) for each paired method, quantifying the added value of expression-based inference.}
    \label{fig:figure6-a}
\end{figure}

\begin{figure}[!htbp]
    \centering
    \includegraphics[width=1.0\textwidth]{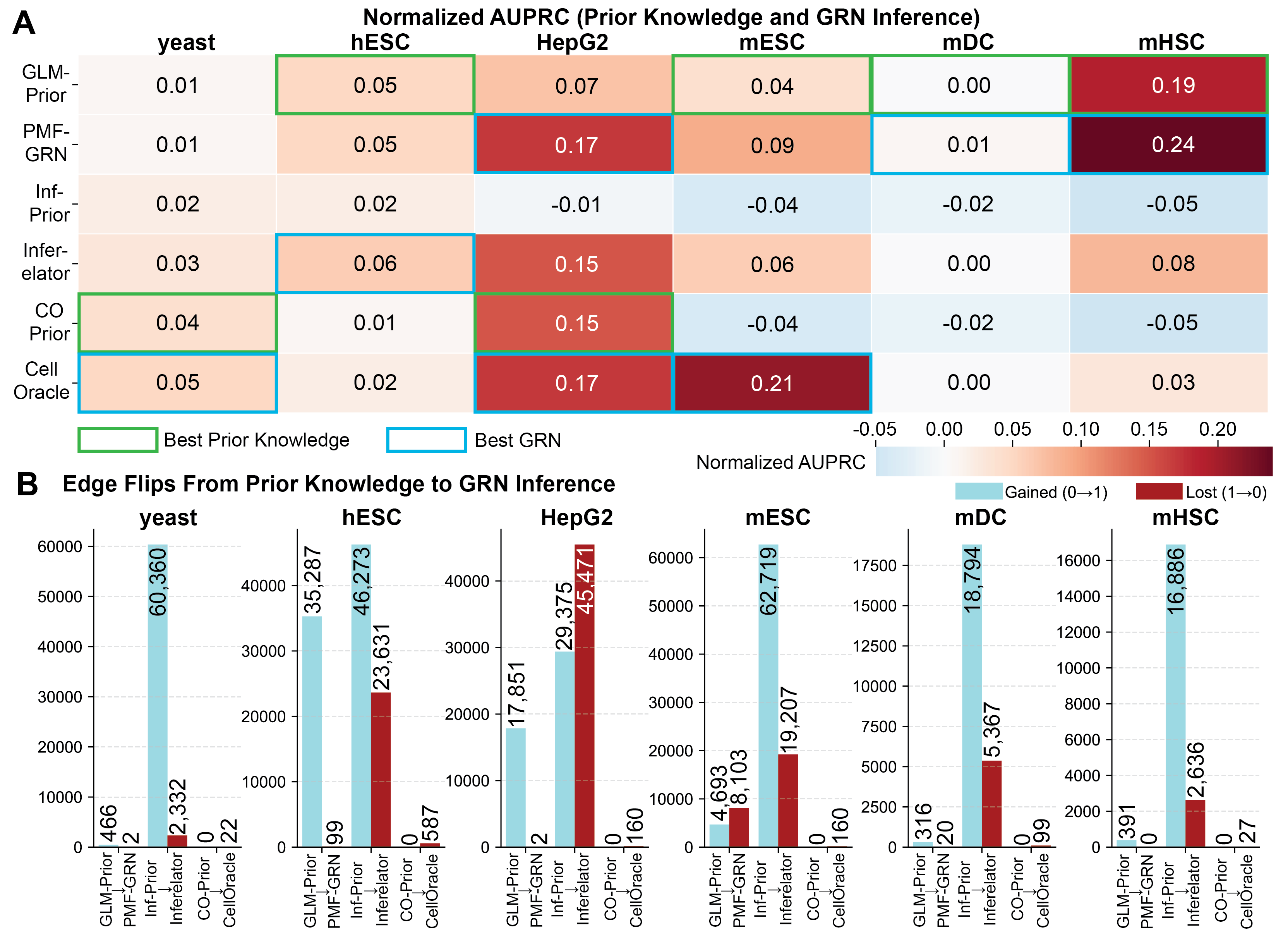} 
    \caption{Paired prior-GRN strategies reveal prior dependent gains from expression-based GRN inference across six cell lines continued. (\textbf{A}) Normalized AUPRC performance for priors and GRNs in each cell line. Green boxes highlight the best prior and blue boxes highlight the best GRN per cell line. (\textbf{B}) Edge flips between prior and GRN for each method demonstrating how GRN inference methods modulate their priors.}
    \label{fig:figure6-b}
\end{figure}

Following our comparative analysis of prior construction approaches, we next asked how each complete prior-GRN pipeline performs across the six cell lines (Figure \ref{fig:figure6-a} and \ref{fig:figure6-b}). GLM-Prior and PMF-GRN form our dual-stage pipeline, combining a finetuned sequence language model for prior construction with a probabilistic matrix factorization model for GRN inference from single cell expression. CellOracle constructs a prior ("CellOracle baseGRN") using motif enrichment within accessible promoter regions and pairs it with a Bayesian ridge regression model for GRN inference. Similarly, Inferelator-Prior uses chromatin accessibility and TF motif enrichment to build a prior edge matrix that is then refined using regularized regression during GRN inference. By comparing these three paired pipelines, GLM-Prior + PMF-GRN, Inferelator-Prior + Inferelator, and CellOracle prior + CellOracle, we can ask two related questions: $(i)$ how well each method converts its own prior into a high-performing GRN, and $(ii)$ to what extent expression-based GRN inference can compensate for poor priors versus refining already strong ones.

To interpret these patterns, we consider the absolute AUPRCs in Figure~\ref{fig:figure6-a}A together with the corresponding $\Delta$AUPRC values in Figure~\ref{fig:figure6-a}B. In yeast and mDC, all priors are at or only slightly above chance, and GRN inference yields minimal gains ($\Delta$AUPRC $\leq 0.02$), indicating that expression-based models struggle to recover meaningful signal when the starting network is near-random. In hESC and mHSC, GLM-Prior already achieves above-chance performance, and PMF-GRN adds only small improvements, whereas Inferelator and CellOracle modestly refine their weaker priors. These contexts illustrate that when priors are already strong, GRN inference primarily fine-tunes edge weights rather than substantially reshaping the network.

By contrast, HepG2 and mESC are the settings where GRN inference provides the largest benefits. Here, priors span a range from near- or slightly below-chance (Inferelator-Prior in HepG2; Inferelator-Prior and CellOracle prior in mESC) to moderately above-chance (GLM-Prior and CellOracle prior in HepG2; GLM-Prior in mESC), leaving partially specified structure that expression data can refine. In HepG2, GLM-Prior + PMF-GRN and Inferelator-Prior + Inferelator show sizeable increases in AUPRC, while CellOracle begins from the strongest prior and improves only modestly. In mESC, regression-based inference on top of weak or near-chance priors yields the single largest observed $\Delta$AUPRC for CellOracle and substantial gains for Inferelator, whereas PMF-GRN makes only a small but consistent improvement over an already above-chance GLM-Prior. Taken together, these six cell lines reveal a coherent pattern: the magnitude of GRN-inference gains is strongly prior-dependent, with the largest boosts arising when priors are of intermediate quality (informative but incomplete), limited performance improvements when priors are very weak, and only modest refinements when priors are already strong.

We next normalize AUPRC by the chance level in each species (as in Section \ref{sec:prior-method-comparison}) to directly compare performance above (or below) chance across methods and cell lines (Figure \ref{fig:figure6-b}A). In this view, GLM-Prior provides the best prior in four of six cell lines (hESC, mESC, mDC, mHSC), while CellOracle prior yields the strongest priors in yeast and HepG2. We observe that in four mammalian cell lines (HepG2, mESC, mDC, and mHSC), Inferelator-Prior rarely exceeds random-chance performance and typically remains close to or below chance. Similarly, in all three mouse cell lines, CellOracle's prior does not construct a prior above chance. Here, complex and long-range regulatory interactions may not be sufficiently captured by these proximity-based approaches, highlighting a key limitation of accessibility-based prior knowledge in complex mammalian cell lines. 

In addition, we find that the best GRNs largely track the best priors, with PMF-GRN producing the top-performing GRN in three of six cell lines, CellOracle in two, and Inferelator in one (Figure~\ref{fig:figure6-b}A). Notably, there are instances where GRN inference can partially compensate for a weaker prior. In mESC, CellOracle's prior is below chance (normalized AUPRC $= -0.04$), yet its regression-based GRN achieves the highest normalized performance across all methods ($0.21$), outperforming GLM-Prior + PMF-GRN despite starting from an inferior prior. This behavior is clarified by examining how each GRN model modulates its prior in terms of edge flips (Figure~\ref{fig:figure6-b}B).

PMF-GRN primarily improves performance by adding edges to the prior, effectively expanding the regulatory network in a data-informed way while leaving most high-confidence edges intact. Inferelator both adds and removes edges, suggesting a balance between edge discovery and denoising. In contrast, CellOracle's regression-based design only allows removal of edges: it can shrink edge weights to zero but cannot introduce new edges, discovered through expression data, that were absent from the accessibility-derived prior. 

The large performance gains observed for CellOracle in settings such as mESC therefore arise from aggressive pruning of false-positive edges in a dense, noisy prior, rather than from discovery of novel regulatory interactions. This pruning behavior can make a weak but overly dense prior appear substantially stronger after GRN inference, yet also highlights an important limitation: when single-cell expression is used only to filter a fixed network, not to propose new edges, it cannot expand the regulatory scaffold beyond what is already encoded in the prior.

Together, these results demonstrate that while GRN inference can substantially improve performance in some settings, particularly when priors are informative but incomplete, the overall ranking of methods is largely dictated by prior quality. Expression-based GRN inference models mostly refine and re-weight the structure supplied by the prior, rather than overturning it, underscoring the central role of prior construction in determining GRN reconstruction performance.

\subsection{Cross-method comparison disentangles prior and GRN inference contributions to performance}

\begin{figure}[!htbp]
    \centering
    \includegraphics[width=1.0\textwidth]{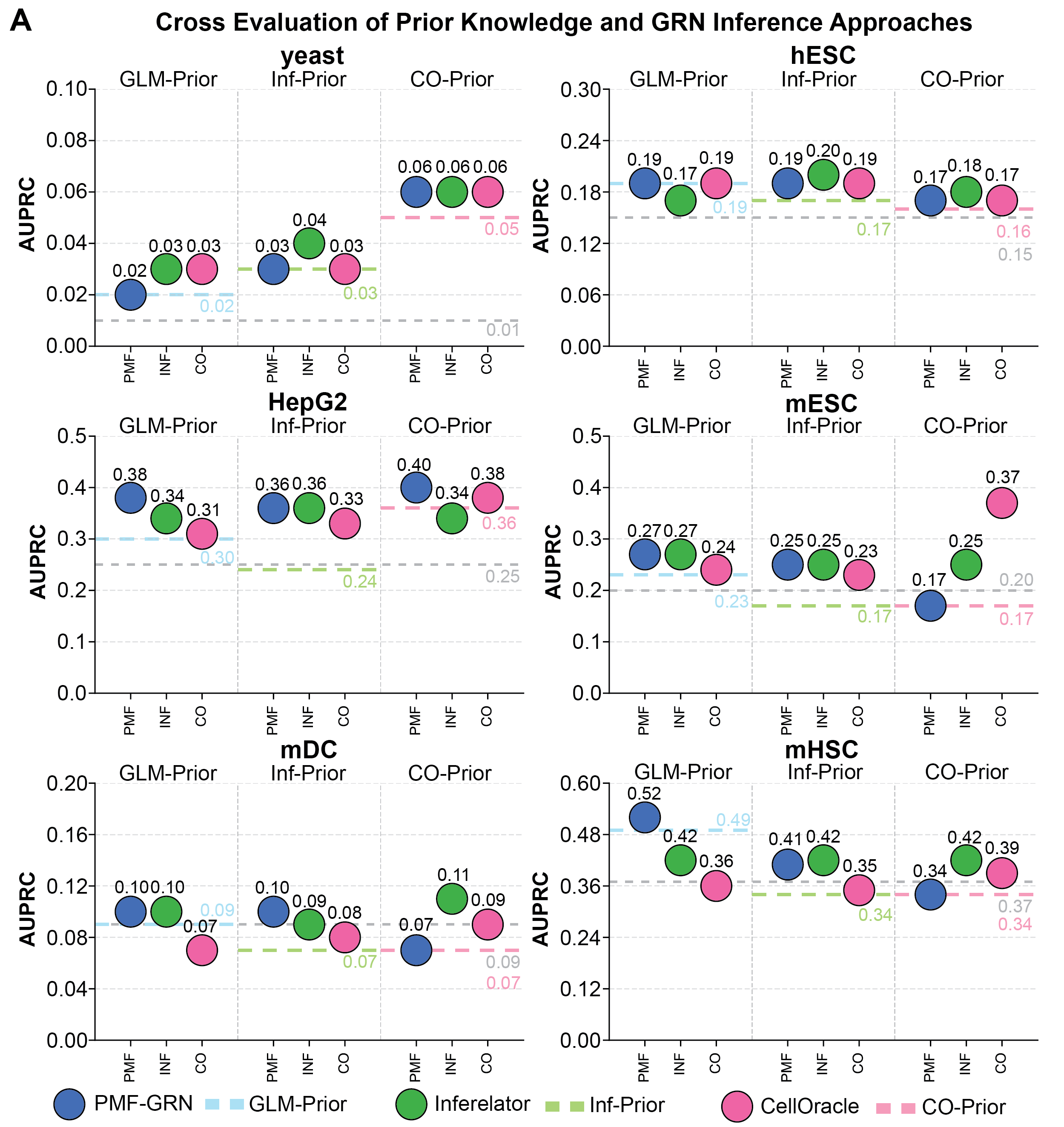} 
    \caption{Cross-method comparison disentangles prior and GRN inference contributions. \textbf{(A)} AUPRC for all nine combinations of prior and GRN inference methods across six cell lines. Gray dashed lines indicate chance performance, while light blue, light green, and light pink dashed lines represent baseline performance of GLM-Prior, Inferelator-Prior, and CellOracle prior, respectively, across each cell line.}
    \label{fig:figure7-a}
\end{figure}

\begin{figure}[!htbp]
    \centering
    \includegraphics[width=1.0\textwidth]{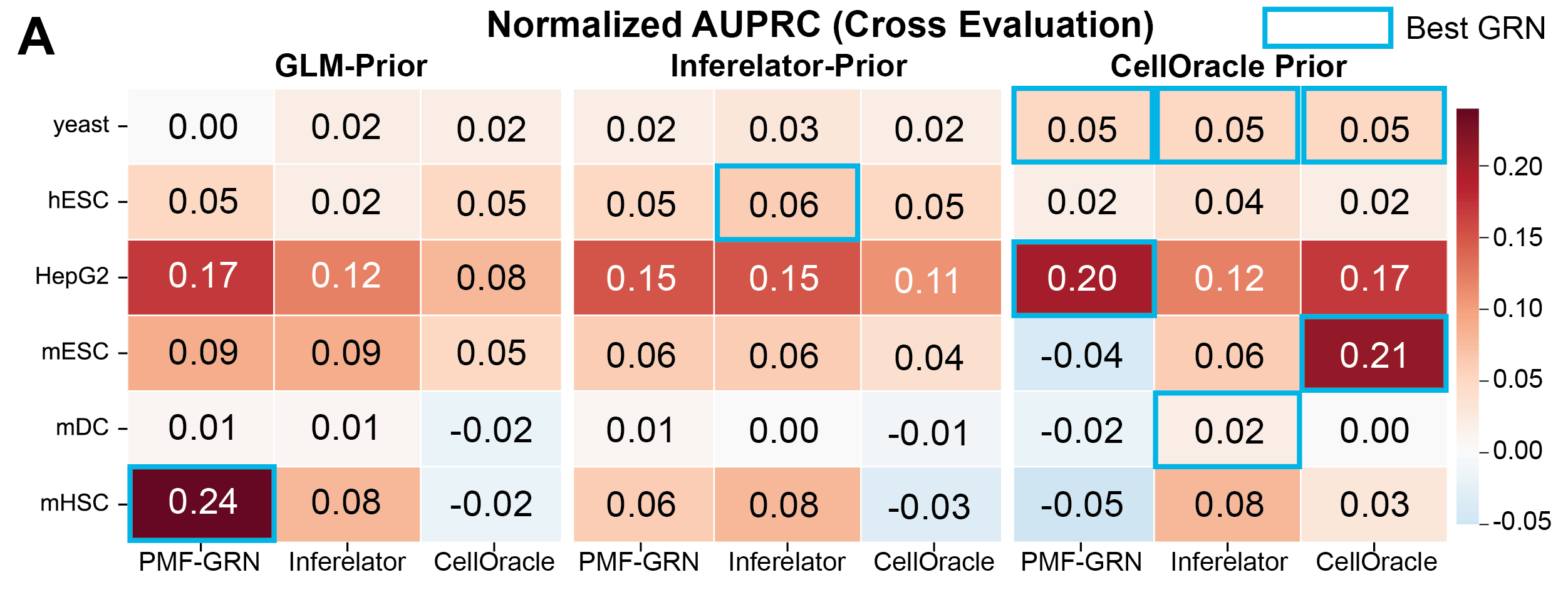} 
    \caption{Cross-method comparison disentangles prior and GRN inference contributions continued. \textbf{(A)} Normalized AUPRC of all nine combinations of prior and GRN for directly comparing across method and cell line. Blue boxes indicate highest GRN performance per cell line.}
    \label{fig:figure7-b}
\end{figure}

To disentangle the contribution of prior construction and GRN inference from the GRN methods themselves, we next performed a full cross-comparison in which each GRN inference method is applied to each prior across all six cell lines (Figure \ref{fig:figure7-a} and \ref{fig:figure7-b}). Specifically, we combine GLM-Prior, Inferelator-Prior, and CellOracle baseGRN with PMF-GRN, Inferelator, and CellOracle, yielding nine prior-GRN combinations per cell line. In Figure \ref{fig:figure7-a}A, we show the resulting AUPRC values, along with species-specific chance levels (gray dashed line), and the baseline performance of the prior (colored dashed line) used for GRN inference. In Figure \ref{fig:figure7-b}A, we report the same results normalized relative to chance, where values above zero indicate performance above random. We focus our interpretation on the normalized AUPRC values, which allows for easier comparison across species and methods, and refer to the raw AUPRCs in Figure \ref{fig:figure7-a}A where relevant.

Across species, the dominant pattern is that the identity of the best GRN for a given cell line is largely determined by the underlying prior, not by the choice of GRN inference method. In yeast, for example, any GRN method paired with CellOracle's prior attains the same normalized AUPRC ($0.05$), whereas all combinations using GLM-Prior or Inferelator-Prior remain at or below $0.03$. Once CellOracle's prior is selected, the choice among PMF-GRN, Inferelator, or CellOracle has no quantifiable impact, indicating that the prior itself, rather than the choice of downstream inference algorithm, is the bottleneck of GRN inference performance.

Similar patterns appear in the mammalian cell lines. In hESC, GLM-Prior and Inferelator-Prior yield the strongest priors ($0.19$ and $0.17$ AUPRC, respectively), and nearly all GRNs inferred from these priors achieve competitive performance. In HepG2, CellOracle's prior is best (AUPRC $=0.36$), and the two highest performing GRNs are obtained by pairing this prior with PMF-GRN or CellOracle, demonstrating again that performance is largely determined by the prior rather than the choice of inference algorithm. An analogous effect is observed in mHSC, where GLM-Prior combined with PMF-GRN yields the highest performing combination (AUPRC $=0.24$), while all GRNs built on Inferelator-Prior or CellOracle's prior plateau at $0.08$ or lower. These results support a consistent conclusion, the quality and structure of the prior largely dictate the achievable performance, and the GRN inference methods primarily modulate performance within a range set by the prior.

The full cross-comparison design also allows us to ask whether any GRN inference method is inherently superior across priors and species. We find that across PMF-GRN, Inferelator, and CellOracle, the overarching evidences suggests no GRN inference method is superior. Each approach wins within different regimes defined by the combination of species and prior. PMF-GRN delivers the highest normalized performance in HepG2 when paired with CellOracle's prior ($0.20$) and in mHSC when paired with GLM-Prior ($0.24$). PMF-GRN also performs well in yeast when paired with CellOracle's prior, and mESC when paired with GLM-Prior or Inferelator-Prior. The Inferelator is best in hESC when paired with Inferelator-Prior ($0.06$) and in mDC when paired with CellOracle's prior ($0.02$), and remains competitive in HepG2 across multiple priors. CellOracle is uniquely optimal in mESC when paired with CellOracle's prior ($0.21$), substantially outperforming all other prior-GRN combinations in that species. The ranking of each GRN inference method thus changes with both prior and cell line, consistent with our claim that GRN methods are context-dependent refiners of information encoded in a prior using expression data, rather than the primary driver of performance with a single universally superior algorithm.

The cross-method comparison further reveals characteristic interactions between each GRN inference method and the priors. PMF-GRN appears highly prior sensitive, where switching priors can produce large swings in performance within a given species. In mHSC, for instance, normalized performance for PMF-GRN ranges from $-0.05$ with CellOracle-Prior to $0.24$ with GLM-Prior. In mESC, performance ranges from $-0.04$ with CellOracle Prior to $0.09$ with GLM-Prior. This pattern indicates that PMF-GRN is very efficient at exploiting high-quality priors, but struggles to rescue very weak ones. The Inferelator, by contrast, is comparatively robust to prior choice. Within each species, the variation in normalized AUPRC across priors is typically modest (often within the range of $0.02$ to $0.04$). In mHSC, all three priors yield identical normalized performance ($0.08$) with the Inferelator. This suggests that the Inferelator leans more heavily on expression-based regularized regression, making it less sensitive to exact prior structure, but also preventing it from reaching the highest performance in most cell lines. CellOracle exhibits strong synergy with its own prior in specific settings. For example, in mESC, CellOracle's prior + Cell Oracle perform well ($0.21$), significantly outperforming all other combinations, indicating that the regression-based pruning used by CellOracle is particularly well matched when the prior necessitates the removal of false positive edges.

As each GRN inference method is applied to every prior, we can also ask whether cross-pairing ever allows a "strong" GRN inference method to elevate a weaker prior above combinations that use a stronger prior with a different inference method. Such reversals are rare. In mHSC, for example, no method applied to Inferelator-Prior or CellOracle prior achieves performance comparable to GLM-Prior + PMF-GRN. In HepG2, there is a modest example of beneficial cross-pairing, where CellOracle prior + PMF-GRN achieves an AUPRC of $0.20$, outperforming the original CellOracle prior + CellOracle pairing ($0.17$). The performance gains here are likely due to PMF-GRN providing additional edges to the CellOracle's prior using expression data. In mDC, CellOracle prior + Inferelator ($0.02$) is better than CellOracle prior + CellOracle ($0$) or CellOracle prior + PMF-GRN ($0.02$), although all values remain near chance. These exceptions are incremental, not complete reversals of the prior-driven ranking, and reinforce the view that GRN inference can fine-tune performance but typically cannot turn a weak prior into one that rivals the best priors.

Overall, the cross-method comparison in Figure \ref{fig:figure7-a} and \ref{fig:figure7-b} supports three main conclusions. First, prior quality remains the primary determinant of GRN reconstruction performance, even when each GRN method is given access to each prior. Second, there is no universally better GRN inference algorithm: PMF-GRN, the Inferelator, and CellOracle perform best in different prior and species regimes, reflecting differences in how they use expression data to refine or re-weight the prior. Third, GRN inference methods modulate, rather than overturn, the structure imposed by the prior. Changing the inference algorithm can yield meaningful gains in some contexts, particularly when paired with an appropriate prior, but rarely elevates a weak prior to the level of a strong one. These findings reinforce our claim that prior construction is the main driver of GRN inference performance, with expression-based GRN inference primarily serving a prior-dependent refining role.

\section{Methods}
\label{sec:glm-methods}

\subsection{The GLM-Prior model}

We developed GLM-Prior, a genomic language model fine-tuned to predict transcription factor (TF)-target gene regulatory interactions directly from sequence data. Specifically, we fine-tuned the 250 million parameter Nucleotide-Transformer model \cite{dalla2024nucleotide}, originally pre-trained on the genomes of 850 species, to infer binary interaction matrices between TFs and their target genes using their nucleotide sequences. These predicted interactions serve as a prior-knowledge for downstream GRN inference, introducing biologically grounded constraints that guide the structure of the inferred networks.

The model uses a transformer encoder architecture to process two distinct sequence types, TF motif sequences from the CisBP database \cite{weirauch2014determination} and gene body sequences derived from genome annotations in a GTF file. We append a classification head to the transformer encoder to predict whether each TF-gene pair represents a true regulatory interaction (positive label) or non-regulatory interaction (negative label). Training labels are obtained from experimentally validated interaction databases such as STRING \cite{mering2003string, szklarczyk2010string, szklarczyk2021string, szklarczyk2023string} and TRRUST \cite{han2015trrust, han2018trrust}, supplying high confidence positive and negative examples of regulatory interactions. This design allows the model to leverage both sequence-level regulatory motifs encoded in the pre-trained transformer and database-derived TF-gene interactions to improve the predictive performance and generalization across genomic contexts.

Each training example is constructed by concatenating the nucleotide sequence of a TF with that of its candidate target gene, separated by a special classification token ($\texttt{<cls>}$). This composite sequence is tokenized, during which a model-specific ($\texttt{<cls>}$) token is automatically prepended to the input. The tokenized sequence is passed through the pre-trained transformer, which produces contextual embeddings across the full input. The embedding corresponding to the prepended ($\texttt{<cls>}$) token is fed into the classification head composed of a dropout layer, a linear projection, a $\tanh$ activation, a second dropout, and a final linear layer that maps to two logits representing the binary interaction label. This setup allows the model to jointly learn representations of TF-binding specificity and gene regulatory potential in a unified sequence-to-interaction framework.

\subsubsection{Dataset construction and preprocessing}
To train the model, we constructed a dataset of TF-gene pairs where positive examples were derived from a curated database of experimentally validated interactions (YEASTRACT \cite{teixeira2018yeastract}, STRING \cite{mering2003string, szklarczyk2010string, szklarczyk2021string, szklarczyk2023string} and TRRUST \cite{han2015trrust, han2018trrust}), and negative examples were sampled from a combination of all remaining TF-gene sequence pairs. This dataset was split into a $99\%$ training and $1\%$ validation set. 

Given the substantial class imbalance between positive and negative samples, we implemented a downsampling strategy to reduce the number of negative samples in the training set. This preserved the diversity of negative samples, while preventing the model from overfitting to the negative class. Specifically, the number of retained negative samples after downsampling can be defined as,

\begin{align}
    N_{sampled} = \lfloor r \cdot N_{-} \rfloor, \label{eq:neg_sampling}
\end{align}

where $r\in (0, 1)$ is the downsampling rate, and $N_{-}$ is the total number of negative examples in the dataset. 

To further address class imbalance during training, we designed a custom DataLoader that performs even class batch sampling. Each batch was constructed to contain an equal number of positive and negative samples, ensuring a balanced signal during training. Specifically, each batch is defined as,

\begin{align}
    B = \{(x_i^+, x_j^-) : x_i^+ \in X_{+}, x_j^- \in X_{-}\}, \label{eq:even_class_data_loader}
\end{align}

where positive examples $x_i^+$ were sampled uniformly with replacement from $X_{+}$ (the positive class) and negative examples $x_j^-$ were cycled through without replacement:

\begin{align}
    x_i^+ \sim X_{+}, \quad x_j^- \sim X_{-} \quad \text{ with } \quad j \bmod  N_{-}.
\end{align}

This strategy ensures that each negative example was seen exactly once during training while positive examples were reused as needed to maintain class balance. Even-class batching was critical for stabilizing training dynamics and improving the model's sensitivity to true positive interactions without inflating the false-negative rate.

\subsubsection{Model Architecture and Classification}

The model architecture builds on the pre-trained Nucleotide Transformer by retaining its embedding and transformer encoding layers. A classification head is applied to the final hidden state corresponding to the prepended ($\texttt{<cls>}$) token to compute logits for a binary classification task. This classification head outputs a two-dimensional logit vector $z \in \mathbb{R}^2$ representing unnormalized scores for the positive and negative classes,

\begin{align}
    z = W \cdot h + b,
\end{align}

where $h$ is the transformed hidden state of the ($\texttt{<cls>}$) token following the intermediate projection and non-linearity, and $W$, $b$ are the weights and bias of the final linear layer. The components of $z$ are denoted $z_{+}$ and $z_{-}$, corresponding to the logits for the positive and negative classes, respectively. Class probabilities are computed using a softmax function,

\begin{align}
    P(y=1 | x_{TF}, x_{gene}) = \frac{\exp (z_{+})}{\exp (z_{+}) + \exp (z_{-})},
\end{align}

Training used a class-weighted cross-entropy loss function to account for residual class imbalance after downsampling. The loss assigns a fixed weight of $1.0$ to positive examples, while the negative class weight $w_{-}$ is tuned through hyperparameter search to optimize the balance between precision and recall. The loss is computed as,

\begin{align}
    \mathcal{L} &= - w_+ y \log p - w_- (1-y) \log (1-p), \quad \text{where} \quad
    p = \frac{\exp(z_+)}{\exp(z_+) + \exp(z_-)}. \label{eq:cross_entropy_loss}
\end{align}

Here, $y \in \{ 0, 1 \}$ is the true label, $z_{+}$ and $z_{-}$ are the predicted logits for the positive and negative classes, respectively. $w_{-}$ is the down-weighted negative class weight, varied during hyperparameter search, and the positive class weight $w_{+}$ is fixed at $1.0$. This approach maintains sensitivity to positive predictions while mitigating bias toward the negative class.

\subsubsection{Training procedure}
\label{yeast-GLM-Prior}

We trained the model on 4 H100 GPUs using PyTorch's Distributed Data Parallel (DDP) framework \cite{li2020pytorch} to enable efficient multi-GPU scaling. The training process was distributed across GPUs to accelerate computation and ensure consistent gradient updates. We used a per-device batch size of 32 and set the gradient accumulation steps to $32$, resulting in an effective batch size of $4096$. The model was optimized using Adam with a learning rate of $10^{-5}$. Training spanned 10 epochs, using optimal hyperparameters selected through a sweep over the negative class weight ($w_{-}$) and downsampling rate (see Appendix \ref{Appendix} for more details). We selected the configuration that achieved the highest F1 score on the validation set for final training.

After training, the model was used to infer a prior-knowledge matrix of TF-gene regulatory interactions from a list of gene-TF sequence pairs without labels. This matrix then served as input for downstream GRN inference, where GRN inference can further tailor these language model derived interactions using cell-type, cell-line, or condition-specific expression data.

\subsubsection{Performance and Evaluation}
We evaluated model performance using standard binary classification metrics, with a focus on metrics that remain robust under class imbalance. Specifically, we report precision, recall, F1 score, area under the receiver operating characteristic curve (AUC-ROC), area under the precision-recall curve (AUPRC), and Matthews correlation coefficient (MCC). 

Precision and recall were computed separately for the positive and negative classes to assess the model's ability to minimize false positives and false negatives, respectively. Let $TP$, $FP$, $FN$ and $TN$ denote the true positives, false positives, false negatives and true negatives. Then,

\begin{align}
    \mbox{Precision} = \frac{TP}{TP + FP}, \quad
    \mbox{Recall} = \frac{TP}{TP + FN}.
\end{align}

The F1 score, which represents the harmonic mean of precision and recall, was used as the primary metric for model selection and hyperparameter optimization. It can be computed as,

\begin{align}
    F_1 = 2 \cdot \frac{\mbox{Precision} \cdot \mbox{Recall}}{\mbox{Precision} + \mbox{Recall}}.
\end{align}

To account for class imbalance and provide a threshold-independent measure of performance, we also computed AUC-ROC and AUPRC. the ROC curve plots true positive rate (TPR) against false positive rate (FPR), defined as:

\begin{align}
    \mbox{True Positive Rate (TPR)} = \frac{TP}{TP+FN}, \quad
    \mbox{False Positive Rate (FPR)} = \frac{FP}{FP+TN}.
\end{align}

While AUC-ROC captures the model's general discrimintative ability, AUPRC is more informative in imbalanced settings, as it directly reflects the trade-off between precision and recall. We used AUPRC to benchmark model predictions against curated gold standard datasets of TF-gene interactions.

To determine the optimal classification threshold, we performed a grid search over the predicted positive class probabilities. The threshold $t^*$ that maximized the F1 score on the validation set was selected for final inference,

\begin{align}
    t^* = \operatorname*{arg\,max} F_1(t)
\end{align}

Finally, we report the Matthews correlation coefficient (MCC), a balanced measure of classification quality that incorporates all four confusion matrix components,

\begin{align}
    \mbox{MCC} = \frac{TP \cdot TN - FP \cdot FN}{\sqrt{(TP + FP) (TP + FN) (TN+FP)(TN+FN)}}.
\end{align}

MCC ranges from $-1$ to $+1$, where $+1$ indicates perfect predictions, $0$ indicates random predictions, and $-1$ indicates incorrect predictions. MCC remains informative even when classes are highly imbalanced, making it a useful complement to F1 and AUPRC.

\subsection{GRN inference with PMF-GRN}

We performed GRN inference using our previously published method, PMF-GRN (Probabilistic Matrix Factorization for Gene Regulatory Network inference) \cite{skok2024pmf} to infer the regulatory interactions between TFs and their target genes. The goal of PMF-GRN is to decompose an observed gene expression matrix into latent factors that represent TF activity and regulatory interactions between TFs and their target genes. These latent factors capture the underlying GRN structure, which cannot be measured directly from gene expression data alone. Further details regarding the PMF-GRN model and the inference strategy used to obtain GRNs can be found in \cite{skok2022high}.

Using PMF-GRN, we perform inference independently on each single-cell dataset to obtain dataset-specific GRNs. These inferred networks are then combined post-inference using a simple averaging strategy to produce a consensus GRN, 

\begin{align}
    \mbox{GRN}_{\text{Consensus}} = \frac{1}{N} \sum_{i=1}^N \mbox{GRN}_i,
\end{align}

where $N$ is the number of datasets and $\mbox{GRN}_i$ is the inferred network for dataset $i$. This consensus GRN captures regulatory interactions that are consistently inferred across datasets, while preserving dataset-specific networks for context-specific analyses.

To evaluate the relationship between posterior uncertainty and predictive accuracy, we rank all TF-gene interaction pairs by their posterior variance, constructing 10 cumulative bins corresponding to the variances in increments of $10\%$. For each bin $k$, we use the posterior point estimates of the interactions within the bin and compute the AUPRC using a gold standard set of validated TF-gene interactions $\mathcal{G}$ as the reference. All interactions in each bin are included in the evaluation, regardless of whether they appear in the gold standard.

Let $\mathcal{B}_k$ be the set of posterior point estimates in the $k\%$ of variances. Then the AUPRC for bin $k$ is given by:

\begin{align}
    \mbox{AUPRC}_k = \mbox{AUPRC}(\mathcal{B}_k; \mathcal{G}),
\end{align}

where the AUPRC is computed by comparing the predicted scores in $\mathcal{B}_k$ to labels derived from $\mathcal{G}$. The cumulative binning strategy ensures that each bin contains a sufficient number of interactions for stable AUPRC estimation.

\section*{Discussion}
\label{sec:discussion}

Accurately reconstructing GRNs remains a central challenge in genomics, particularly in complex systems where direct experimental measurements are incomplete or unavailable. In this work, we present GLM-Prior, a genomic language model fine-tuned to predict TF-gene regulatory interactions directly from DNA sequence, and integrate it with PMF-GRN, a probabilistic matrix factorization framework for expression-based GRN inference. This dual-stage design decouples prior construction from downstream GRN inference, allowing us to evaluate how different priors and GRN inference algorithms jointly shape GRN reconstruction across yeast, mouse, and human cell lines. 

Our results first demonstrate that GLM-Prior's performance is tightly linked to the composition of the training data. Across six cell lines spanning yeast, human, and mouse, predictive accuracy scales with the abundance and diversity of positive labels and TF coverage. Human and mouse cell lines with rich, well-annotated regulatory datasets support substantially higher AUPRCs than yeast, where extreme class imbalance and sparse labels limit generalization despite good within-training metrics. These trends highlight that even with powerful sequence models, generalization is fundamentally constrained by the available supervision. Here, we find that when only a small subset of true interactions are labeled, the model can capture some regulatory grammar but cannot fully resolves novel gene-TF combinations in held-out contexts.

By systematically varying training paradigms, we further show that GLM-Prior generalizes in a biologically interpretable manner across species. Transfer learning between evolutionarily related species, such as human and mouse, maintains or modestly improves performance relative to single-species models, indicating that GLM-Prior captures conserved sequence-level features that can be reused across lineages. In contrast, models trained in mammals transfer poorly to yeast, where performance remains near chance, consistent with deeper divergence in regulatory grammar, TF binding preferences, and genome organization. Multi-species training on human, mouse, and yeast preserves accuracy comparable to the best single-species or transfer models in most cell lines, with only mild trade-offs in some human contexts, and yields a domain-stable prior that can be applied to new systems without species-specific retraining. Together, these findings establish GLM-Prior as a flexible framework for constructing priors across diverse data regimes, with strong performance in well-annotated human and mouse cell lines and more limited gains in sparsely labeled settings such as yeast.

Furthermore, benchmarking GLM-Prior against widely used accessibility-based priors, including Inferelator-Prior and CellOracle's baseGRN, across six cell lines demonstrates that GLM-Prior consistently performs at or above chance and provides the strongest prior in four of six contexts (hESC, mESC, mDC, and mHSC). At the same time, accessibility-based priors remain competitive and even superior in settings dominated by proximal regulatory logic, such as yeast, where promoter-focused chromatin accessibility and motif enrichment capture a large fraction of regulatory edges. In more complex mammalian systems, accessibility-based priors often fall below chance while GLM-Prior remains predictive, suggesting that sequence-based models may better capture distal enhancer regulation and long-range TF-gene coupling that are not easily recovered from proximity-based peak-gene assignment. These complementary performance profiles indicate that GLM-Prior provides the most robust and transferable prior knowledge across diverse mammalian contexts, while accessibility-based priors remain particularly effective in promoter-dominated scenarios such as yeast, together capturing distinct and complementary facets of regulatory architecture.

Integrating these priors with multiple GRN inference algorithms reveals a consistent pattern: the quality and structure of the prior largely determines the achievable performance, and GRN inference algorithms mostly modulate performance within a range set by the prior. First, when we use GLM-Prior as input to PMF-GRN, AUPRC is equal to or higher than the prior alone in all six cell lines, with modest gains when the prior is already strong and larger improvements when the prior is weaker. Edge-flip analyses in this setting show that PMF-GRN typically makes relatively small, targeted changes, treating the sequence-derived prior as a stable scaffold and adding or pruning a limited subset of edges rather than reconstructing the network from scratch. We then extend this analysis to the full cross-method setting, in which all three priors (GLM-Prior, Inferelator-Prior, and CellOracle’s baseGRN) are combined with all three GRN inference methods (PMF-GRN, Inferelator, and CellOracle). In this fully crossed design, the best-performing GRNs in each cell line almost always arise from combinations that use the strongest prior for that context, regardless of which inference algorithm is applied. At the same time, no GRN inference method is uniformly superior across priors or species. Each GRN algorithm performs best in different prior–species regimes and interacts with priors in characteristic ways, from highly prior-sensitive behavior to more robust, expression-driven refinement. Together, these analyses indicate that GRN performance is primarily limited by prior construction, with inference algorithms providing prior-dependent gains rather than defining the overall performance ceiling.

Our results recast GRN inference as a problem in which the central bottleneck is the construction of high-quality prior knowledge, rather than the specific choice of the downstream expression-based inference algorithm. GLM-Prior addresses this bottleneck by leveraging a transformer-based genomic language model to learn regulatory grammar directly from DNA, enabling priors that generalize across cell types, species, and training paradigms. Critically, because GLM-Prior operates directly on genome sequence, it removes the dependence on cell-type- and assay-specific measurements such as ATAC-seq, and thus makes it possible, in principle, to construct regulatory priors in understudied or experimentally inaccessible species where chromatin accessibility or expression data are sparse or unavailable. While accessibility-based priors remain valuable in contexts dominated by relatively simple, proximal promoter–gene regulation, they struggle to capture regulatory architecture in more complex mammalian cell lines and cannot be readily transferred to unseen species or cell types, in contrast to sequence-based priors such as GLM-Prior.

Several limitations of our study point to promising directions for future work. First, GRN benchmarks are constrained by incomplete and noisy interaction databases, in which many negative labels likely represent unknown or untested edges rather than true absences of regulation. This incompleteness affects both prior construction and evaluation, and motivates the development of benchmarks that better distinguish between confirmed negatives and unlabeled interactions. Further, as GLM-Prior downsamples negative examples from the input training data, we likely discard false negative edges during training and thus never learn these regulatory interactions using our sequence-based approach. Future work could include more sampling techniques to avoid losing access to interactions previously defined as negative example that could be in fact classified as positive examples during training. Second, our models are trained on a limited set of cell lines and regulatory labels. Extending GLM-Prior to broader compendia of cell types, developmental stages, and species will be important for assessing how far sequence-based priors can be pushed as a general regulatory scaffold that other methods can refine. Third, while we consider multiple GRN inference algorithms, they share a common edge-centric view of network reconstruction. Exploring integration strategies that move beyond edge-wise scoring, such as jointly modeling regulatory modules, trajectories, or causally inferred dynamics \cite{kim2025large}, may reveal additional ways that expression data can complement strong sequence-derived priors.

In summary, this work introduces GLM-Prior as a scalable, sequence-based framework for constructing regulatory priors and systematically characterizes how these priors interact with GRN inference methods across yeast, mouse, and human. By demonstrating that prior quality is the dominant determinant of GRN reconstruction performance, our study shifts the emphasis in GRN modeling toward learning robust, generalizable priors from sequence, and provides a foundation for future methods that integrate these priors with expression and other modalities in more flexible and biologically grounded ways.

\section{Supplementary Material for Chapter \ref{sec:GLM-Prior-Chapter}}
\label{Appendix}

\subsection{Single-Species Experiments}
\label{appendix:single-species}

\subsubsection{Yeast}
\label{appendix:single-yeast}
To train our GLM-Prior model in yeast, we first obtained all $5,999$ gene body nucleotide sequences from the ENSEMBL \textit{S. cerevisiae} (R64-1-1.UTR.gtf) genome. We obtained $212$ TF sequences from the CisBP database \cite{weirauch2014determination} under \textit{S. cerevisiae}. 

Next, to pair these gene and TF nucleotide sequences with interaction labels, we used the YEASTRACT database of interactions \cite{teixeira2018yeastract} ($6,885$ genes by $220$ TFs). From these $220$ YEASTRACT TFs, $46$ did not have sequences associated with them from CisBP. Due to this large portion of data loss ($20\%$), we used the promoter regions of the target genes for each missing TF as a proxy for it's binding sequence. We defined the promoter sequence following YEASTRACT's definition of $1000$bp upstream or downstream of the gene TSS (depending on the strand orientation).  Training for yeast was conducted across $5,893$ genes and $123$ TFs, with $660$ positive labels and $724,179$ negative labels for these interactions derived from YEASTRACT.

\begin{table}[!htbp]
    \centering
    \renewcommand{\arraystretch}{1.2}  
    \begin{tabular}{l|c}
        \multicolumn{2}{c}{\textbf{Validation Metrics for Yeast Single-Species GLM-Prior}} \\ \hline
        Metric & Score \\ \hline
        Best F1 Score & $0.76$ \\
        ROC AUC & $1.00$ \\
        Best Classification Threshold & $1.00$ \\
        Positive Class Precision & $0.89$ \\
        Positive Class Recall & $0.67$ \\
        Negative Class Precision & $1.00$ \\
        Negative Class Recall & $1.00$ \\
        AUPRC (vs. held-out set \cite{tchourine2018condition})& $0.02$ \\
    \end{tabular}
    \caption{Validation performance of the GLM-Prior model trained on yeast. Evaluation was conducted on a held-out validation set that assess positive and negative class contributions during training (top) and AUPRC against an independent gold standard for the final inferred prior-knowledge matrix (bottom).}
    \label{table:yeast_GLM_validation_metrics}
\end{table}

A hyperparameter sweep over $1$ epoch of training using different class-weights and downsampling rates for the negative class revealed $[0.5, 1.0]$ to be the optimal class-weights and $0.4$ to be the optimal negative class downsampling rate. These hyperparameters were used during final training over $10$ epochs, obtaining the validation metrics on the held-out $1\%$ of training data found in Table \ref{table:yeast_GLM_validation_metrics}. 

Following training of the yeast single-species GLM-Prior model on YEASTRACT database labels, we ran inference on all genes and TFs, including those in a held-out set not seen during training. These held-out genes and TFs corresponded to those present in the gold standard from \cite{tchourine2018condition}. Our held-out set comprised $933$ genes and $98$ TFs, with $986$ positive labels, and $95,911$ negative labels. After inference, the model predictions were evaluated using AUPRC with the corresponding labels from the gold standard, with a chance baseline calculated by taking the positive rate of the held-out set.

\subsubsection{Sequence homology experiments in yeast}

Sequence homology refers to shared sequence similarity between loci, often introduced by duplication and other genome rearrangements. Because nucleotide language models are optimized to exploit recurring sequence patterns, high homology across train-test splits can, in principle, make held-out prediction easier by allowing the model to match familiar fragments rather than rely on fully generalizable features. For GLM-Prior, which is trained on labeled TF-sequence pairs, this motivates a simple check: whether genes or TFs are unusually similar to sequences seen during training, and whether controlling for that similarity changes performance at inference time. We therefore use hashFrag \cite{rafi2025detecting}, which leverages BLAST \cite{madden2013blast} to quantify cross-split sequence similarity and generate homology-aware partitions (or prune highly similar training sequences), and then re-evaluate GLM-Prior on these homology-controlled splits to test whether sequence similarity contributes to our yeast performance.

\begin{figure}[!htbp]
    \centering
    \includegraphics[width=1.0\textwidth]{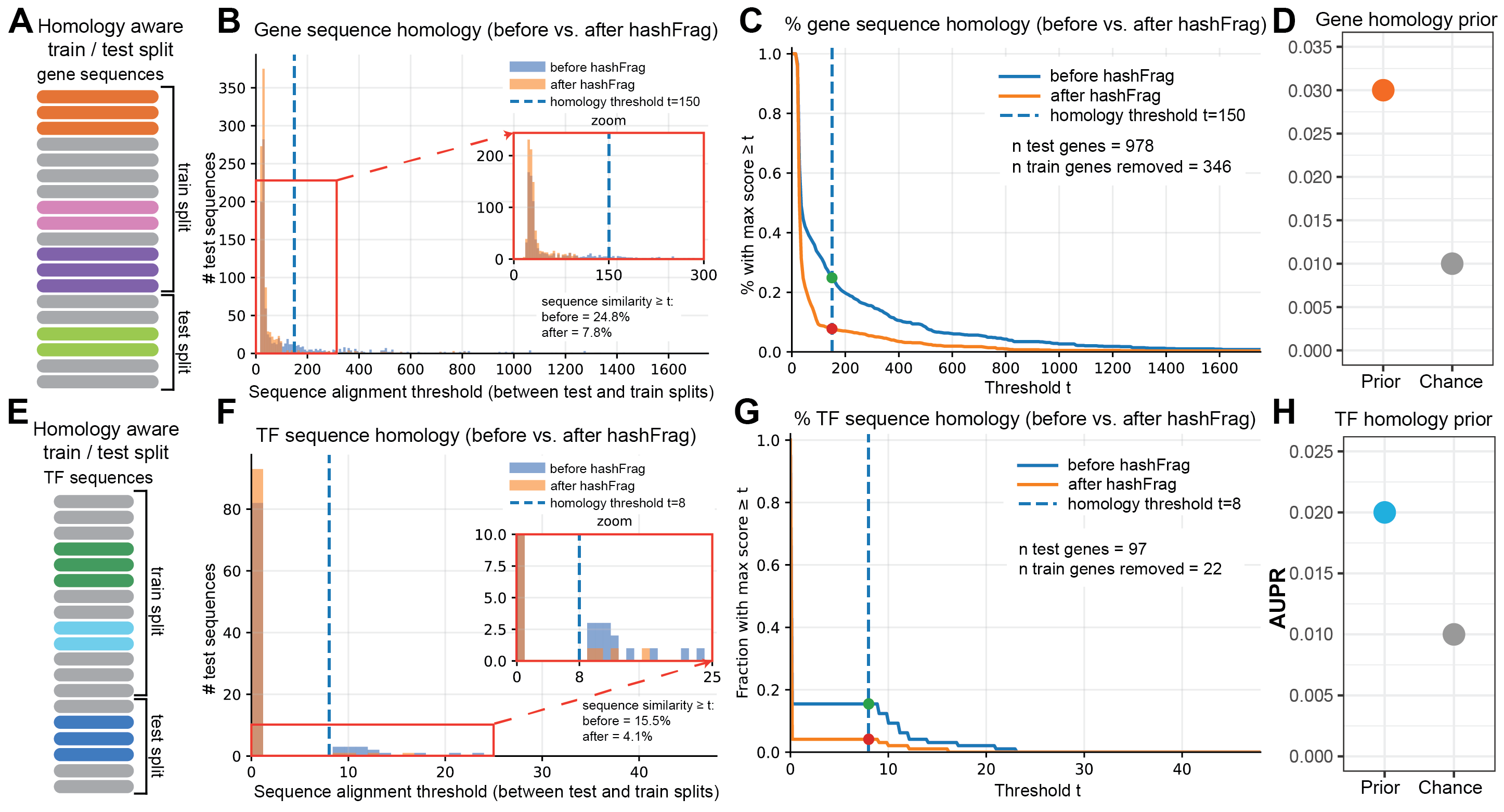} 
    \caption{\textbf{Sequence-homology analysis of yeast gene and TF splits using hashFrag.}
    \textbf{(A)} Schematic of homology-aware partitioning, illustrating how highly similar sequences are identified and separated to reduce cross-split homology between training and test sets.
    \textbf{(B)} Histogram of per-test-gene maximum BLAST alignment score to any training gene, shown before (blue) and after (orange) hashFrag pruning of the training gene set. Blue dashed line indicates the similarity threshold used to define leakage.
    \textbf{(C)} Leakage curve for genes showing, for each similarity threshold $t$, the fraction of test genes whose maximum BLAST score to any training gene is $\geq t$. Blue dashed line marks the leakage threshold.
    \textbf{(D)} AUPRC evaluation of GLM-Prior trained and tested on the hashFrag-defined gene partitions, with the gray dot indicating chance performance.
    \textbf{(E)} Schematic of homology-aware partitioning for TF sequences, analogous to panel A.
    \textbf{(F)} Histogram of per-test-TF maximum BLAST alignment score to any training TF, shown before (blue) and after (orange) hashFrag pruning of the training TF set. Blue dashed line indicates the similarity threshold.
    \textbf{(G)} Leakage curve for TFs showing, for each similarity threshold $t$, the fraction of test TFs whose maximum BLAST score to any training TF is $\geq t$. Blue dashed line marks the leakage threshold.
    \textbf{(H)} AUPRC evaluation of GLM-Prior trained and tested on the hashFrag-defined TF partitions, with the gray dot indicating chance performance.}
    \label{fig:sequence-homology}
\end{figure}

In Figure \ref{fig:sequence-homology}A we depict the homology-aware partitioning scheme. We then quantify leakage among genes by computing, for each test gene, its maximum BLAST score to any training gene (Figure \ref{fig:sequence-homology}B-C). Using an operational cutoff $t=150$, and holding the test set fixed ($n=978$), pruning $346$ homologous training genes reduces the fraction of "leaking" test genes from $24.8\%$ to $7.8\%$.

We next ask whether removing this homology affects model accuracy. Training on the hashFrag-defined gene partitions yields a test AUPRC of $0.03$, compared to $0.01$ chance performance (Figure \ref{fig:sequence-homology}D). This magnitude is comparable to the yeast results (AUPRC = $0.02$) reported in Section \ref{sec:GLM-Prior-Model-Results}, indicating that sequence leakage is not responsible for our observed performance in yeast.

We repeat the same analysis for TF sequences (Figure \ref{fig:sequence-homology}E-H). As with genes, hashFrag pruning decreases cross-split similarity, and model performance on the homology-clean TF splits remains similar to the main yeast model, again motivating against leakage-driven gains. 

Overall, these experiments show that while homology-aware partitioning is methodologically important and successfully reduces cross-split similarity, it does not materially affect our model performance, which is primarily dependent on access to sufficient positive training examples. Because aggressive pruning removes substantial numbers of sequences, and thus labels, from both training and evaluation, it can reduce statistical power without measurable benefit here. For this reason, we include this analysis in the appendix to demonstrate its value, and motivate why we avoid homology pruning in the primary experiments to preserve sufficient data for training and evaluation.

\subsubsection{Human cell lines (hESC \& HepG2)}
\label{appendix:single-human}
To train our GLM-Prior model in human, we first obtained hg38 gene body nucleotide sequences from ENSEMBL (RCh38.113.gtf). Due to the lengthy nature of human genes, and the inherent limitations of context length in large language models, we filtered our gene sequences to retain all sequences for a gene body $\leq 12,000$ nucleotides in length. This provided us with a list of $23,533$ genes. We obtained TF binding motif sequences from the CisBP database \cite{weirauch2014determination}, under \textit{H. sapiens}. 

\begin{table}[!htbp]
    \centering
    \renewcommand{\arraystretch}{1.2}  
    \begin{tabular}{l|@{\hspace{.5cm}}c@{\hspace{.75cm}}|@{\hspace{.25cm}}c}
        \multicolumn{3}{c}{\textbf{Validation Metrics for Human Single-Species GLM-Prior}} \\ \hline
        \textbf{Metric} & \textbf{hESC} & \textbf{HepG2} \\ \hline
        Best F1 Score & $0.29$ & $0.35$ \\
        ROC AUC & $0.85$ & $0.83$ \\
        Best Classification Threshold & $0.94$ & $0.97$ \\
        Positive Class Precision & $0.39$ & $0.53$ \\
        Positive Class Recall & $0.22$ & $0.27$ \\
        Negative Class Precision & $0.99$ & $0.99$ \\
        Negative Class Recall & $1.00$ & $1.00$ \\
        AUPRC (vs. reference network)& $0.19$ & $0.30$ \\
    \end{tabular}
    \caption{Validation performance of the GLM-Prior model trained on human, evaluated in hESC and HepG2. Evaluation is conducted on held-out sets using validation metrics that consider the contribution of the positive and negative classes and AUPRC against ChIP-seq-derived reference networks after inference of the prior-knowledge matrix.}
    \label{table:human_GLM_validation_metrics}
\end{table}

Training for hESC was conducted across $1,801$ genes and $504$ TFs, with $5,305$ positive labels and $902,399$ negative labels for these interactions derived from STRING and TRRUST. Training for hESC was conducted across $2,121$ genes and $527$ TFs, with $8,087$ positive labels and $1,109,680$ negative labels for these interactions derived from STRING and TRRUST.

A hyperparameter sweep over $1$ epoch of training using different class-weights and downsampling rates for the negative class revealed $[0.8, 1.0]$ to be the optimal class-weights and $0.3$ to be the optimal negative class downsampling rate in hESC, and $[1.0, 1.0]$ to be the optimal class-weights and $0.4$ to be the optimal negative class downsampling rate in HepG2. These weights were used during final training and achieved the validation metrics found in Table \ref{table:human_GLM_validation_metrics}. 

Following training of the human single-species GLM-Prior model on STRING and TRRUST database labels, we ran inference on all genes and TFs, including those in a held-out set not seen during training. These held-out genes and TFs corresponded to those present in the BEELINE hESC and HepG2 reference ChIP-seq networks, respectively. For hESC, our held-out set comprised $4,773$ genes and $79$ TFs, with $58,337$ positive labels, and $318,730$ negative labels. For HepG2, our held-out set comprised $4,338$ genes and $53$ TFs, with $56,567$ positive labels, and $173,347$ negative labels. After inference, the model predictions were evaluated using AUPRC with the corresponding labels from the hESC and HepG2 reference ChIP-seq networks, with a chance baseline calculated by taking the positive rate of the held-out set.

\subsubsection{Mouse cell lines (mESC, mDC, \& mHSC)}
\label{appendix:single-mouse}
To train our GLM-Prior model in mouse, we first obtained the mm10 gene body nucleotide sequences from ENSEMBL (GRCm39.113.gtf). Similarly to the single-species human experiments, we again filtered the length of our mouse genes to retain all sequences for a gene body $\leq 12,000$ nucleotides in length. This provided us with a list of $37,755$ genes. We obtained TF binding motif sequences from the CisBP database \cite{weirauch2014determination}, under \textit{M. musculus}. 

\begin{table}[!htbp]
    \centering
    \renewcommand{\arraystretch}{1.2}  
    \begin{tabular}{l|c@{\hspace{0.25cm}}|@{\hspace{0.25cm}}c@{\hspace{0.25cm}}|@{\hspace{0.25cm}}c}
        \multicolumn{4}{c}{\textbf{Validation Metrics for Mouse Single-Species GLM-Prior}} \\ \hline
        \textbf{Metric} & \textbf{mESC} & \textbf{mDC} & \textbf{mHSC} \\ \hline
        Best F1 Score & $0.19$ & $0.20$ & $0.19$ \\
        ROC AUC & $0.70$ & $0.80$ & $0.74$ \\
        Best Classification Threshold & $0.97$ & $0.99$ & $0.98$ \\
        Positive Class Precision & $0.23$ & $0.17$ & $0.50$ \\
        Positive Class Recall & $0.16$ & $0.19$ & $0.12$ \\
        Negative Class Precision & $0.99$ & $0.99$ & $0.99$ \\
        Negative Class Recall & $0.99$ & $0.99$ & $1.00$ \\
        AUPRC (vs. reference network) & $0.23$ & $0.09$ & $0.49$ \\
    \end{tabular}
    \caption{Validation performance of the GLM-Prior model trained on mouse, evaluated in mESC, mDC, and mHSC. Evaluation is conducted on held-out sets using validation metrics that consider the contribution of the positive and negative classes and AUPRC against ChIP-seq-derived reference networks after inference of the prior-knowledge matrix.}
    \label{table:mouse_GLM_validation_metrics}
\end{table}

Training for mESC was conducted across $1,097$ genes and $437$ TFs, with $3,129$ positive labels and $476,260$ negative labels for these interactions derived from STRING and TRRUST. Training for mDC was conducted across $2,837$ genes and $462$ TFs, with $11,551$ positive labels and $1,299,134$ negative labels for these interactions derived from STRING and TRRUST. Training for mHSC was conducted across $1,129$ genes and $419$ TFs, with $2,808$ positive labels and $470,243$ negative labels.

A hyperparameter sweep over $1$ epoch of training using different class-weights and downsampling rates for the negative class revealed $[0.2, 1.0]$ to be the optimal class-weights and $0.5$ to be the optimal negative class downsampling rate in mESC, and $[0.1, 1.0]$ to be the optimal class-weights and $0.4$ to be the optimal negative class downsampling rate in mDC, and $[0.1, 1.0]$ to be the optimal class-weights and $0.5$ to be the optimal negative class downsampling rate in mHSC. These weights were used during final training, achieving the validation metrics shown in Table \ref{table:mouse_GLM_validation_metrics}. 

Following training of the mouse single-species GLM-Prior model on STRING and TRRUST database labels, we ran inference on all genes and TFs, including those in a held-out set not seen during training. These held-out genes and TFs corresponded to those present in the BEELINE mESC, mDC and mHSC reference ChIP-seq networks, respectively. For mESC, our held-out set comprised $5,711$ genes and $59$ TFs, with $65,933$ positive labels, and $271,016$ negative labels. For mDC, our held-out set comprised $3,373$ genes and $29$ TFs, with $8,708$ positive labels, and $89,109$ negative labels. For mHSC, our held-out set comprised $6,777$ genes and $72$ TFs, with $182,005$ positive labels, and $305,939$ negative labels. After inference, the model predictions were evaluated using AUPRC with the corresponding labels from the mESC, mDC and mHSC reference ChIP-seq networks, with a chance baseline calculated by taking the positive rate of the held-out set.

\subsection{Transfer-Learning Experiments}
Transfer learning experiments were conducted by first ensuring that no overlap existed between the training set (human, mouse, or yeast), and the corresponding inference set with held-out evaluation labels. For the human and mouse model that transferring knowledge to yeast, no information was overlapping between training and inference \textit{a priori}. 

\begin{table}[!htbp]
    \centering
    \renewcommand{\arraystretch}{1.2}  
    \begin{tabular}{l|l|c}
        \multicolumn{3}{c}{\textbf{AUPRC Results from Transfer Learning}} \\ \hline
        Training Dataset & Inference Dataset & AUPRC \\ \hline
        Human and Mouse  & Yeast             & $0.02$ \\
        Mouse            & hESC              & $0.19$ \\
        Mouse            & HepG2             & $0.31$ \\
        Human            & mESC              & $0.23$ \\
        Human            & mDC               & $0.09$ \\
        Human            & mHSC              & $0.46$ \\
    \end{tabular}
    \caption{AUPRC scores for cross-species transfer learning. Each row indicates the species used to train the GLM-Prior model and the target species on which GRN inference was performed. Evaluation was conducted against species-specific ChIP-seq or curated reference datasets.}
    \label{table:transfer-learning-inference-auprc}
\end{table}

For the mouse trained model which transferred knowledge to hESC and HepG2, two separate models were trained, one in which overlaps between the mouse genes, TFs and labels with hESC were removed, and the second in which the overlaps between mouse genes, TFs and labels with HepG2 were removed. For the human trained model which transferred knowledge to mESC, mDC, and mHSC, three separate models were trained. In the first model, overlaps between human genes, TFs, and labels with mESC were removed. In the second model, overlaps between human genes, TFs, and labels with mDC were removed. In the third model, overlaps between human genes, TFs, and labels with mHSC were removed. 

Training six separate models allowed us to ensure an optimal number of gene and TF sequences and their corresponding labels were seen during training, while maintaining complete independence from the inference set and corresponding evaluation performance. Results for the transfer learning experiments can be found in Table \ref{table:transfer-learning-inference-auprc}.

\subsection{Multi-Species Experiments}
\label{appendix:multi-species}
To train a multi-species model, we consecutively trained GLM-Prior on human, mouse, and yeast. To ensure no sequence overlap between the genes, TFs, and labels used during training and inference, we removed all overlapping inference sets from each individual organisms training set. Here, we ensured there was no inter-species leakage, as well as intra-species leakage, between training and inference sets.

Following the consecutive training of the multi-species model, we ran inference in yeast, hESC, HepG2, mESC, mDC, and mHSC held-out sets, respectively. Results for this multi-species model inference can be found in Table \ref{table:multi-species-inference-auprc}.

\begin{table}[!htbp]
    \centering
    \renewcommand{\arraystretch}{1.2}  
    \begin{tabular}{l|l|c}
        \multicolumn{3}{c}{\textbf{AUPRC Results from Multi-Species Training}} \\ \hline
        \textbf{Training Dataset} & \textbf{Inference Dataset} & \textbf{AUPRC} \\ \hline
        \multirow{6}{*}{Human + Mouse + Yeast} 
            & Yeast & $0.02$ \\
            & hESC  & $0.18$ \\
            & HepG2 & $0.29$ \\
            & mESC  & $0.23$ \\
            & mDC   & $0.09$ \\
            & mHSC  & $0.49$ \\
    \end{tabular}
    \caption{AUPRC scores for multi-species model inference in human, mouse, and yeast. GRNs were inferred using a unified model trained jointly on all three species, and evaluated against cell-line specific reference networks.}
    \label{table:multi-species-inference-auprc}
\end{table}

\subsection{Inferelator-Prior and CellOracle baseGRN}

We generated prior knowledge for Inferelator-Prior \cite{skok2022high} and CellOracle's baseGRN \cite{kamimoto2023dissecting}, using each respective methods published Python software. To generate prior knowledge with Inferelator-Prior and CellOracle baseGRN for each of the six cell lines, we obtained ATAC-seq datasets from the following accessions: Human embryonic stem cells: 4DNFIPGM38K4, Human HepG2 cells: ENCFF913MQB, Mouse embryonic stem cells: 4DNFIAEQI3RP, Mouse dendritic cells: ENCFF109TUH, and Mouse hematopoietic stem cells: ENCFF931CIR. In yeast, prior knowledge datasets were obtained from \cite{skok2022high} without further modification. These priors were constructed using motifs from CisBP, the same motifs used in GLM-Prior to ensure compatibility and fairness during evaluation. 

\begin{table}[!htbp]
    \centering
    \renewcommand{\arraystretch}{1.2}  
    \begin{tabular}{l|ccc}
        \multicolumn{1}{c}{} & \multicolumn{3}{c}{\textbf{AUPRC}} \\ \cline{2-4}
        \textbf{Cell Line} & \textbf{GLM-Prior} & \textbf{Inferelator-Prior} & \textbf{CellOracle baseGRN} \\ \hline
        Yeast & $0.02$ & $0.03$ & $0.05$ \\
        hESC  & $0.19$ & $0.17$ & $0.16$ \\
        HepG2 & $0.30$ & $0.24$ & $0.36$ \\
        mESC  & $0.23$ & $0.17$ & $0.17$ \\
        mDC   & $0.09$ & $0.07$ & $0.07$ \\
        mHSC  & $0.49$ & $0.34$ & $0.34$ \\
    \end{tabular}
    \caption{AUPRC of GLM-Prior, Inferelator-Prior and CellOracle baseGRN across six cell lines.}
    \label{table:prior_auprc_comparison}
\end{table}

Results for the Inferelator-Prior and CellOracle baseGRN prior knowledge for each six cell lines using the same reference sets described in Section \ref{appendix:single-species} are provided in Table \ref{table:prior_auprc_comparison}.

\subsection{GRN Inference with PMF-GRN, the Inferelator, and CellOracle}

GRN inference for each of the six cell lines was performed using PMF-GRN, the Inferelator 3.0, and CellOracle Python software, respectively. 

Single cell gene expression datasets were obtained from GSE125162 ($38,225$ cells by $6,763$ genes) \cite{jackson2020gene} and GSE144820 ($6,118$ cells by $6,763$ genes) \cite{jariani2020new} without modification. The dataset was then combined by concatenating GSE125162 and GSE144820 on the cells axis ($44,343$ cells by $6,763$ genes). The remaining hESC ($758$ cells by $17,735$ genes), HepG2 ($425$ cells by $11,515$ genes), mESC ($421$ cells by $18,385$ genes), mDC ($383$ cells by $7,371$ genes), and mHSC ($2,807$ cells by $4,762$ genes) single-cell expression datasets were obtained from BEELINE \cite{pratapa2020benchmarking}.

\begin{table}[!htbp]
    \centering
    \renewcommand{\arraystretch}{1.2}  
    \begin{tabular}{l|ccc}
        \multicolumn{1}{c}{} & \multicolumn{3}{c}{\textbf{AUPRC}} \\ \cline{2-4}
        \textbf{Cell Line} & \textbf{PMF-GRN} & \textbf{Inferelator 3.0} & \textbf{CellOracle} \\ \hline
        Yeast & $0.02$ & $0.04$ & $0.06$ \\
        hESC  & $0.19$ & $0.20$ & $0.17$ \\
        HepG2 & $0.38$ & $0.36$ & $0.38$ \\
        mESC  & $0.27$ & $0.25$ & $0.37$ \\
        mDC   & $0.10$ & $0.09$ & $0.09$ \\
        mHSC  & $0.52$ & $0.42$ & $0.39$ \\
    \end{tabular}
    \caption{AUPRC of three GRN inference methods (PMF-GRN, Inferelator 3.0, and CellOracle) across six cell lines.}
    \label{table:grn_method_auprc_comparison}
\end{table}

Prior knowledge for each method is required to produce a GRN from single-cell expression data. For the experiments in this section, we use the specific cell line prior knowledge generated by the corresponding algorithm (described in Section \ref{appendix:single-species} for GLM-Prior and Section \ref{table:prior_auprc_comparison} for Inferelator-Prior and CellOracle baseGRN. Results for these GRN inference experiments can be found in Table \ref{table:grn_method_auprc_comparison}).

\begin{figure}[!htbp]
    \centering
    \includegraphics[width=0.9\textwidth]{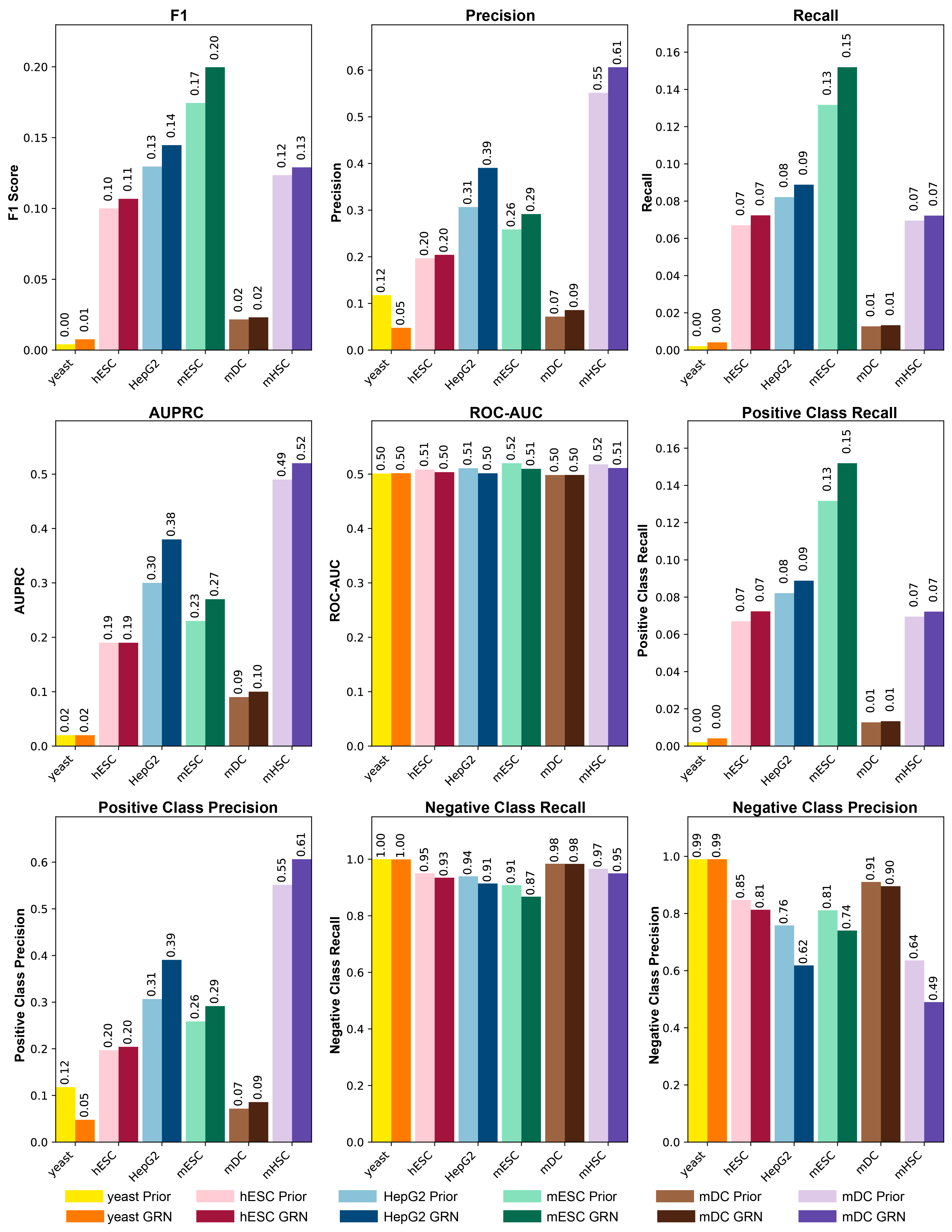} 
    \caption{\textbf{Performance Metrics}. Additional performance metrics for GLM-Prior and PMF-GRN across six cell lines.}
    \label{fig:supplemental_metrics}
\end{figure}

Additional metrics for GLM-Prior and PMF-GRN, including F1 score, precision, recall, ROC-AUC, positive class recall, positive class precision, negative class recall, and negative class precision, are provided in \ref{fig:supplemental_metrics}.

\subsection{Cross-Method Comparison}

In order to disentangle the performance contributions provided by each prior knowledge method (GLM-Prior, Inferelator-Prior, and CellOracle baseGRN) from their corresponding GRN inference method (PMF-GRN, Inferelator 3.0, and CellOracle), we provide a cross-method comparison in which each GRN inference algorithm is applied to each prior. No new datasets are introduced in this section. The results from the cross-method comparison can be found in Table \ref{table:cross_method_prior_grn_auprc}.

\begin{table}[!htbp]
    \centering
    \renewcommand{\arraystretch}{1.2}
    \begin{tabular}{l|ccc|ccc|ccc}
        \multicolumn{1}{c}{} 
            & \multicolumn{3}{c|}{\textbf{CellOracle Prior}} 
            & \multicolumn{3}{c|}{\textbf{Inferelator-Prior}} 
            & \multicolumn{3}{c}{\textbf{GLM-Prior}} \\ \cline{2-10}
        \textbf{Cell Line} 
            & \textbf{PMF} & \textbf{Inf} & \textbf{CO}
            & \textbf{PMF} & \textbf{Inf} & \textbf{CO}
            & \textbf{PMF} & \textbf{Inf} & \textbf{CO} \\ \hline
        Yeast 
            & $0.06$ & $0.06$ & $0.06$
            & $0.03$ & $0.04$ & $0.03$
            & $0.02$ & $0.03$ & $0.03$ \\
        hESC  
            & $0.17$ & $0.18$ & $0.17$
            & $0.19$ & $0.20$ & $0.19$
            & $0.19$ & $0.17$ & $0.19$ \\
        HepG2 
            & $0.40$ & $0.34$ & $0.38$
            & $0.36$ & $0.36$ & $0.33$
            & $0.38$ & $0.34$ & $0.31$ \\
        mESC  
            & $0.17$ & $0.25$ & $0.37$
            & $0.25$ & $0.25$ & $0.23$
            & $0.27$ & $0.27$ & $0.24$ \\
        mDC   
            & $0.07$ & $0.11$ & $0.09$
            & $0.10$ & $0.09$ & $0.08$
            & $0.10$ & $0.10$ & $0.07$ \\
        mHSC  
            & $0.34$ & $0.42$ & $0.39$
            & $0.41$ & $0.42$ & $0.35$
            & $0.52$ & $0.42$ & $0.36$ \\
    \end{tabular}
    \caption{Cross-method AUPRC comparison for six cell lines, combining three prior-knowledge constructions (CellOracle prior, Inferelator-Prior, GLM-Prior) with three GRN inference methods (PMF-GRN, Inferelator, CellOracle).}
    \label{table:cross_method_prior_grn_auprc}
\end{table}

\chapter{Conclusion}
\label{chp-conclusion}
\label{sec:conclusion}

GRNs provide a mechanistic view of how TFs coordinate gene expression programs to sustain stable cell identities and mediate dynamic responses to signals and perturbations. Reconstructing these networks from genome-wide data is a central goal in modern genomics, yet remains challenging under realistic data and methodological constraints. In practice, several methodological limitations are especially consequential. GRN inference methods often tightly couple a chosen generative model to a specific inference procedure. As a result, methods developed for specific experimental modalities often require substantial redesign as new measurement technologies and modeling assumptions emerge. In addition, model selection is often treated heuristically, with many approaches committing to a single algorithmic form or fixed set of hyperparameters without systematic comparison to plausible alternatives, despite the fact that these choices can yield qualitatively different networks from the same dataset. 

Evaluation presents an additional obstacle that arises from three related limitations. First, GRN accuracy is typically assessed by comparing predicted edges to a reference network that is treated as a proxy for ground truth. Second, these reference networks are often incomplete, context dependent, or unavailable. Here, the absence of an edge in the reference frequently indicates missing evidence or limited coverage, rather than evidence that the interaction does not occur. Third, inferred networks are frequently reported only as point estimates, without an explicit measure of confidence to support interpretation when a predicted edge lacks a reference label. These limitations reduce the interpretability of agreement and disagreement with reference networks and limit how reliably predictions can be prioritized and interpreted.

Finally, GRN inference in practice depends critically on prior knowledge to anchor and constrain regulatory structure, yet commonly used priors often rely on cell type-specific experimental assays, emphasize promoter-proximal regulatory logic, and do not readily transfer to new species or less-characterized biological systems. Curated databases also remain biased toward non-model organisms, resulting in priors whose quality is highly variable across settings. This thesis aims to address these challenges by developing inference and prior construction frameworks that remain modular across data regimes, support principled model selection, provide uncertainty-aware regulatory estimates, and enable transferable prior knowledge to guide GRN reconstruction when direct regulatory evidence is sparse.

In Chapter \ref{sec:PMF-GRN-Chapter}, we introduce PMF-GRN to address these methodological limitations by posing GRN inference as a probabilistic graphical model. In this framework, observed single cell gene expression data is decomposed into latent variables with explicitly defined distributions, providing a generative interpretation of regulatory structure. This probabilistic formulation decouples the generative model, which encodes assumptions about how TFs regulate genes, from the variational inference procedure used to fit the model, enabling straightforward adaptation to new datasets and modeling assumptions without redesigning the learning algorithm. Using the ELBO objective function, PMF-GRN supports principled model selection by comparing alternative generative models and hyperparameter configurations and selecting an optimal model under explicit performance criteria. Finally, posterior inference yields full distributions over TF-target gene interaction parameters, providing well-calibrated uncertainty estimates that serve as an interpretable measure of confidence in inferred edges, even when comprehensive ground-truth networks are incomplete or unavailable. 

Across single-cell datasets in yeast, human and mouse, we show that PMF-GRN recovers biologically meaningful regulatory structure, performs competitively with or better than state-of-the-art regression-based GRN inference methods, and scales efficiently across systems. At the same time, these experiments highlight a practical limitation that remains even under a principled probabilistic framework. Accurate GRN recovery depends critically on the quality of the prior network used to specify candidate TF-target gene edges, with performance often constrained by the strength and correctness of the signal encoded in this prior. 

In Chapter \ref{sec:GLM-Prior-Chapter}, we introduce GLM-Prior to address this limitation by constructing prior knowledge directly from nucleotide sequence. GLM-Prior fine-tunes the 250 million parameter Nucleotide Transformer \cite{dalla2024nucleotide}, pretrained on the genomes of 850 species, to predict TF-target gene regulatory interactions as a supervised sequence-to-interaction problem. For each TF-gene pair, the model receives associated genic and TF-binding nucleotide sequences together with available interaction labels. These paired sequences are jointly encoded by the transformer and passed through a classification head to produce an interaction probability. Since positive regulatory interactions are sparse, GLM-prior is trained with class-weighted loss and balanced sampling to address extreme class imbalance. By leveraging representations learned across diverse genomes and the transformer's attention mechanisms, GLM-Prior can transfer regulatory information to new cell types and species, capturing regulatory sequence dependencies that are difficult to recover from promoter-proximal, assay-dependent prior construction strategies. 

Across yeast, mouse, and human cell lines, we show that GLM-Prior produces sequence-derived priors that outperform accessibility-based priors in several mammalian contexts. We also show that GLM-Prior generalizes across single-species, transfer learning, and multi-species training regimes, enabling prior construction in understudied systems. More broadly, we observe that when priors are informative, expression-based GRN inference primarily reweights and refines predicted TF-target gene edges rather than discovering them de novo. These results suggest that prior construction, rather than the choice of downstream GRN inference algorithm, is often the dominant bottleneck in GRN recovery.

Taken together, Chapters \ref{sec:PMF-GRN-Chapter} and \ref{sec:GLM-Prior-Chapter} support a dual-stage perspective on GRN inference in which prior construction and expression-based inference play distinct, complementary roles. Prior knowledge provides TF-specific structure that constrains the space of plausible regulatory explanations and anchor interpretability, while probabilistic inference refines this scaffold in a context-dependent manner and quantifies uncertainty in the resulting network. This division of labor directly targets the core constraints emphasized in this thesis, yielding GRN reconstruction pipelines that are more flexible across data regimes, less dependent on heuristic model selection, more interpretable under incomplete evaluation resources, and more transferable through priors learned by fine-tuning foundation models that generalize across species and cellular contexts.

Several directions follow naturally from this work. On the prior construction side, an important next step is to move beyond DNA-only predictors toward foundation models that integrate multiple modalities along the central dogma, including DNA \cite{brixi2025genome}, RNA, and protein \cite{hayes2025simulating}, with the goal of encoding regulatory specificity and context more directly. Another valuable avenue would be to develop a prior framework to jointly fine-tune multiple foundation models, including models that incorporate cell-type and tissue context from spatial transcriptomics \cite{tejada2025nicheformer} in order to construct priors that are explicitly conditioned on cellular contexts. 

These advances in prior modeling naturally motivate parallel progress on the inference side. A natural extension is to incorporate causal and temporal structure more explicitly, moving beyond static network reconstruction toward models that learn how regulatory interactions evolve over time. Such frameworks could support prediction of future regulatory states under perturbations \cite{kim2024large}, providing a more direct route to forecasting how drugs, signaling events, or disease progression reshapes regulatory programs. More broadly, this thesis motivates future GRN frameworks that couple context-aware, transferable priors with uncertainty aware and temporally grounded inference frameworks, yielding regulatory hypotheses that are both biologically grounded and practically testable at genome-scale.










\cleardoublepage
\phantomsection
\bibliographystyle{apalike}
\addcontentsline{toc}{chapter}{Bibliography}

\bibliography{intro-bib, background-bib, PMF-GRN-bib, GLM-Prior-bib, conclusion-bib}


\end{document}